\documentclass[runningheads]{llncs}

\usepackage{eccv}

\usepackage{eccvabbrv}

\usepackage{graphicx}
\usepackage{booktabs}

\usepackage[accsupp]{axessibility}  

\usepackage[pagebackref,breaklinks,colorlinks,citecolor=eccvblue]{hyperref}

\usepackage{orcidlink}
\usepackage{wrapfig}

\usepackage{minitoc}
\usepackage{float}
\usepackage{pifont}
\usepackage{footnote}
\usepackage{enumitem}
\usepackage{bm}
\usepackage{arydshln}
\usepackage{booktabs}
\usepackage{multicol}
\usepackage{multirow}
\usepackage{color}
\usepackage{xcolor}     
\usepackage{colortbl}
\usepackage{soul}
\usepackage{bbding}
\usepackage{makecell}
\usepackage{mathtools}
\usepackage{imakeidx}
\usepackage{setspace}
\usepackage{threeparttable}

\definecolor{orchid}{rgb}{0.85, 0.44, 0.84}
\definecolor{rubinred}{rgb}{0.82, 0.0, 0.28}
\definecolor{flagship}{rgb}{0.93, 0.06, 0.41}
\definecolor{radiologist}{rgb}{0.50, 0.50, 1}

\newcommand{\ourmodel}{{\fontfamily{ppl}\selectfont SynthX}}

\newcommand{\ourbench}{{\fontfamily{ppl}\selectfont BenchX}}

\newcommand{\numofct}{85,355}

\newcommand{\numofmodels}{12}
\newcommand{\numofhospitals}{112}

\newcolumntype{P}[1]{>{\centering\arraybackslash}p{#1}}
\newlength\savewidth

\begin{document}

\doparttoc 
\faketableofcontents 

\title{\ourbench: Benchmarking AI Models for  \\
Cancer Detection and Localization with \\ Demographic and Protocol Biases} 

\titlerunning{\ourbench}

\author{Qi Chen\inst{1} \and
Wenxuan Li\inst{1} \and
Pedro R. A. S. Bassi\inst{1} \and
Xinze Zhou\inst{1} \and
Jakob Wasserthal\inst{2,3} \and
Ibrahim Ethem Hamamci\inst{4,5,6} \and
Sezgin Er\inst{4,5,6} \and
Ashwin Kumar\inst{7} \and
Yiwen Ye\inst{8} \and
Yuhan Wang\inst{9} \and
Yuyin Zhou\inst{10} \and
Akshay S. Chaudhari\inst{7} \and
Curtis Langlotz\inst{7} \and
Kang Wang\inst{11} \and
Yang Yang\inst{11} \and
Alan L. Yuille\inst{1} \and
Zongwei Zhou\inst{1,12}\thanks{Correspondence to: Zongwei Zhou (\href{mailto:zzhou82@jh.edu}{\texttt{zzhou82@jh.edu}}).}
}

\authorrunning{Q. Chen et al.}

\institute{
Johns Hopkins University \and
German Cancer Research Center \and
University Hospital Basel \and
University of Zurich \and
ETH AI Center \and
Istanbul Medipol University \and
Stanford University \and
École Polytechnique Fédérale de Lausanne \and
Nanyang Technological University \and
University of California, Santa Cruz \and
University of California, San Francisco \and
Johns Hopkins Medicine
}

\maketitle

\begin{abstract}
  Artificial intelligence (AI) has achieved remarkable success in medical imaging, but it is widely recognized that these models often perform inconsistently across real-world clinical settings. Such inconsistencies occur when patient demographics and imaging protocols vary, for example, in detecting small tumors, analyzing scans from different contrast phases, or evaluating patients of different ages or sexes. To quantify these inconsistencies, we develop a large-scale, open benchmark of \numofct\ CT scans that systematically evaluates \numofmodels\ tumor-detection AI models across tumor size, location, patient subgroup, and imaging protocol. We leverage large language models (LLMs) to extract and organize subgroup information from clinical data, which makes the analysis both scalable and reproducible. Our benchmark reveals that current state-of-the-art AI models, optimized for average accuracy, perform poorly in rare or underrepresented subgroups, such as young, female African Americans. However, collecting sufficient annotated data for these rare cases is often impractical. The benchmark provide a foundation for building more reliable and robust AI models for tumor detection and highlighting the need for rigorous, subgroup-level evaluation in medical imaging and computer vision. Datasets, code, and models will be publicly released.
  
\end{abstract}

\section{Introduction}
\label{sec:intro}

Bias in data-driven algorithms has long been a recognized challenge. In the 1980s, studies in both medicine and computer vision showed that models trained on narrow populations often failed to generalize. Early face-recognition systems, such as \textit{Eigenfaces} in the late 1980s and early 1990s~\cite{sirovich1987low,turk1991face,phillips2000feret,buolamwini2018gender}, relied mainly on images of adult men and performed poorly on women and other groups. Similar biases existed in biomedical research, where most clinical trials enrolled only men \cite{holdcroft2007gender,mastroianni1994women}, leading to diagnostic errors and prompting the NIH Revitalization Act of 1993 \cite{chen2014twenty}. These early lessons underscore the need to test models across diverse populations---a principle especially crucial in medical imaging, where variations in patients and imaging protocols can strongly affect AI performance.

In the 2020s, while AI has greatly advanced tumor detection, offering more accurate and accessible cancer care worldwide~\cite{xia2022felix,bassi2025scaling,chen2023cancerunit}, equitable performance across patient populations remains a big, unsolved challenge. Recent advances in large-scale medical imaging models demonstrate impressive accuracy when evaluated on single-institution datasets with homogenous patient cohorts (Asian population, for example, in Cao~\etal~\cite{cao2023large} and Hu~\etal~\cite{hu2025ai}). However, there is a significant knowledge gap in how current models handle population and image acquisition protocol diversity. 

\begin{wrapfigure}{r}{0.50\textwidth}
	\vspace{-12mm} 
	\begin{center}
		\includegraphics[width=0.50\textwidth]{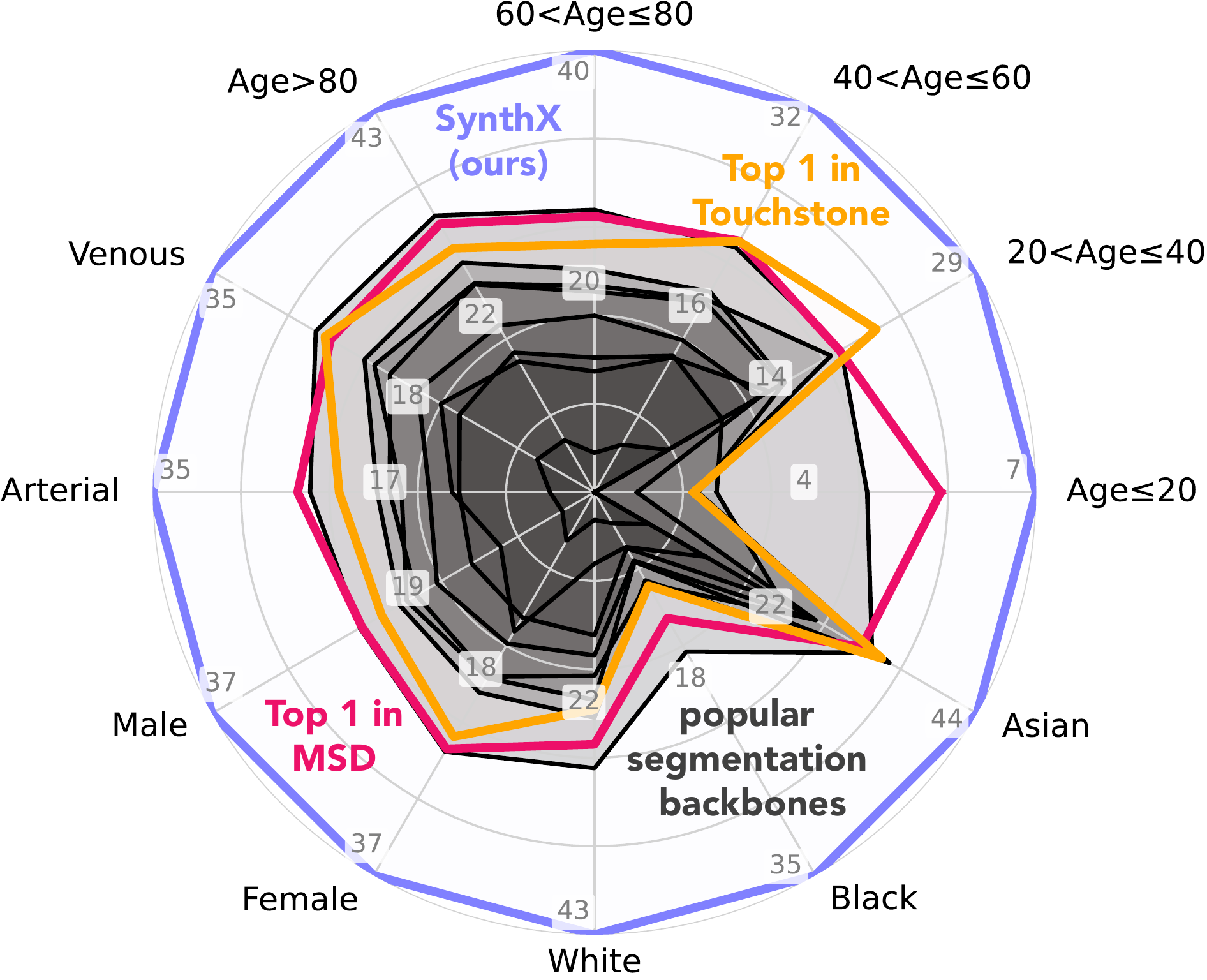}
	\end{center}
	\vspace{-2mm}
	\caption{Existing tumor-detection models, including top leaderboard performers \cite{antonelli2021medical,bassi2024touchstone}, show inconsistent performance across patient demographics (age, race, sex) and imaging protocols (contrast phases). This benchmark (\ourbench) systematically measures and exposes these performance gaps.}
	\vspace{-6mm}
	\label{fig:teaser}
\end{wrapfigure} 


The main bottlenecks are the lack of training data and evaluation benchmarks that explicitly account for demographic and acquisition variability. As a result, most models are implicitly biased toward well-represented groups \cite{bassi2024touchstone}. This paper seeks to answer the critical question: \textit{How can we build AI models that generalize robustly across patient demographics and imaging protocols for tumor detection?} Existing solutions like domain adaptation partially address this issue, but they rely on narrow datasets and fail to assess group-wise performance biases in a systematic way \cite{jain2022unsupervised,yin2024source,hu2024one}.

To address these two bottlenecks, we construct a comprehensive dataset with detailed demographic and protocol factors. Specifically, we establish a benchmark, called \textbf{\ourbench}, that measures model reliability across patient demographics (age, sex, race) and CT imaging protocols (non-contrast, arterial, venous, delayed). To build this benchmark at scale, we leverage large language models (LLMs) \cite{bai2023qwen} to curate and organize data from over 450,000 CT scans from \numofhospitals\ global hospitals. LLMs can automatically read radiology reports and clinical notes quickly and accurately, summarizing key details such as patient characteristics and scan types. This automated process allows us to assemble a large and well-structured dataset of \numofct\ CT scans with precisely matched patient demographics and imaging protocols. By balancing the representation of both common and less common subgroups, the resulting benchmark enables systematic evaluation of model robustness across clinically and demographically varied groups rather than relying on overall average accuracy. Unlike traditional benchmarks such as DeepLesion \cite{yan2018deeplesion}, MSD \cite{antonelli2022medical}, and FLARE \cite{ma2024automatic}, which report only overall accuracy, \ourbench\ reveals where models perform inconsistently across different patient demographics and imaging protocols, and examines large-scale tumor segmentation with more than 10$\times$ the number of test patients.

Our benchmark reveals clear and consistent patterns of performance degradation across patient demographics and imaging protocols. As shown in \figureautorefname~\ref{fig:teaser}, all widely used segmentation backbones, including the top models from MSD \cite{antonelli2022medical,chen2025scaling} and Touchstone \cite{bassi2024touchstone,liu2024vsmtrans}, lose substantial accuracy in young patients, minority racial groups, and non-venous CT phases. These gaps match the quantitative results across six external cohorts (see \figureautorefname~\ref{fig:benchmark_results}). Arterial and non-contrast scans reduce F1 and sensitivity by 15–30\%, pediatric and young-adult patients show the lowest detection accuracy, and minority groups such as Black or Pacific Islander patients experience the largest drops (e.g., F1 as low as 2.5\% on Merlin \cite{blankemeier2024merlin}). Tumor characteristics further magnify these challenges: small tumors and pancreatic tail lesions consistently show the lowest performance across datasets. These results demonstrate that demographics, protocols, and tumor characteristics jointly drive the largest performance drops.

In summary, this paper makes \textbf{two key contributions}:
(1) \ourbench, a large-scale benchmark with fine-grained demographic and protocol annotations that systematically exposes fairness and domain-generalization gaps in tumor detection and localization models, and
(2) Five critical observations through comprehensive evaluation: (i) demographic disparities across age, sex, and race, (ii) performance degradation across CT imaging protocols, (iii) amplified challenges with small tumors and tumor location, (iv) consistent patterns across multiple state-of-the-art models, and (v) validation across six external cohorts demonstrating generalizability of these findings.
\ul{All benchmark data, models, and code will be made publicly available} to promote transparent and equitable medical AI research.

\begin{table}[t]
    \centering
    \caption{\textbf{Related works \& our contributions.}
    Data in \ourbench\ differ significantly across most clinical and imaging variables, including age, sex distribution, image resolution, and contrast phases.
    }
    \label{tab:data_comparison}
    \setlength{\tabcolsep}{1.1pt}
\renewcommand{\arraystretch}{1.1}
    \scriptsize
    \begin{tabular}{p{0.05\linewidth}p{0.25\linewidth}P{0.22\linewidth}P{0.22\linewidth}P{0.22\linewidth}}
        \toprule
        \multicolumn{2}{l}{\makecell[l]{Variable}} & \makecell[c]{\ourbench\ (ours) \\($n$ = \numofct)} & \makecell[c]{MSD-Pancreas \cite{antonelli2021medical}\\($n$ = 281)} & \makecell[c]{Panorama \cite{alves2026artificial}\\($n$ = 2,238)}  \\
        \midrule
        \multicolumn{2}{l}{Age, mean (SD)} &60.5\tiny{~(10.2)}    & - & 65.0\tiny{~(14.0)} \\
        \multicolumn{3}{l}{Sex} & &   \\
         & Female, no. (\%) &41,706\tiny{~(48.9)}   & - & 900\tiny{~(40.2)}\\
         & Male, no. (\%) & 37,846\tiny{~(43.3)}  & - &  1,064\tiny{~(47.5)} \\
         & \textcolor{lightgray}{Unknown, no. (\%)} &\textcolor{lightgray}{5,803\tiny{~(6.8)}}   & \textcolor{lightgray}{281\tiny{~(100.0)}} & \textcolor{lightgray}{274\tiny{~(12.2)}}\\
         \multicolumn{3}{l}{Race} & &   \\
         & Asian, no. (\%) &6,325\tiny{~(7.4)}  & - & - \\
         & Black, no. (\%) &2,478\tiny{~(2.9)}  & - & - \\
         & White, no. (\%) &21,469\tiny{~(25.2)}  & - & - \\
         & \textcolor{lightgray}{Unknown, no. (\%)} & \textcolor{lightgray}{63,886\tiny{~(64.5)}} &\textcolor{lightgray}{281\tiny{~(100.0)}}&\textcolor{lightgray}{2,238\tiny{~(100.0)}}\\ 
        \midrule
        \multicolumn{2}{l}{CT phase} &  &  & \\
         & Non-contrast, no. (\%) & 8,950\tiny{~(10.5)}  & - & - \\
         & Venous, no. (\%) & 29,593\tiny{~(34.7)}  & - & - \\
         & Arterial, no. (\%) & 15,342\tiny{~(18.0)}  & - & - \\
         & Delayed, no. (\%) & 2,219\tiny{~(2.6)}   & - & - \\
         & \textcolor{lightgray}{Unknown, no. (\%)} & \textcolor{lightgray}{29,251\tiny{~(34.3)}} &\textcolor{lightgray}{281\tiny{~(100.0)}}
         &\textcolor{lightgray}{2,238\tiny{~(100.0)}}\\
        \midrule
        \multicolumn{2}{l}{Tumor} & & &  \\
         & Yes, no. (\%) &19,816\tiny{~(23.2)}  & 281\tiny{~(100.0)} & 676\tiny{~(30.2)} \\
         & No, no. (\%) & 65,539\tiny{~(76.8)} & 0\tiny{~(0.0)} & 1,562\tiny{~(69.8)}\\
         \multicolumn{2}{l}{Tumor size} & & &  \\
         & Small, no. (\%) & 6,836\tiny{~(34.5)}  &  110\tiny{~(39.1)} & 248\tiny{~(36.5)}\\
         & Large, no. (\%) & 12,999\tiny{~(65.6)}  &  171\tiny{~(60.9)} & 431\tiny{~(63.5)}\\
         \bottomrule
    \end{tabular}
    \vspace{-2pt}
\end{table}

\section{Related Work}\label{sec:related_works} 
\textbf{Benchmarking AI robustness to biases.}
Bias has long been recognized in pedestrian detection, skin lesion analysis, and speech processing, where models trained on narrow populations tend to underperform on minority groups \cite{ktena2024generative,lotter2024acquisition,glocker2023risk}. These findings highlight that data imbalance can lead to systematic errors unless models are evaluated and improved across diverse subgroups \cite{tian2023fairseg,yang2024limits,wang2024drop}. Bias appears in tumor detection and localization as well, but existing studies remain limited and fragmented. Prior work reports that model accuracy drops for small tumors. In this paper, we take a step beyond average accuracy and perform the first systematic study of demographic and protocol biases in tumor detection. We build a large-scale, comprehensive benchmark that evaluates mainstream AI models across age, sex, race, tumor size, and CT phase. 

\smallskip\noindent\textbf{Benchmarking AI for tumor detection and localization.}
Public benchmarks such as MSD~\cite{antonelli2022medical}, LiTS~\cite{bilic2019liver}, KiTS~\cite{heller2019kits19}, and FLARE~\cite{FLARE23-ma2024automaticorganpancancersegmentation} have advanced tumor detection and localization. Most benchmarks report only overall averages and miss biases from population and protocol differences. Many are single- or few-site and lack key metadata, which prevents systematic analysis. Recent works (Touchstone~\cite{bassi2024touchstone}, OpenMIBOOD~\cite{gutbrod2025openmibood}) reveal generalization gaps but still do not assess group-wise reliability. In this paper, \ourbench\ adds structured annotations (age, sex, race, CT phase, scanner type, voxel spacing) and enables large-scale, comprehensive evaluation across demographic and protocol factors. This provides a fuller view of robustness and bias than accuracy-only reports and reveals specific weaknesses in current models.

\section{Dataset}\label{sec:Dataset}
\subsection{Dataset Construction}
\label{sec:dataset_construction}
\noindent\textbf{Overview.}
\ourbench\ aggregates 85,355 abdominal CT scans from six clinical cohorts (\textsl{E-Coast}, \textsl{Merlin}, \textsl{PanTS}, \textsl{N--California}, \textsl{S.~Europe}, \textsl{N--Europe}; Table~\ref{tab:dataset_composition}). For each scan, we curate tumor annotations and a unified set of demographic and acquisition metadata to enable subgroup-aware evaluation, including age, sex, race, tumor size, contrast phase, scanner type, voxel spacing, tumor attributes.

\smallskip\noindent\textbf{Preprocessing.}
During dataset construction, all scans are converted and stored in NIfTI (.nii) format for consistency. We convert DICOM intensities to Hounsfield Units (HU) when applicable and clip the intensity values to a fixed range of [-1000, 1000]. All scans are preserved at their original spatial resolution.

\begin{figure}[t]
	\centering
	\includegraphics[width=\linewidth]{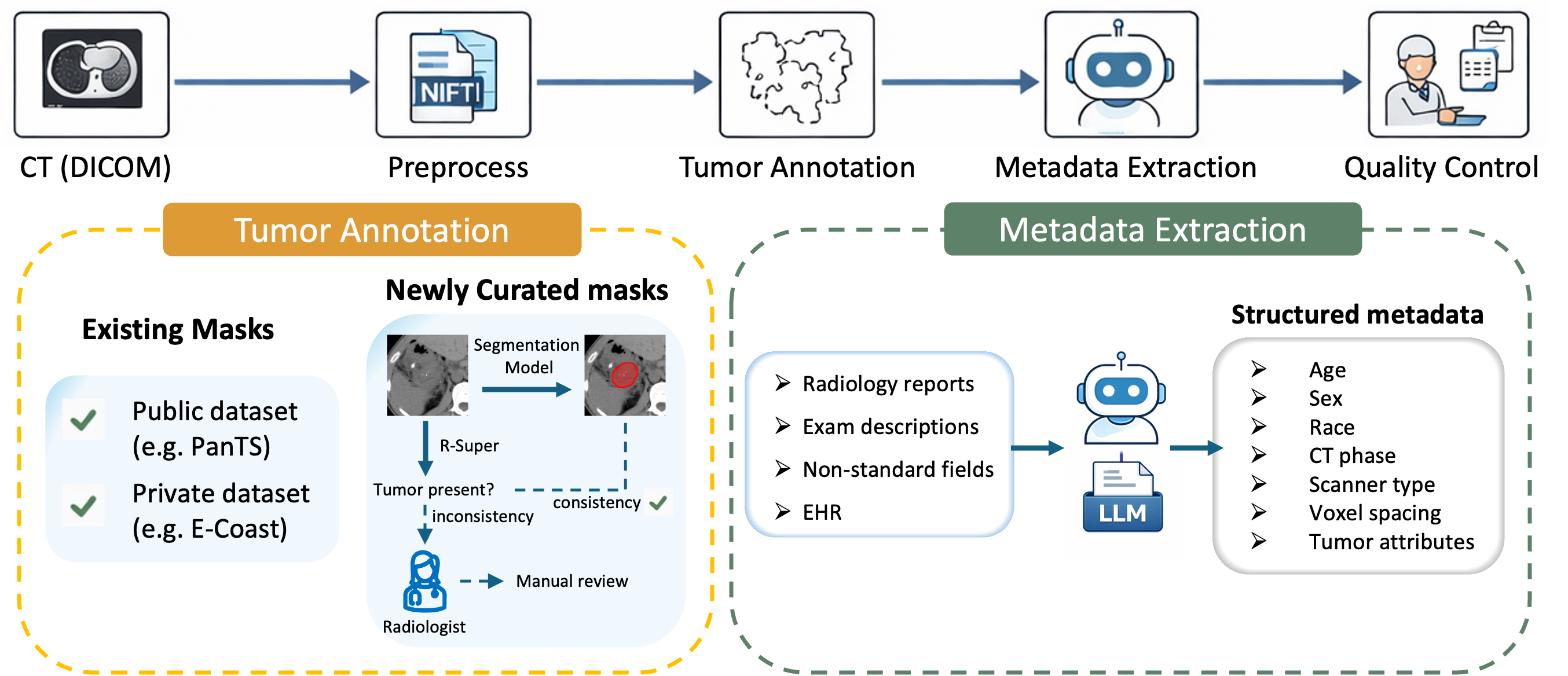}
	\caption{\textbf{The \ourbench\ Curation Pipeline.} We unify \numofct\ heterogeneous scans into a standardized benchmark using a scalable, semi-automated pipeline. Tumor Annotation: A consistency-check loop where only discordant predictions between segmentation and detection models trigger radiologist review, ensuring expert-level ground truth with minimal manual effort. LLM Metadata Extraction: An automated parser converts unstructured clinical reports into fine-grained attributes (e.g., race, contrast phase), enabling large-scale analysis of model fairness.
}
    \label{fig:dataset_curation}
\end{figure}

\smallskip\noindent\textbf{Tumor annotations.}
BenchX uses scan-level tumor presence labels, indicating whether each CT scan contains a pancreatic tumor. For cohorts that provide expert-reviewed tumor masks, such as PanTS~\cite{li2025pants} and \textsl{E-Coast}, we directly use the provided annotations and convert them to scan-level labels. For cohorts without tumor masks, we create scan-level labels with a double-check process.
First, an automatic segmentation model proposes tumor regions with a high-sensitivity setting.
Second, an independent detection model (e.g., R-super~\cite{bassi2025learning}) predicts tumor presence at the scan level.
Cases with inconsistent results are flagged for manual review.
Radiologists then check the original CT scan and reports, and decide the final scan-level label. Annotator roles, clinical background, and additional quality checks are provided in the Appendix~\ref{sec:supp_dataset_curation}.

\smallskip\noindent\textbf{Metadata extraction with LLMs.}
Clinical cohorts store subgroup and protocol information in multiple sources, including radiology reports, exam descriptions, non-standard metadata fields, and electronic health record data. These sources are often fragmented and inconsistently formatted, which makes large-scale subgroup analysis difficult. To make metadata curation scalable and reproducible, we use a pre-trained LLM to extract structured attributes from free-text reports. We define a fixed schema that specifies the target fields, including age, sex, race, contrast phase, and tumor attributes. The LLM extracts normalized categorical labels according to this schema. It is instructed to quote the exact text spans that support each attribute and to avoid inferring any information that is not explicitly stated in the report.

\smallskip\noindent\textbf{Quality control.}
We apply a two-stage quality control protocol for extracted metadata.
First, an LLM performs an automated consistency check by re-extracting key attributes from the original report and comparing them with the normalized outputs. If any extracted attribute does not match the source report or breaks the predefined schema rules, such as more than one phase assigned or unsupported labels, the case is flagged for manual review. A human reviewer then checks the original report and corrects the value if needed, or sets it to \texttt{Unknown} when the information is unclear or not explicitly stated. Second, we randomly sample a subset of cases for manual verification and compute attribute level accuracy. We also summarize common error modes, such as ambiguous phase descriptions, missing demographic mentions, and inconsistent terminology. Overall, the observed accuracy and error rates fall within an acceptable range for large scale metadata curation. Detailed statistics and representative examples are provided in the Appendix~\ref{sec:supp_dataset_curation}.

\subsection{Dataset Statistical Analysis}

\begin{table}[t]
\centering
\begin{threeparttable}
\scriptsize
\caption{\textbf{Dataset composition in \ourbench.}
\label{tab:dataset_composition}
\ourbench\ aggregates \textbf{85,355} CT scans drawn from six large-scale clinical cohorts. 
Together, these datasets span broad demographic, geographic, and imaging diversity. 
We report dataset-level population statistics, race coverage, CT phase availability, and each dataset’s proportional contribution to the full benchmark.}
\renewcommand{\arraystretch}{1.15}

\begin{tabular}{
p{0.13\linewidth}  
P{0.1\linewidth}  
P{0.08\linewidth}  
P{0.08\linewidth}  
P{0.08\linewidth}  
p{0.22\linewidth}  
P{0.13\linewidth}  
P{0.12\linewidth}  
}
\toprule
\textbf{Dataset} & \textbf{\# Scans} & \textbf{Male} & \textbf{Female} & \textbf{Age} & \textbf{Race} & \textbf{Phase}$^{\ddag}$ & \textbf{Prop. (\%)} \\
\midrule

\rowcolor{eccvblue!10}
\textbf{\ourbench} & \textbf{85,355} & \textbf{37,846} & \textbf{41,706} & \textbf{0--113} &
Asian, Black, White, Hispanic, American Indian, Native Hawaiian, Pacific Islander, African, Arab, Balkan, Georgian, Russian, Turkish &
\textbf{A, V, P, D} & \textbf{Full} \\
\midrule

\textsl{E-Coast} & 5,102 & 1,744 & 2,198 & 16--89 &
Asian, Black, White, Hispanic & A, V & 6.0 \\

\rowcolor{gray!5}
Merlin \cite{blankemeier2024merlin} & 25,393 & 11,100 & 14,291 & 1--113 &
Asian, Black, White, American Indian, Pacific Islander & A, V, P & 29.7 \\

PanTS \cite{li2025pants} & 9,901 & 2,924 & 2,358 & 0--98 &
Unknown$^{\dag}$ & A, V, P, D & 11.6 \\

\rowcolor{gray!5}\textsl{N-California} & 42,096 & 20,687 & 21,387 & 2--101 &
Asian, Black, White, American Indian, Native Hawaiian & A, V, P & 49.3 \\

\textsl{S-Europe} & 1,043 & 523 & 520 & 14--95 &
African, Arab, Balkan, Georgian, Russian, Turkish & A, V, P & 1.2 \\

\rowcolor{gray!5}
\textsl{N-Europe} & 1,820 & 868 & 952 & 30--101 &
Unknown$^{\dag}$ & A, V, P & 2.1 \\
\bottomrule
\end{tabular}

\begin{tablenotes}
\item[$^{\dag}$] Race metadata was not collected for these datasets.
\item[$^{\ddag}$] A: Arterial, V: Venous, P: Plain (non-contrast), D: Delayed.
\end{tablenotes}
\end{threeparttable}
\end{table}

\begin{figure}[t]
	\centering
	\includegraphics[width=\linewidth]{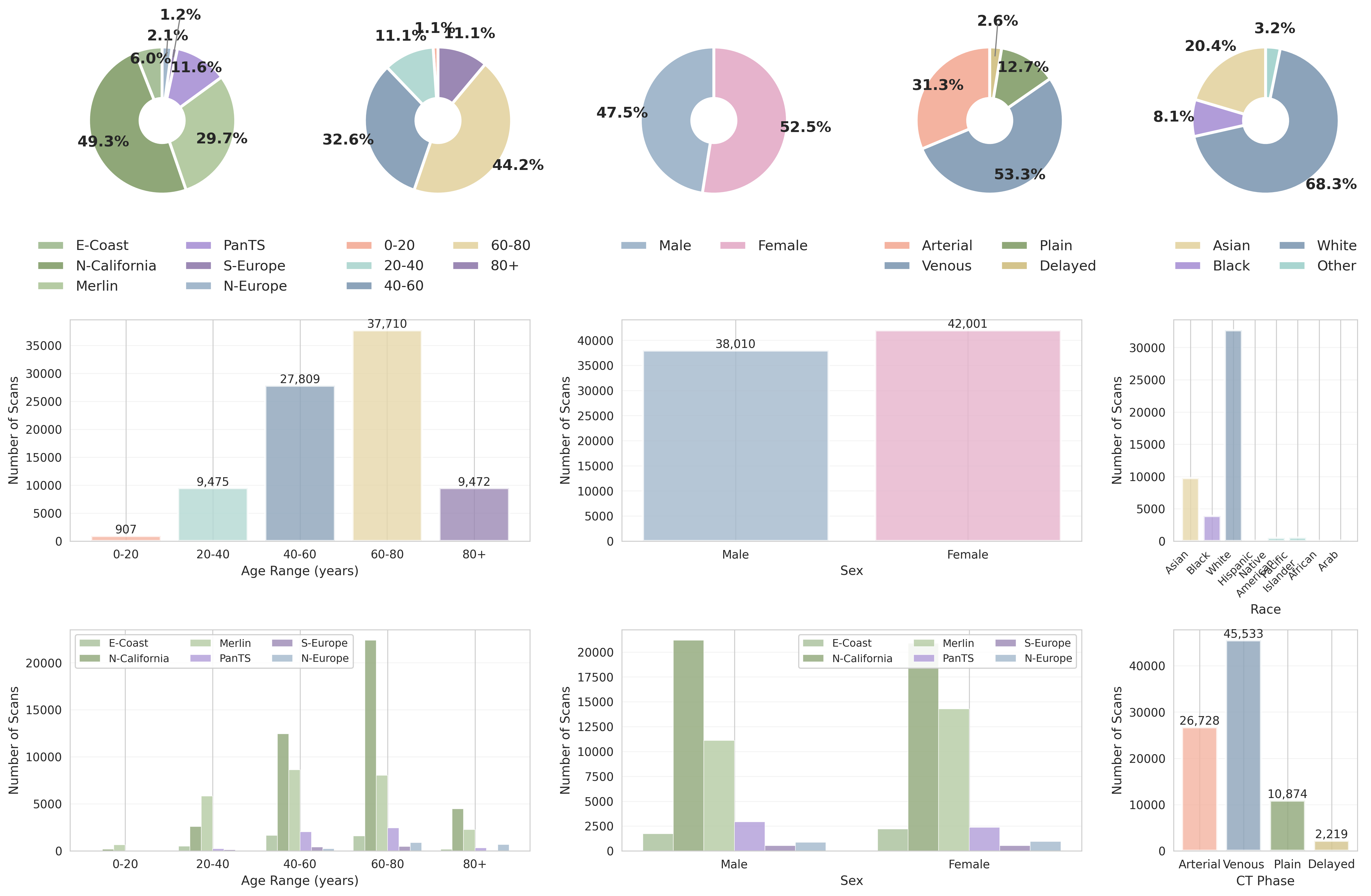}
    \caption{\textbf{Summary of \ourbench\ metadata.} These metadata reveal the breadth and heterogeneity of \ourbench, capturing extensive demographic diversity, protocol variation, and dataset-specific population biases. This comprehensive characterization underscores the importance of subgroup-aware evaluation and motivates the need for robust, fair, and generalizable AI models in medical imaging.}
\label{fig:metadata_distribution}
\end{figure}

We use six external datasets to perform a comprehensive out-of-distribution (OOD) benchmark evaluation. These datasets were collected from hospitals across the globe. \tableautorefname~\ref{tab:data_comparison} illustrates the differences between \ourbench\ and existing pancreatic tumor datasets in terms of tumor characteristics, patient demographics, and imaging protocols. MSD-Pancreas and the Panorama dataset provide only limited metadata, such as sex and tumor size, with the latter derived from the tumor mask. \ourbench\ provides richer metadata, including five additional types compared with MSD-Pancreas. \figureautorefname~\ref{fig:metadata_distribution} also provides a clear and intuitive visualization of the metadata distributions.

\tableautorefname~\ref{tab:dataset_composition} summarizes \ourbench's composition of six cohorts: four cohorts collected by us (\textsl{E-Coast}, \textsl{N-California}, \textsl{S-Europe}, and \textsl{N-Europe}) and two public cohorts (Merlin and PanTS). For Merlin, we annotate scan-level tumor presence labels to support tumor detection evaluation. Representative examples spanning diverse demographics and protocols are shown in Appendix~\ref{sec:vis_E-Coast_appendix}--\ref{sec:vis_n-europe_appendix}. \ul{We will release the benchmark and its datasets to the community.}

\textbf{\textsl{E-Coast}} \textit{$N$=5,102}. USA cohort with venous/arterial phases and rich demographic (age, sex, race) and protocol metadata (phase, scanner, spacing). High resolution (0.5 mm/slice) with per-voxel annotations. 
\textbf{Merlin \cite{blankemeier2024merlin}} \textit{$N$=25,393}. Stanford Hospital abdominal CT collection (mostly venous) with demographic metadata and tumor labels derived from reports. 
\textbf{PanTS \cite{li2025pants}} \textit{$N$=9,901}. Large public dataset from 145 institutions across 12 countries with voxel-wise tumor annotations and demographic/protocol metadata. Validation set: $n$=901.
\textbf{\textsl{N-California}} \textit{$N$=42,096}. Largest cohort by scan count; offers broad demographic coverage (age, sex, race) and diverse protocols (arterial/venous/non-contrast phases). Access restricted.
\textbf{\textsl{S-Europe}} \textit{$N$=1,043}. European cohort (ages 14 to 95) with rich demographic metadata including underrepresented races (African, Arab, Balkan, Georgian, Russian, Turkish) and diverse protocols.
\textbf{\textsl{N-Europe}} \textit{$N$=1,820}. European cohort (ages 30 to 101) with demographic and protocol metadata; race metadata not collected.

\begin{figure*}[t]
	\centering
	\includegraphics[width=\linewidth]{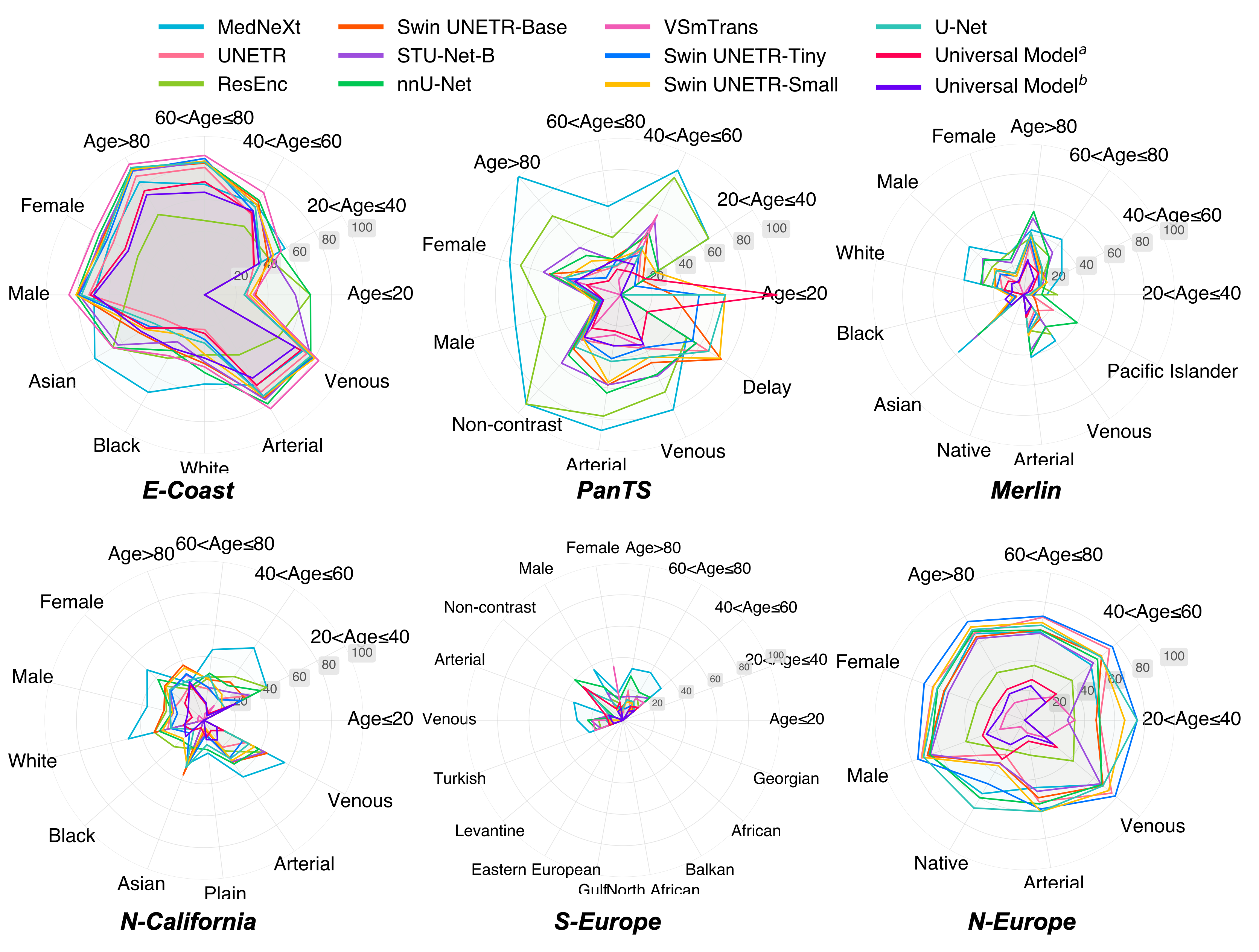}
    \caption{\textbf{Benchmark performance biases in \ourbench.} 
    Performance varies substantially across cohorts, revealing strong dataset and protocol shifts between institutions. Accuracy is highest on E-Coast, PanTS, and N-Europe but drops on Merlin, N-California, and S-Europe. Arterial and non-contrast scans consistently reduce performance relative to venous CT, and younger and minority populations show lower detection accuracy. Although MedNeXt is generally the most robust model, no architecture performs best across all groups, indicating that average metrics hide important subgroup failures. More results are in Appendix~\ref{sec:full_benchmark_results}.}
 
\label{fig:benchmark_results}
\end{figure*}

\section{Benchmark}
\subsection{Evaluation Protocols}\label{sec:evaluation_protocols}
\noindent\textbf{Evaluated models.}
In \ourbench, we evaluated \numofmodels\ tumor-detection AI models. Appendices \ref{sec:AI_Description_appendix}--\ref{sec:implementation_details_appendix} provide in-depth descriptions and configuration details for each architecture and framework. We used the same hardware configuration to evaluate all submitted models. Its specifications are: CPU: AMD EPYC 7713 @ 2.0,GHz $\times$ 64; GPU: NVIDIA Ampere A100 (80,GB); RAM: 2,TB. 

\smallskip\noindent\textbf{Evaluation Metrics.}
All evaluated methods are tumor \emph{segmentation} models trained with voxel-wise tumor masks. In this paper, we report \emph{tumor detection and localization} as a by-product, using scan-level labels that indicate whether each CT scan contains a pancreatic tumor. This scan-level criterion, \emph{if a tumor is present, we detect it}, is widely used in recent works~\cite{xia2022felix, cao2023large} and public benchmarks such as PANORAMA~\cite{alves2024panorama}. Given a CT scan, a model outputs a tumor prediction mask. We convert the mask to a scan-level prediction using a simple rule: the scan is marked as tumor-positive if the count of predicted tumor voxels exceeds a fixed threshold $m$; otherwise, it is tumor-negative. A scan is defined as a true positive when scan-level prediction is tumor-positive and the ground truth label confirms tumor presence. We report sensitivity, specificity, AUC, and F1-score for tumor detection and localization.

\smallskip\noindent\textbf{Evaluation setup.}
We evaluate \numofmodels\ tumor-detection models that are trained on a public dataset~\cite{chen2025scaling}.
All models are evaluated on \ourbench\ under the same inference setting and hardware configuration, without any additional fine-tuning on \ourbench.
Importantly, the six cohorts in \ourbench\ are mutually exclusive, with no patient overlap across cohorts.

\smallskip\noindent\textbf{Data splits.}
All \ourbench\ data are used only for model evaluation and do not participate in training or fine-tuning. We therefore define two splits, a validation set and a test set. The six cohorts come from different institutions, so there is no patient overlap across cohorts. Within each cohort, we perform a patient-level random split, where all scans from the same patient are kept in the same split. We use a 1:9 ratio for validation and test.

\smallskip\noindent\textbf{Handling Missing Metadata.}
Not all patients have complete demographic or protocol information, which is common in real-world clinical practice. For certain attributes such as race or contrast phase, metadata may be unavailable or not explicitly documented. In \ourbench, subgroup evaluation is conducted only on cases with definitive labels for the corresponding factor. Cases labeled as \textit{Unknown} are excluded when computing subgroup-specific metrics.
\subsection{Comprehensive Benchmark Analysis}
\label{sec:comprehensive_bench_analysis}

We evaluate model performance on \ourbench\ across six external cohorts using tumor detection F1 (\figureautorefname~\ref{fig:benchmark_results}), and report full detection metrics in Appendix~\ref{sec:full_benchmark_results} (Tables~\ref{tab:benchmark_jhh}--\ref{tab:benchmark_totalsegmentator}). Overall, performance varies widely across cohorts, showing strong site and protocol shifts. MedNeXt is the most stable model overall, but no single model is best for every group, so average scores can hide subgroup failures.

\smallskip\noindent\textbf{1) Cohort difficulty varies.}
Across models and factors, \textsl{Merlin}, \textsl{N-California}, and \textsl{S-Europe} are consistently more challenging than \textsl{E-Coast}, \textsl{PanTS}, and \textsl{N-Europe}. The best model reaches top-1 subgroup F1 of $\sim$70--74\% on \textsl{E-Coast}, \textsl{PanTS} and \textsl{N-Europe}, but only $\sim$30--35\% on \textsl{Merlin}, \textsl{N-California} and \textsl{S-Europe}, indicating large distribution shifts across cohorts and protocols.

\smallskip\noindent\textbf{2) Architectural insights.}
Aggregating results over all six cohorts and reported factors, MedNeXt achieves the best overall performance. VSmTrans and ResEnc form the next tier and remain competitive across multiple cohorts. This ordering is consistent across all detection metrics (Sen./Spe./AUC/F1).

\smallskip\noindent\textbf{3) Diversified OOD evaluation is necessary.}
For the same model, absolute performance can change by $>$30 F1 points across cohorts. These large shifts lead to unstable rankings under single-cohort testing, highlighting the need for diversified external evaluation.

\smallskip\noindent\textbf{4) No model is uniformly strong across factors within a cohort.}
Within each cohort, subgroup performance differs across demographic and protocol factors. For example, arterial (and when available non-contrast) phases reduce Sen./F1 by $\sim$15--30 points compared to venous on \textsl{E-Coast} and \textsl{N-Europe}, even for top-performing models.

\smallskip\noindent\textbf{5) Consistent hard factors across cohorts.}
Across the six cohorts, arterial and non-contrast phases are among the hardest protocol groups, and younger age groups tend to have lower detection accuracy. 

\smallskip\noindent\textbf{6) Error pattern in hard groups.}
In many difficult subgroups, specificity remains relatively high while sensitivity drops, suggesting that errors are mainly missed positives. For cohorts with race labels, some minority groups also show pronounced F1 drops.

\subsection{Case Study on E-Coast: Age and Sex}
\label{sec:case_study_e-coast}

\begin{table}[t]
    \centering
    \tiny
    \caption{\textbf{Validation on \textsl{E-Coast} across age and sex subgroups.}
Subgroup analysis reveals substantial performance disparities across 
\colorbox{orange!20}{age}, with markedly lower F1 scores for younger patients 
(Age $\leq$20 and 20$<$Age$\leq$40). 
Detection performance across \colorbox{blue!20}{sex} shows minimal variation.}

    \begin{tabular}{p{0.16\linewidth}P{0.044\linewidth}P{0.044\linewidth}P{0.044\linewidth}P{0.044\linewidth}P{0.044\linewidth}P{0.044\linewidth}P{0.044\linewidth}P{0.044\linewidth}P{0.044\linewidth}P{0.044\linewidth}P{0.044\linewidth}P{0.044\linewidth}P{0.044\linewidth}P{0.044\linewidth}P{0.044\linewidth}P{0.044\linewidth}} 
    \toprule
    & \multicolumn{4}{c}{\cellcolor{orange!20}Age $\leq 20$} & \multicolumn{4}{c}{\cellcolor{orange!20}$20 < $ Age $\leq 40$} & \multicolumn{4}{c}{\cellcolor{orange!20}$40 < $ Age $\leq 60$} & \multicolumn{4}{c}{\cellcolor{orange!20}$60 < $ Age $\leq 80$} \\
    \cmidrule(lr){2-5}\cmidrule(lr){6-9}\cmidrule(lr){10-13}\cmidrule(lr){14-17}
    Method & Sen. & Spe. & AUC & F1 & Sen. & Spe. & AUC & F1 & Sen. & Spe. & AUC & F1 & Sen. & Spe. & AUC & F1 \\
    \midrule
    U-Net & 100.0 & 45.5 & 72.8 & 25.0 & 62.5 & 65.0 & 63.8 & 41.0 & 70.2 & 62.2 & 66.2 & 63.9 & 76.0 & 65.2 & 70.6 & 83.6 \\
    Swin UNETR-T & 100.0 & 45.5 & 72.8 & 25.0 & 66.7 & 61.1 & 63.9 & 41.0 & 70.3 & 54.7 & 62.5 & 61.2 & 80.7 & 58.9 & 69.8 & 86.0 \\
    Swin UNETR-S & 100.0 & 54.5 & 77.3 & 28.6 & 66.7 & 75.5 & 71.1 & 50.0 & 69.3 & 68.2 & 68.8 & 65.7 & 76.3 & 72.8 & 74.6 & 84.4 \\
    Swin UNETR-B & 100.0 & 63.6 & 81.8 & 33.3 & 51.0 & 79.8 & 65.4 & 43.8 & 67.1 & 76.3 & 71.7 & 67.7 & 76.2 & 78.1 & 77.2 & 84.7 \\
    Universal Model$^a$ & 0.0 & 63.6 & 31.8 & 0.0 & 55.2 & 62.9 & 59.1 & 36.0 & 65.1 & 59.5 & 62.3 & 59.7 & 59.4 & 57.1 & 58.3 & 71.3 \\
    Universal Model$^b$ & 0.0 & 90.9 & 45.5 & 0.0 & 35.4 & 89.3 & 62.4 & 39.5 & 52.5 & 84.9 & 68.7 & 61.0 & 49.1 & 83.9 & 66.5 & 64.7 \\
    \hline
    nnU-Net & 100.0 & 90.9 & 95.5 & 66.7 & 52.1 & 90.5 & 71.3 & 54.6 & 65.5 & 80.8 & 73.2 & 68.7 & 75.6 & 76.8 & 76.2 & 84.2 \\
    ResEnc & 50.0 & 100.0 & 75.0 & 66.7 & 30.2 & 100.0 & 65.1 & 46.4 & 33.5 & 99.5 & 66.5 & 49.9 & 30.6 & 98.7 & 64.7 & 46.8 \\
    UNETR & 100.0 & 50.0 & 75.0 & 26.7 & 66.7 & 49.1 & 57.9 & 35.6 & 72.9 & 42.4 & 57.7 & 58.7 & 74.1 & 38.4 & 56.3 & 80.4 \\
    STU-Net-B & 100.0 & 86.4 & 93.2 & 57.1 & 47.9 & 87.7 & 67.8 & 48.4 & 64.3 & 75.9 & 70.1 & 65.6 & 75.1 & 75.5 & 75.3 & 83.8 \\
    MedNeXt & 0.0 & 100.0 & 50.0 & 0.0 & 41.7 & 99.7 & 70.7 & 58.4 & 49.9 & 98.9 & 74.4 & 65.9 & 53.7 & 97.8 & 75.8 & 69.7 \\
    VSmTrans & 100.0 & 59.1 & 79.6 & 30.8 & 61.5 & 85.4 & 73.5 & 55.7 & 75.6 & 79.0 & 77.3 & 74.4 & 81.3 & 79.0 & 80.2 & 88.0 \\
    \midrule

    & \multicolumn{4}{c}{\cellcolor{orange!20}Age $> 80$} & \multicolumn{4}{c}{\cellcolor{blue!20}Female} & \multicolumn{4}{c}{\cellcolor{blue!20}Male} & \multicolumn{4}{c}{Avg.} \\
    \cmidrule(lr){2-5}\cmidrule(lr){6-9}\cmidrule(lr){10-13}\cmidrule(lr){14-17}
    Method & Sen. & Spe. & AUC & F1 & Sen. & Spe. & AUC & F1 & Sen. & Spe. & AUC & F1 & Sen. & Spe. & AUC & F1 \\
    \midrule
    U-Net & 86.4 & 100.0 & 93.2 & 92.7 & 72.2 & 64.9 & 68.6 & 72.0 & 76.7 & 60.4 & 68.6 & 77.6 & 74.9 & 66.0 & 71.3 & 65.0 \\
    Swin UNETR-T & 83.3 & 50.0 & 66.7 & 90.6 & 72.0 & 60.5 & 66.3 & 70.7 & 82.5 & 51.0 & 66.8 & 79.1 & 77.0 & 54.2 & 65.6 & 63.4 \\
    Swin UNETR-S & 84.6 & 50.0 & 67.3 & 91.3 & 70.3 & 70.2 & 70.3 & 72.3 & 78.7 & 71.1 & 74.9 & 81.1 & 75.8 & 66.0 & 71.6 & 68.6 \\
    Swin UNETR-B & 85.2 & 100.0 & 92.6 & 92.0 & 68.3 & 81.7 & 75.0 & 74.6 & 78.0 & 70.4 & 74.2 & 80.5 & 72.6 & 78.0 & 76.8 & 70.7 \\
    Universal Model$^a$ & 61.1 & 100.0 & 80.6 & 75.9 & 52.6 & 64.1 & 58.4 & 57.9 & 70.0 & 53.7 & 61.9 & 72.0 & 52.8 & 63.7 & 58.7 & 52.5 \\
    Universal Model$^b$ & 57.4 & 100.0 & 78.7 & 72.9 & 42.3 & 87.8 & 65.1 & 55.6 & 58.5 & 82.9 & 70.7 & 69.8 & 42.2 & 88.5 & 65.1 & 51.2 \\
    \hline
    nnU-Net & 85.8 & 100.0 & 92.9 & 92.4 & 70.7 & 82.7 & 76.7 & 76.5 & 74.0 & 82.8 & 78.4 & 80.8 & 74.8 & 86.4 & \textbf{80.5} & 74.3 \\
    ResEnc & 41.4 & 100.0 & 70.7 & 58.5 & 32.3 & 99.5 & 65.9 & 48.6 & 32.3 & 99.5 & 65.9 & 48.7 & 35.7 & \textbf{99.6} & 68.3 & 50.8 \\
    UNETR & 76.5 & 50.0 & 63.3 & 86.4 & 69.9 & 49.5 & 59.7 & 66.4 & 77.5 & 34.0 & 55.8 & 72.9 & 72.1 & 44.8 & 58.1 & 53.3 \\
    STU-Net-B & 82.7 & 50.0 & 66.4 & 90.2 & 69.4 & 79.3 & 74.4 & 74.6 & 73.1 & 77.8 & 75.5 & 79.0 & 69.2 & 76.0 & 73.3 & 67.0 \\
    MedNeXt & 69.8 & 100.0 & 84.9 & 82.2 & 52.6 & 98.8 & 75.7 & 68.5 & 53.6 & 99.2 & 76.4 & 69.6 & 48.2 & 99.5 & 73.8 & 63.5 \\
    VSmTrans & 90.7 & 100.0 & 95.4 & 95.2 & 78.4 & 77.5 & 78.0 & 79.8 & 80.4 & 85.5 & 83.0 & 85.5 & \textbf{81.2} & 80.8 & 81.3 & \textbf{75.6 }\\
    \bottomrule
    \end{tabular} 
    \label{tab:benchmark_jhh_age_sex}
\end{table}

Age and sex are key patient attributes that relate to body structure and can affect CT appearance and disease presentation. Among the six cohorts in \ourbench, \textsl{E-Coast} provides relatively complete metadata for age and sex, enabling subgroup evaluation with minimal missing labels. As shown in Table~\ref{tab:benchmark_jhh_age_sex}, VSmTrans achieves the best and most balanced performance across both age and sex dimensions. 
nnU-Net serves as a robust second choice. 
ResEnc, UNETR, and Universal~Model each suffer from distinct sensitivity--specificity trade-off limitations. 

\smallskip\noindent\textbf{Age.}
The hardest age group is Age $\leq$ 20. Many models have low F1 in this group (often $\leq$ 33.3), and several methods fail completely with F1 $=$ 0.0 (Universal Model$^a$, Universal Model$^b$, and MedNeXt), mainly due to very low sensitivity. A few models remain stable, such as nnU-Net and ResEnc (both 66.7 F1), showing that the youngest group is difficult but not impossible. The 20 $<$ Age $\leq$ 40 group is also challenging, with F1 ranging from 35.6 to 58.4. Performance improves with age: the 40--60 group reaches 49.9 to 74.4, the 60--80 group reaches 46.8 to 88.0, and multiple models exceed 90 in Age $>$ 80.

\smallskip\noindent\textbf{Sex.}
Sex differences are smaller than age differences on \textsl{E-Coast}. For most models, Female and Male F1 differs by less than 10 points, for example ResEnc (48.6 vs.\ 48.7), nnU-Net (76.5 vs.\ 80.8), and MedNeXt (68.5 vs.\ 69.6). Universal Models show a larger gap (Female 55.6--57.9, Male 69.8--72.0).

\smallskip\noindent\textbf{Clinical justification.}
Younger patients have smaller organs and thinner anatomical structures. As a result, the pancreas occupies fewer voxels and tumor boundaries are less clear at the same scan resolution. Small lesions can be easily mixed with nearby vessels and normal tissue, which increases missed detections and makes segmentation harder.

\subsection{Case Study on N--California: Contrast Enhancement Phase}
\label{sec:case_study_pants}

\begin{table*}[t]
    \centering
    \scriptsize
    \caption{\textbf{Phase subgroup analysis on N--California}. 
    Performance is reported across venous, non-contrast, and arterial CT phases, along with the phase-averaged metrics. 
    Venous scans consistently yield the highest sensitivity and F1, while non-contrast and arterial scans pose substantially greater challenges, with many models exhibiting near-zero sensitivity on non-contrast inputs.}
    \begin{tabular}{p{0.27\linewidth}P{0.07\linewidth}P{0.07\linewidth}P{0.09\linewidth}P{0.07\linewidth}P{0.07\linewidth}P{0.07\linewidth}P{0.09\linewidth}P{0.07\linewidth}} 
    \toprule
    \multicolumn{1}{c}{\cellcolor{purple!20}\textbf{Phase Groups}} & \multicolumn{4}{c}{\cellcolor{purple!20}Venous} & \multicolumn{4}{c}{\cellcolor{purple!20}Non-Contrast} \\
    \cmidrule(lr){2-5}\cmidrule(lr){6-9}
    Method & Sen. & Spe. & AUC & F1 & Sen. & Spe. & AUC & F1 \\
    \midrule
    U-Net~\cite{ronneberger2015u} & 67.9 & 71.5 & 69.7 & 31.9 & 30.6 & 79.7 & 55.2 & 15.5 \\
    Swin UNETR-T~\cite{tang2022self} & 85.7 & 51.8 & 68.8 & 27.6 & 19.4 & 76.9 & 48.2 & 9.3 \\
    Swin UNETR-S~\cite{tang2022self} & 85.7 & 67.6 & 76.7 & 35.8 & 5.6 & 89.3 & 47.5 & 4.5 \\
    Swin UNETR-B~\cite{tang2022self} & 70.4 & 76.9 & 73.7 & 37.6 & 26.5 & 71.4 & 49.0 & 10.9 \\
    Univ.\ Model$^a$~\cite{liu2023clip} & 21.4 & 79.5 & 50.5 & 13.9 & 22.2 & 78.8 & 50.5 & 11.2 \\
    Univ.\ Model$^b$~\cite{liu2023clip} & 10.7 & 86.2 & 48.5 & 9.1 & 13.9 & 91.2 & 52.6 & 12.2 \\
    \hline
    nnU-Net~\cite{isensee2021nnu} & 53.6 & 86.6 & 70.1 & 39.0 & 30.6 & 84.8 & 57.7 & 18.6 \\
    ResEnc~\cite{isensee2024nnu} & 46.4 & 92.5 & 69.5 & 43.3 & 5.6 & 99.6 & 52.6 & 10.0 \\
    UNETR~\cite{hatamizadeh2022unetr} & 64.3 & 70.8 & 67.6 & 30.0 & 8.3 & 89.9 & 49.1 & 7.0 \\
    STU-Net-B~\cite{huang2023stu} & 60.7 & 86.6 & 73.7 & 43.0 & 16.7 & 80.3 & 48.5 & 9.0 \\
    MedNeXt~\cite{roy2023mednext} & 71.4 & 91.3 & 81.4 & 57.1 & 13.9 & 98.5 & 56.2 & 20.8 \\
    VSmTrans~\cite{liu2024vsmtrans} & 7.1 & 97.6 & 52.4 & 11.1 & 0.0 & 100.0 & 50.0 & 0.0 \\
    \midrule

    \multicolumn{1}{c}{} & \multicolumn{4}{c}{\cellcolor{purple!20}Arterial} & \multicolumn{4}{c}{\cellcolor{purple!20}Average} \\
    \cmidrule(lr){2-5}\cmidrule(lr){6-9}
    Method & Sen. & Spe. & AUC & F1 & Sen. & Spe. & AUC & F1 \\
    \midrule
    U-Net~\cite{ronneberger2015u} & 44.0 & 78.8 & 61.4 & 29.3 & 47.5 & 76.7 & 62.1 & 25.6 \\
    Swin UNETR-T~\cite{tang2022self} & 56.0 & 67.9 & 62.0 & 28.6 & 53.7 & 65.5 & 59.7 & 21.8 \\
    Swin UNETR-S~\cite{tang2022self} & 48.0 & 77.7 & 62.9 & 30.8 & 46.4 & 78.2 & 62.4 & 23.7 \\
    Swin UNETR-B~\cite{tang2022self} & 63.6 & 73.4 & 68.5 & 34.1 & 53.5 & 73.9 & 63.7 & 27.5 \\
    Univ.\ Model$^a$~\cite{liu2023clip} & 8.0 & 87.0 & 47.5 & 7.8 & 17.2 & 81.8 & 49.5 & 11.0 \\
    Univ.\ Model$^b$~\cite{liu2023clip} & 12.0 & 93.5 & 52.8 & 15.0 & 12.2 & 90.3 & 51.3 & 12.1 \\
    \hline
    nnU-Net~\cite{isensee2021nnu} & 40.0 & 86.4 & 63.2 & 33.3 & 41.4 & 85.9 & 63.7 & 30.3 \\
    ResEnc~\cite{isensee2024nnu} & 16.0 & 97.3 & 56.7 & 23.5 & 22.7 & 96.5 & 59.6 & 25.6 \\
    UNETR~\cite{hatamizadeh2022unetr} & 28.0 & 80.4 & 54.2 & 20.6 & 33.5 & 80.4 & 57.0 & 19.2 \\
    STU-Net-B~\cite{huang2023stu} & 32.0 & 90.2 & 61.1 & 31.4 & 36.5 & 85.7 & 61.1 & 27.8 \\
    MedNeXt~\cite{roy2023mednext} & 33.3 & 97.3 & 65.3 & 43.2 & 39.5 & 95.7 & 67.6 & 40.4 \\
    VSmTrans~\cite{liu2024vsmtrans} & 0.0 & 100.0 & 50.0 & 0.0 & 2.4 & 99.2 & 50.8 & 3.7 \\
    \bottomrule
    \end{tabular} 
    \label{tab:phase_analysis}
\end{table*}

As shown in Table~\ref{tab:phase_analysis}, the contrast phase has a strong impact on detection performance. Venous phase scans provide the most favorable imaging condition, under which most models achieve their highest sensitivity and F1. For instance, MedNeXt~\cite{roy2023mednext} achieves 71.4\% sensitivity and 57.1\% F1 on venous phase scans, substantially outperforming the other evaluated architectures. Non-contrast scans present the greatest difficulty: without contrast enhancement, the lesion to background contrast diminishes significantly, causing sensitivity to approach zero for several models (\eg, ResEnc~\cite{isensee2024nnu} 5.6\%, VSmTrans~\cite{liu2024vsmtrans} 0\%). Arterial phase scans are slightly more informative than noncontrast scans, but they still trail venous phase performance, likely because enhancement is shorter and less uniform.

Among all evaluated models, MedNeXt achieves the highest phase averaged F1 of 40.4\%, while maintaining consistently high specificity above 95\% across all three phases, indicating a favorable balance between detection and false positive control. nnU-Net~\cite{isensee2021nnu} and STU-Net-B~\cite{huang2023stu} deliver moderate and relatively stable performance (average F1 of 30.3\% and 27.8\%, respectively). In contrast, ResEnc exhibits a highly conservative prediction pattern: its average specificity reaches 96.5\%, but sensitivity is substantially reduced (22.7\%). Both variants of Universal Model~\cite{liu2023clip} and VSmTrans yield the lowest phase averaged F1 scores, rendering them unsuitable under this evaluation setting.

Overall, these results support two key findings. First, the contrast phase is a dominant factor in model performance, with venous scans being the most amenable to automated detection. Second, among the compared methods, MedNeXt demonstrates the strongest robustness to phase variation and the highest overall F1, making it the most competitive architecture in this benchmark.

\smallskip\noindent\textbf{Clinical  justification.} CT appearance changes with contrast timing.
Non-contrast scans show baseline attenuation without enhancement, so tumors may have weak intensity difference from surrounding tissue.
Arterial phase highlights early enhancement and vessels, which can increase local contrast but also introduce strong vessel signals that may confuse models. Venous phase usually provides more uniform organ enhancement and clearer soft-tissue boundaries, often giving more stable tumor visibility. 

\subsection{Case Study on Merlin: Race}
\label{sec:case_study_merlin}

\begin{table*}[t]
    \centering
    \tiny
    \caption{\textbf{Race subgroup analysis on Merlin.} 
    Performance across race groups reveals substantial disparities. 
    Asian patients consistently achieve the highest F1 scores, 
    while Black and Pacific Islander patients suffer severe performance degradation across nearly all methods. 
    White patients also show pronounced drops compared to Asian patients.}
    \begin{tabular}{p{0.16\linewidth}P{0.044\linewidth}P{0.044\linewidth}P{0.044\linewidth}P{0.044\linewidth}P{0.044\linewidth}P{0.044\linewidth}P{0.044\linewidth}P{0.044\linewidth}P{0.044\linewidth}P{0.044\linewidth}P{0.044\linewidth}P{0.044\linewidth}P{0.044\linewidth}P{0.044\linewidth}P{0.044\linewidth}P{0.044\linewidth}}
    \toprule
    & \multicolumn{4}{c}{\cellcolor{green!20}Asian} & \multicolumn{4}{c}{\cellcolor{green!20}White} & \multicolumn{4}{c}{\cellcolor{green!20}Black} & \multicolumn{4}{c}{\cellcolor{green!20}Pacific Isl.} \\
    \cmidrule(lr){2-5}\cmidrule(lr){6-9}\cmidrule(lr){10-13}\cmidrule(lr){14-17}
    Method & Sen. & Spe. & AUC & F1 & Sen. & Spe. & AUC & F1 & Sen. & Spe. & AUC & F1 & Sen. & Spe. & AUC & F1 \\
    \midrule
    U-Net & 54.5 & 80.7 & 67.6 & 31.6 & 47.9 & 68.7 & 58.3 & 21.5 & 0.0 & 72.5 & 36.3 & 0.0 & 0.0 & 56.2 & 28.1 & 0.0 \\
    Swin UNETR-T & 54.5 & 56.9 & 55.7 & 18.8 & 56.2 & 52.1 & 54.2 & 18.4 & 50.0 & 52.2 & 51.1 & 5.6 & 0.0 & 43.8 & 21.9 & 0.0 \\
    Swin UNETR-S & 63.6 & 72.5 & 68.1 & 29.2 & 50.0 & 62.6 & 56.3 & 19.8 & 50.0 & 65.2 & 57.6 & 7.4 & 0.0 & 50.0 & 25.0 & 0.0 \\
    Swin UNETR-B & 45.5 & 76.2 & 60.9 & 23.8 & 45.8 & 64.8 & 55.3 & 19.1 & 50.0 & 73.9 & 62.0 & 9.5 & 100.0 & 43.8 & 71.9 & 18.2 \\
    Universal Model$^a$ & 0.0 & 90.8 & 45.4 & 0.0 & 20.8 & 79.4 & 50.1 & 13.2 & 0.0 & 79.7 & 39.9 & 0.0 & 0.0 & 50.0 & 25.0 & 0.0 \\
    Universal Model$^b$ & 9.1 & 93.6 & 51.4 & 10.5 & 8.3 & 88.6 & 48.5 & 7.7 & 50.0 & 81.2 & 65.6 & 12.5 & 0.0 & 68.8 & 34.4 & 0.0 \\
    \hline
    nnU-Net & 54.5 & 88.1 & 71.3 & 40.0 & 45.8 & 82.3 & 64.1 & 29.1 & 0.0 & 82.6 & 41.3 & 0.0 & 100.0 & 81.2 & 90.6 & 40.0 \\
    ResEnc & 36.4 & 97.2 & 66.8 & 44.4 & 20.8 & 96.9 & 58.9 & 27.8 & 0.0 & 94.2 & 47.1 & 0.0 & 0.0 & 100.0 & 50.0 & 0.0 \\
    UNETR & 54.5 & 62.4 & 58.5 & 20.7 & 41.7 & 54.7 & 48.2 & 14.6 & 0.0 & 62.3 & 31.2 & 0.0 & 100.0 & 56.2 & 78.1 & 22.2 \\
    STU-Net-B & 54.5 & 90.8 & 72.7 & 44.4 & 39.6 & 84.9 & 62.3 & 27.9 & 0.0 & 87.0 & 43.5 & 0.0 & 0.0 & 75.0 & 37.5 & 0.0 \\
    MedNeXt & 60.0 & 95.2 & 77.6 & 57.1 & 40.0 & 94.5 & 67.3 & 40.5 & 0.0 & 91.2 & 45.6 & 0.0 & 0.0 & 92.9 & 46.5 & 0.0 \\
    VSmTrans & 0.0 & 98.2 & 49.1 & 0.0 & 2.1 & 99.1 & 50.6 & 3.8 & 0.0 & 100.0 & 50.0 & 0.0 & 0.0 & 100.0 & 50.0 & 0.0 \\
    \bottomrule
    \end{tabular}
    \label{tab:race_groups_merlin}
\end{table*}
Among the six cohorts in \ourbench, \textsl{Merlin} offers the most comprehensive race annotations across multiple groups, enabling a detailed race stratified evaluation. Table~\ref{tab:race_groups_merlin} reveals a pronounced performance gap across racial groups. Across most models, Asian patients achieve the highest detection performance, followed by White patients, whereas Black and Pacific Islander groups exhibit the largest declines. Notably, many models attain 0.0 sensitivity and F1 on the Black group, indicating frequent missed positive cases within this subgroup.

MedNeXt achieves the highest F1 on Asian patients and also leads on White patients, but its sensitivity falls to 0\% for both Black and Pacific Islander groups. nnU Net is more robust for the Pacific Islander group, achieving 100.0\% sensitivity and 40.0\% F1, yet it still fails on the Black group. For Black patients, the only nonzero sensitivities are primarily observed in Swin UNETR variants, yet their F1 scores remain very low.

\smallskip\noindent\textbf{Clinical justification.}
Race groups in real clinical data can differ in sample size, imaging practice, and patient body habitus, which can change CT appearance and tumor visibility. When some groups are underrepresented in training data, models may not learn stable features for them and can miss positive cases. These factors can contribute to the large subgroup gaps observed on \textsl{Merlin}.

\begin{table*}[t]
    \centering
    \tiny
    \caption{\textbf{Synthetic data reduces demographic and protocol gaps.}
Results on the E-Coast dataset stratified by \colorbox{orange!20}{age}, \colorbox{blue!20}{sex}, \colorbox{purple!20}{phase}, and \colorbox{green!20}{race}. Compared with the baseline, training with synthetic data improves sensitivity in every subgroup and also increases the average results, showing that synthetic augmentation can partially mitigate demographic and protocol biases.}
    \begin{tabular}{p{0.16\linewidth}P{0.06\linewidth}P{0.06\linewidth}P{0.06\linewidth}P{0.06\linewidth}P{0.06\linewidth}P{0.06\linewidth}P{0.05\linewidth}P{0.06\linewidth}P{0.06\linewidth}P{0.06\linewidth}P{0.06\linewidth}P{0.06\linewidth}} 
    \toprule
     & \multicolumn{3}{c}{\cellcolor{orange!20}20 \textless Age $\leq 40$} & \multicolumn{3}{c}{\cellcolor{orange!20}40 \textless Age $\leq 60$} & \multicolumn{3}{c}{\cellcolor{orange!20}60 \textless Age $\leq 80$} & \multicolumn{3}{c}{\cellcolor{orange!20}Age \textgreater 80} \\
    \cmidrule(lr){2-4}\cmidrule(lr){5-7}\cmidrule(lr){8-10}\cmidrule(lr){11-13}
    Method & Sen. & AUC & F1 & Sen. & AUC & F1 & Sen. & AUC & F1 & Sen. & AUC & F1 \\
    \midrule
    MedNeXt & 41.7 & 70.7 & 58.4 & 49.9 & 74.4 & 65.9 & 53.7 & 75.8 & 69.7 & 69.8 & 84.9 & 82.2 \\ 
    \quad + \ourmodel & 60.4 & 76.8 & 64.1 & 78.8 & 83.4 & 81.0 & 86.6 & 86.4 & 91.7 & 95.1 & 97.5 & 97.5 \\ 
    $\Delta$ & \textbf{\textcolor{red}{+18.7}} & \textbf{\textcolor{red}{+6.1}} & \textbf{\textcolor{red}{+5.7}} & \textbf{\textcolor{red}{+28.9}} & \textbf{\textcolor{red}{+9.0}} & \textbf{\textcolor{red}{+15.1}} & \textbf{\textcolor{red}{+32.9}} & \textbf{\textcolor{red}{+10.6}} & \textbf{\textcolor{red}{+22.0}} & \textbf{\textcolor{red}{+25.3}} & \textbf{\textcolor{red}{+12.6}} & \textbf{\textcolor{red}{+15.3}} \\
    \midrule
     & \multicolumn{3}{c}{\cellcolor{blue!20}Female} & \multicolumn{3}{c}{\cellcolor{blue!20}Male} & \multicolumn{3}{c}{\cellcolor{purple!20}Arterial} & \multicolumn{3}{c}{\cellcolor{purple!20}Venous} \\
    \cmidrule(lr){2-4}\cmidrule(lr){5-7}\cmidrule(lr){8-10}\cmidrule(lr){11-13}
    Method & Sen. & AUC & F1 & Sen. & AUC & F1 & Sen. & AUC & F1 & Sen. & AUC & F1 \\
    \midrule
    MedNeXt & 52.6 & 75.7 & 68.5 & 53.6 & 76.4 & 69.6 & 49.1 & 74.2 & 65.7 & 55.5 & 77.1 & 71.1 \\
    \quad + \ourmodel & 81.7 & 84.7 & 85.2 & 85.8 & 88.4 & 90.0 & 82.3 & 85.6 & 87.7 & 79.3 & 84.1 & 85.9 \\ 
    $\Delta$ & \textbf{\textcolor{red}{+29.1}} & \textbf{\textcolor{red}{+9.0}} & \textbf{\textcolor{red}{+16.7}} & \textbf{\textcolor{red}{+32.2}} & \textbf{\textcolor{red}{+12.0}} & \textbf{\textcolor{red}{+20.4}} & \textbf{\textcolor{red}{+33.2}} & \textbf{\textcolor{red}{+11.4}} & \textbf{\textcolor{red}{+22.0}} & \textbf{\textcolor{red}{+23.8}} & \textbf{\textcolor{red}{+7.0}} & \textbf{\textcolor{red}{+14.8}} \\
    \midrule
     & \multicolumn{3}{c}{\cellcolor{green!20}Asian} & \multicolumn{3}{c}{\cellcolor{green!20}Black} & \multicolumn{3}{c}{\cellcolor{green!20}White} & \multicolumn{3}{c}{\textbf{Avg}} \\
    \cmidrule(lr){2-4}\cmidrule(lr){5-7}\cmidrule(lr){8-10}\cmidrule(lr){11-13}
    Method & Sen. & AUC & F1 & Sen. & AUC & F1 & Sen. & AUC & F1 & Sen. & AUC & F1 \\
    \midrule
    MedNeXt & 66.7 & 83.4 & 80.0 & 55.0 & 77.5 & 71.0 & 42.9 & 71.0 & 56.2 & 53.5 & 76.5 & 68.0 \\
    \quad + \ourmodel & 100.0 & 95.9 & 75.0 & 95.0 & 92.6 & 66.7 & 95.2 & 92.2 & 64.0 & 85.5 & 87.8 & 80.8 \\
    $\Delta$ & \textbf{\textcolor{red}{+33.3}} & \textbf{\textcolor{red}{+12.5}} & \textbf{\textcolor{blue}{-5.0}} & \textbf{\textcolor{red}{+40.0}} & \textbf{\textcolor{red}{+15.1}} & \textbf{\textcolor{blue}{-4.3}} & \textbf{\textcolor{red}{+52.3}} & \textbf{\textcolor{red}{+21.2}} & \textbf{\textcolor{red}{+7.8}} & \textbf{\textcolor{red}{+32.0}} & \textbf{\textcolor{red}{+11.3}} & \textbf{\textcolor{red}{+12.8}} \\
    \bottomrule
    \end{tabular}
    \label{tab:combined_results}
\end{table*}

\subsection{Synthetic Data for Domain Adaptation}
\label{sec:synthetic_data}

We hypothesize that synthetic data can help when some demographic groups or scan settings have few training cases. This can reduce subgroup gaps by adding more diverse examples. We test this hypothesis on the \textsl{E-Coast} dataset.

We modified existing conditional diffusion model approaches \cite{wu2024freetumor,chen2024towards,li2024text} to generate synthetic tumors conditioned on patient demographics and imaging protocols. We call this method \ourmodel. Specifically, we adapt existing conditional diffusion models to generate synthetic tumors by conditioning on patient sex, age, and image contrast phases (non-contrast, arterial, venous). This allows us to synthesize tumors with specific demographic and protocol attributes, filling gaps in underrepresented subgroups. Implementation details are provided in the Appendix~\ref{sec:implementation_details_appendix}. We compare two settings: MedNeXt trained on real data only, and the same backbone trained with added SynthX-generated synthetic tumors. Results are shown in \figureautorefname~\ref{fig:teaser} and Table~\ref{tab:combined_results}. Adding synthetic tumors from SynthX increases sensitivity in every reported subgroup. The average results also improve. This demonstrates that our approach helps tumor-detection models generalize better across demographic groups and imaging protocols.

\section{Conclusion}\label{sec:conclusion}

\ourbench\ is a global benchmark built from over \numofct\ CT scans representing different ages, races, and imaging protocols. Our evaluation exposes critical limitations in current tumor detection models that are often overlooked:
\textbf{(1) Age dependent performance degradation is severe.} Younger patients (Age $\leq$ 40) show dramatic performance drops with near-zero sensitivity. Top-performing models completely fail on patients under 20 years old, a vulnerability rarely discussed in medical imaging literature.
\textbf{(2) Imaging protocol is a dominant factor, rivaling architecture.} Venous phase scans yield F1 scores up to 57.1\%, while non-contrast scans produce near-zero sensitivity. Single-protocol benchmarks fundamentally misrepresent real-world performance.
\textbf{(3) Racial disparities are severe and systemic.} Black and Pacific Islander patients exhibit near-zero sensitivity across most top-ranked models, indicating the issue stems from training data imbalance rather than individual model weaknesses.
\textbf{(4) Standard fairness metrics mask catastrophic failures.} Overall performance can appear acceptable (70–80\% F1) while demographic or protocol subgroups collapse to zero sensitivity. Average metrics alone are insufficient for safety-critical medical applications.
\textbf{(5) Synthetic data provides incomplete mitigation.} Adding SynthX-generated data improves sensitivity across subgroups (notably +22.0\% F1 for age 60–80 and White), but gaps persist. Synthetic data should be combined with other fairness-aware strategies. 

\noindent\textbf{Acknowledgments.}
This work was supported by the Lustgarten Foundation for Pancreatic Cancer Research and the National Institutes of Health (NIH) under Award Number R01EB037669. We would like to thank the Johns Hopkins Research IT team in \href{https://researchit.jhu.edu/}{IT@JH} for their support and infrastructure resources where some of these analyses were conducted; especially \href{https://researchit.jhu.edu/research-hpc/}{DISCOVERY HPC}. We thank Jaimie Patterson for writing a \href{}{news article} about this project. Paper content is covered by patents pending.

\clearpage  

%
%
\bibliographystyle{splncs04}
\bibliography{refs,zzhou}

\clearpage
\appendix

\renewcommand \thepart{}
\renewcommand \partname{}

\part{Appendix} 
\setcounter{secnumdepth}{4}
\setcounter{tocdepth}{4}

\renewcommand{\ptctitle}{Appendix}
\renewcommand{\openparttoc}{\vspace{1.2em}}
\parttoc 
\setcounter{page}{1}

\section{Dataset Curation Details}\label{sec:supp_dataset_curation}

\subsection{Tumor Annotations Details}

\noindent\textbf{Annotator roles.}
The final scan-level tumor labels in \ourbench\ were established through a multi-reader review workflow designed to balance scalability and clinical reliability. The annotation team consisted of 8 radiologists with experience in abdominal CT interpretation, including 3 senior radiologists (each with more than 10 years of clinical experience) and 2 junior radiologists. Junior readers first screened cases flagged by the automated consistency-check pipeline. Their task was to verify whether the proposed lesion corresponded to a plausible pancreatic tumor in the original CT volume and to cross-check the imaging evidence with the accompanying radiology report when available. Senior radiologists handled difficult or ambiguous cases, including scans with subtle lesions, low image quality, atypical enhancement patterns, postoperative anatomy, or disagreement between imaging findings and report evidence. These senior readers were responsible for adjudicating uncertain cases and assigning the final scan-level tumor label. Before formal annotation, all annotators followed a shared labeling protocol that defined tumor-positive criteria at the scan level and specified how to treat equivocal findings. The protocol also described common edge cases such as cystic lesions, postoperative anatomy, severe motion artifacts, and incomplete report evidence. During review, annotators had access to the original volumetric CT scan, the automatic tumor proposal generated from segmentation, the independent scan-level detector output, and relevant report text when available.

\smallskip
\noindent\textbf{Additional quality checks.}
Beyond the automated double-check procedure described, we performed several additional quality-control steps. First, 100\% of cases with disagreement between the segmentation-based proposal and the scan-level detector were routed to human review. Second, we randomly sampled approximately 5\% of automatically accepted scans for manual verification to estimate potential error rates of the automatic filtering pipeline. If systematic error patterns were observed, the screening rules were updated and the affected cases were re-evaluated. Third, difficult cases were escalated for senior review. These typically included scans with very small suspected lesions ($<4$mm), poor contrast timing, heavy motion or streak artifacts, or reports with uncertain wording. Fourth, a subset of reviewed cases (10\%) was independently re-checked by a second reader to monitor inter-reader consistency at the scan-label level. Discrepancies were resolved through adjudication by a senior radiologist.

We also conducted targeted audits of common failure modes identified during annotation. These audits focused on false positives caused by vessels, bowel loops, postoperative changes, or inflammatory findings, and false negatives caused by very small tumors, iso-attenuating lesions, or atypical enhancement patterns. For each error category, representative scans were inspected and the review guidelines were refined when necessary. Importantly, the automated models were used only to prioritize cases for review and surface suspicious candidates. They did not determine the final ground truth. The final scan-level label was assigned only after human verification of the original CT scan and supporting report evidence when available.

\subsection{Metadata Extraction Statistics and Quality Analysis}
\label{sec:supp_dataset_curation}

This section provides additional statistics and examples for the metadata extraction pipeline. The goal is to evaluate the reliability of the LLM-based attribute extraction process and illustrate typical extraction outcomes.

\smallskip
\noindent\textbf{Cross-model extraction consistency.}
To assess the robustness of the extraction process, we apply two large language models, \textit{Llama-3.1-70B} and \textit{DeepSeek-R1-Distill-Qwen-32B}, to the same radiology reports using an identical extraction schema. Both models generate structured attributes including patient age, sex, race, and CT contrast phase. Across the dataset, the two models produce identical normalized outputs for approximately 98\% of scans after schema normalization. These consistent cases pass the automated validation stage and are directly accepted. The remaining 2\% of scans show discrepancies between the two model outputs and are automatically flagged for manual verification.

\smallskip
\noindent\textbf{Sources of extraction discrepancies.}
Most disagreements between the two models arise from heterogeneous report wording or incomplete report information. For example, CT contrast phase may be described using different expressions such as `portal venous phase'', `venous phase'', or abbreviated terminology. In other cases, demographic attributes such as age or sex may not appear explicitly in the report text, which can lead to missing or inconsistent extraction results. Additional discrepancies may arise from implicit protocol descriptions or institutional variations in reporting style.

\smallskip
\noindent\textbf{Manual verification and audit.}
All discrepant cases are reviewed by a radiology using the original report text. When the report provides clear evidence, the extracted attribute is corrected according to the predefined schema. If the report evidence remains insufficient or ambiguous, the attribute is assigned the label \texttt{Unknown} to avoid introducing uncertain metadata into subgroup analyses. In addition to reviewing discrepant cases, we randomly sample approximately 1,000 scans across cohorts for manual verification to estimate the overall reliability of the extraction pipeline. Age and sex extraction achieve high agreement with the source reports, as these attributes are typically stated explicitly in clinical documentation. CT phase extraction is slightly more challenging due to heterogeneous protocol terminology across institutions.

\begin{figure}[h]
	\centering
	\includegraphics[width=\linewidth]{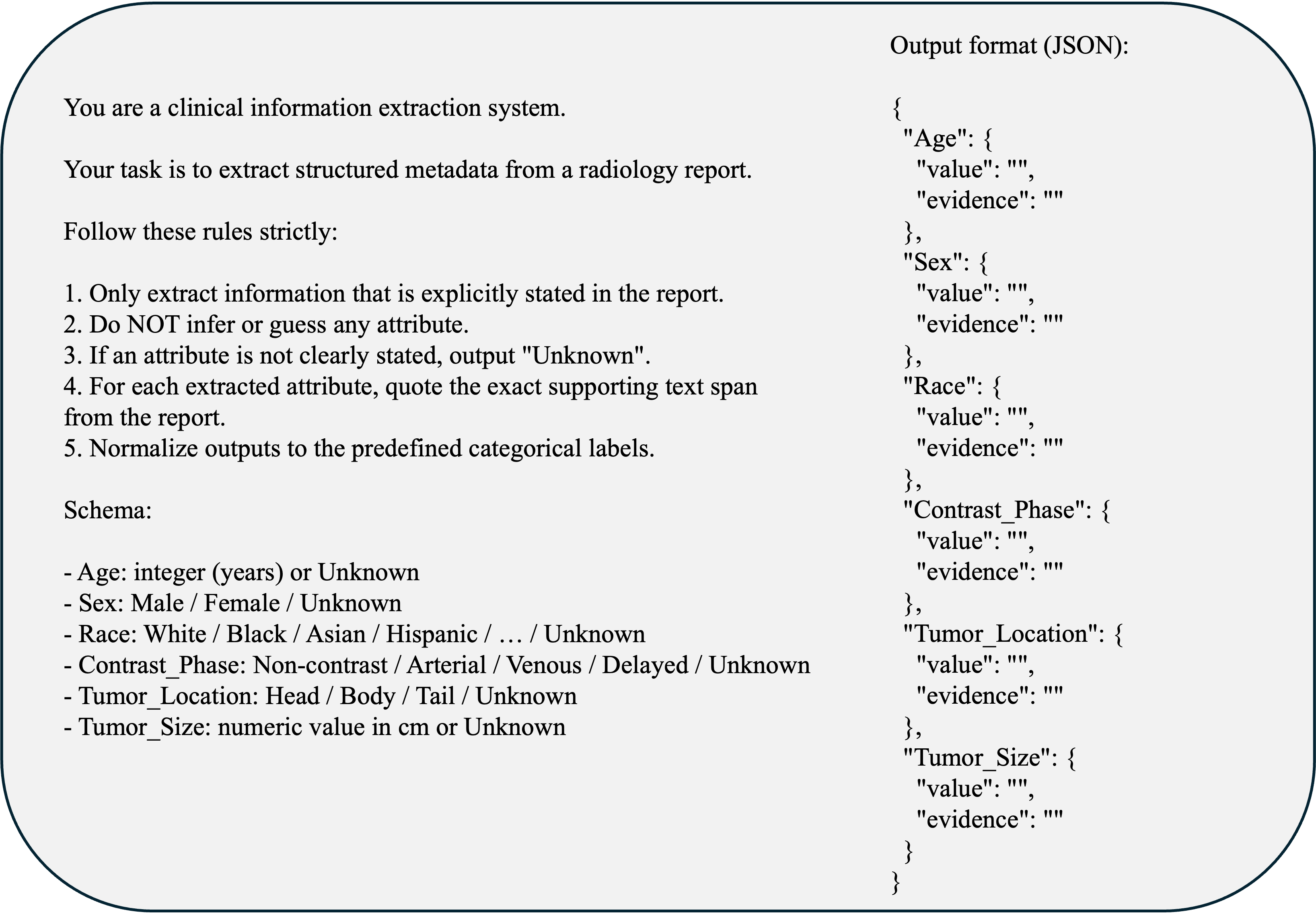}
    \caption{LLM Prompt for extracting structured attributes.}
\label{fig:prompt_metadata_extraction}
\end{figure}

\clearpage
\newpage
\section{Example Visualization in E-Coast Dataset}\label{sec:vis_E-Coast_appendix}

\begin{figure*}[ht]
    \centering
    \includegraphics[width=0.65\linewidth]{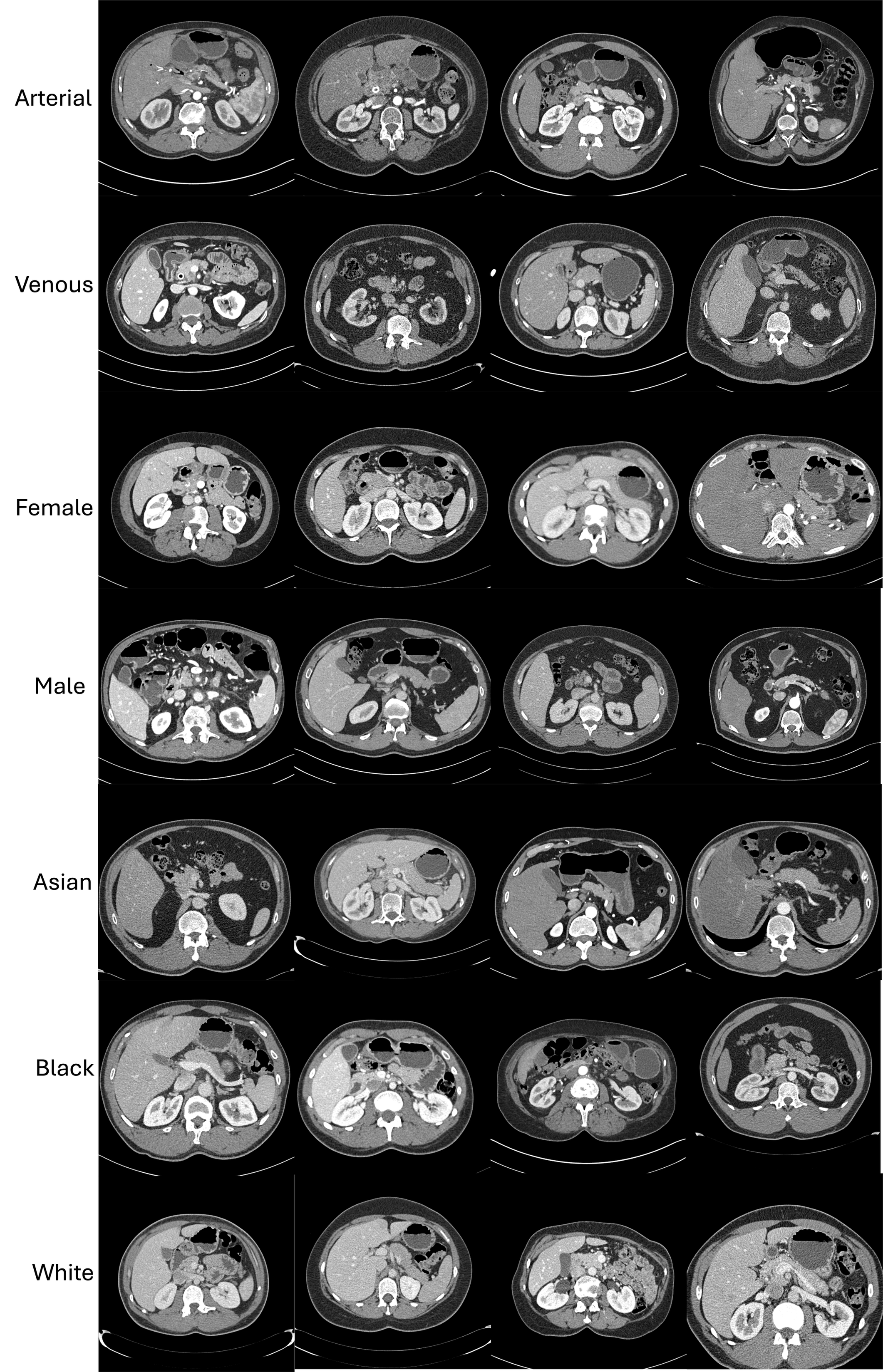}
    \caption{\textbf{Example CT slices across ct phase, sex and race subgroups in the E-Coast dataset.}
    We visualize representative subgroups across \emph{CT phase} (arterial, venous), 
    \emph{sex} (female, male), and \emph{race} (Asian, Black, White). 
    All images are preprocessed by clipping intensities to $[-175, 250]$ Hounsfield units
    and linearly normalizing them to the range $[0, 255]$.}
    \label{fig:supp_jhh_example}
\end{figure*}

\begin{figure*}[h]
	\centering
	\includegraphics[width=0.9\linewidth]{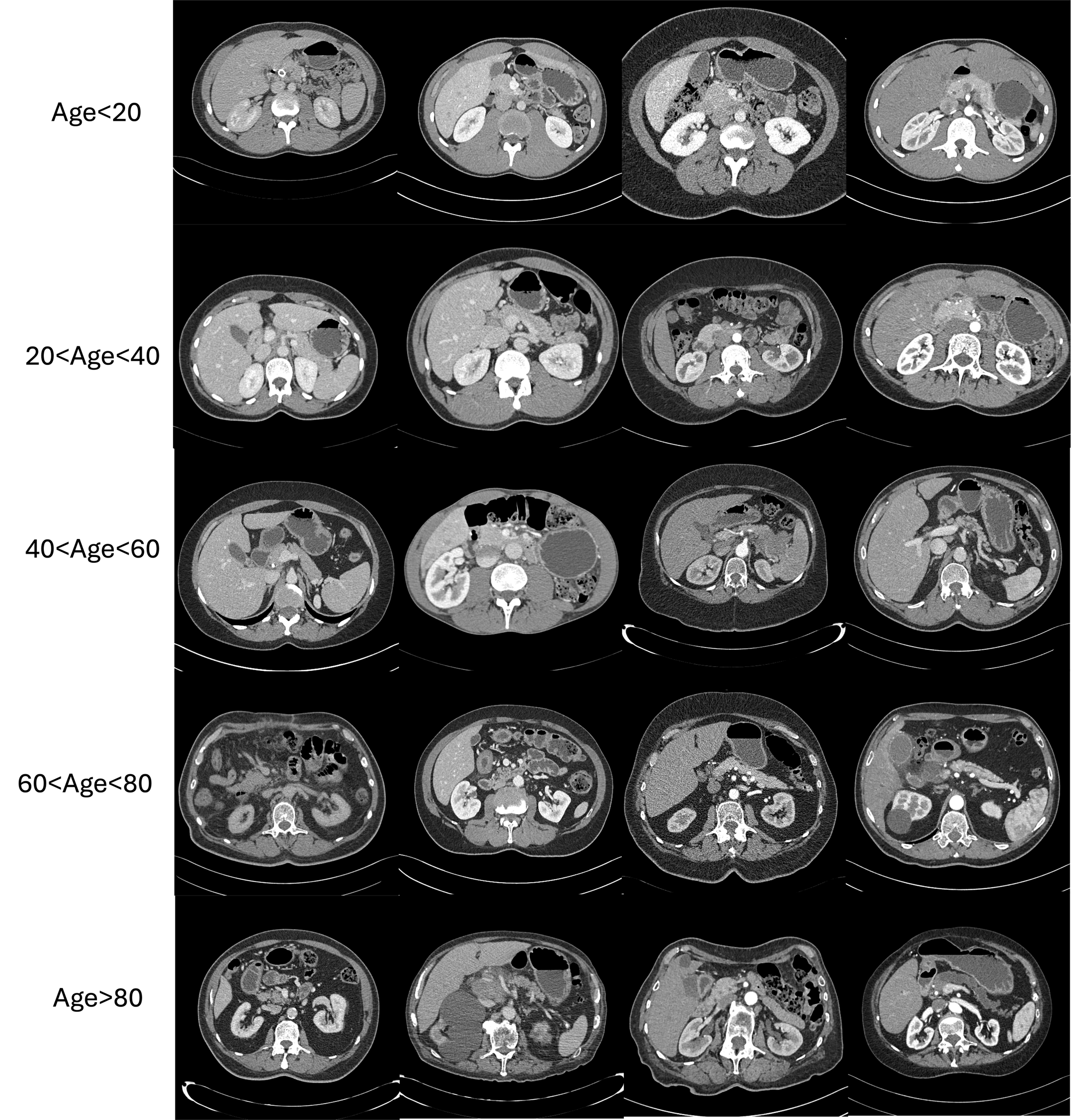}
    \caption{\textbf{Example CT slices across age subgroups in the E-Coast dataset.}
We visualize representative cases from five age groups (\emph{Age}<20, 20<\emph{Age}<40, 
40<\emph{Age}<60, 60<\emph{Age}<80, and \emph{Age}>80). 
All images are preprocessed by clipping intensities to $[-175, 250]$ Hounsfield units
and linearly normalizing them to the range $[0, 255]$.}

\label{fig:supp_jhh_example_age}
\end{figure*}

\clearpage
\newpage
\section{Example Visualization in Merlin Dataset}\label{sec:vis_merlin_appendix}

\begin{figure*}[h]
	\centering
	\includegraphics[width=0.7\linewidth]{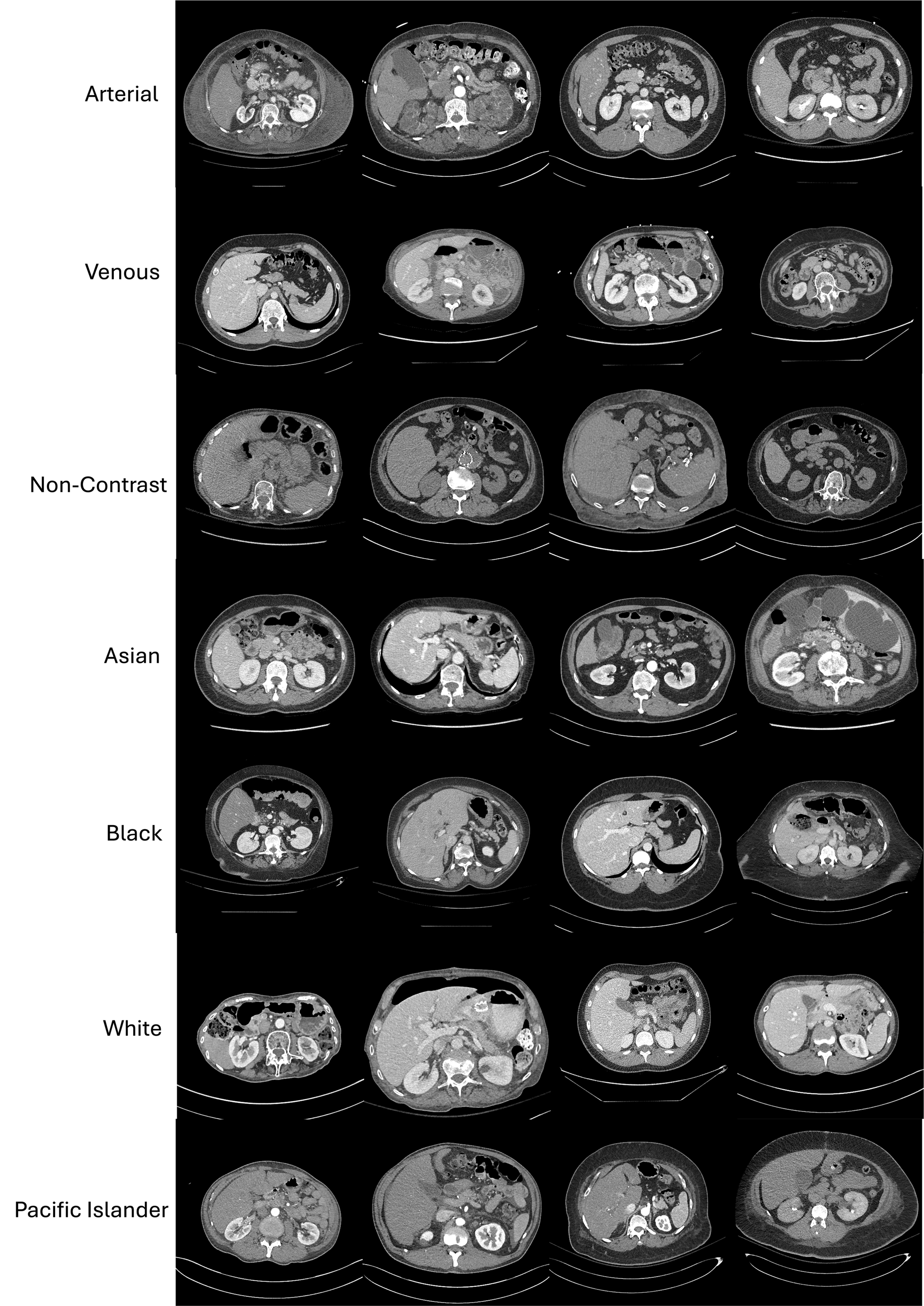}
    \caption{\textbf{Example CT slices across ct phase and race subgroups in the Merlin dataset.}
We visualize representative subgroups across \emph{CT phase} (arterial, venous, non-contrast) 
and \emph{race} (Asian, Black, White, Pacific Islander). 
All images are preprocessed by clipping intensities to $[-175, 250]$ Hounsfield units 
and linearly normalizing them to the range $[0, 255]$.}
\end{figure*}

\begin{figure*}[t]
	\centering
	\includegraphics[width=0.8\linewidth]{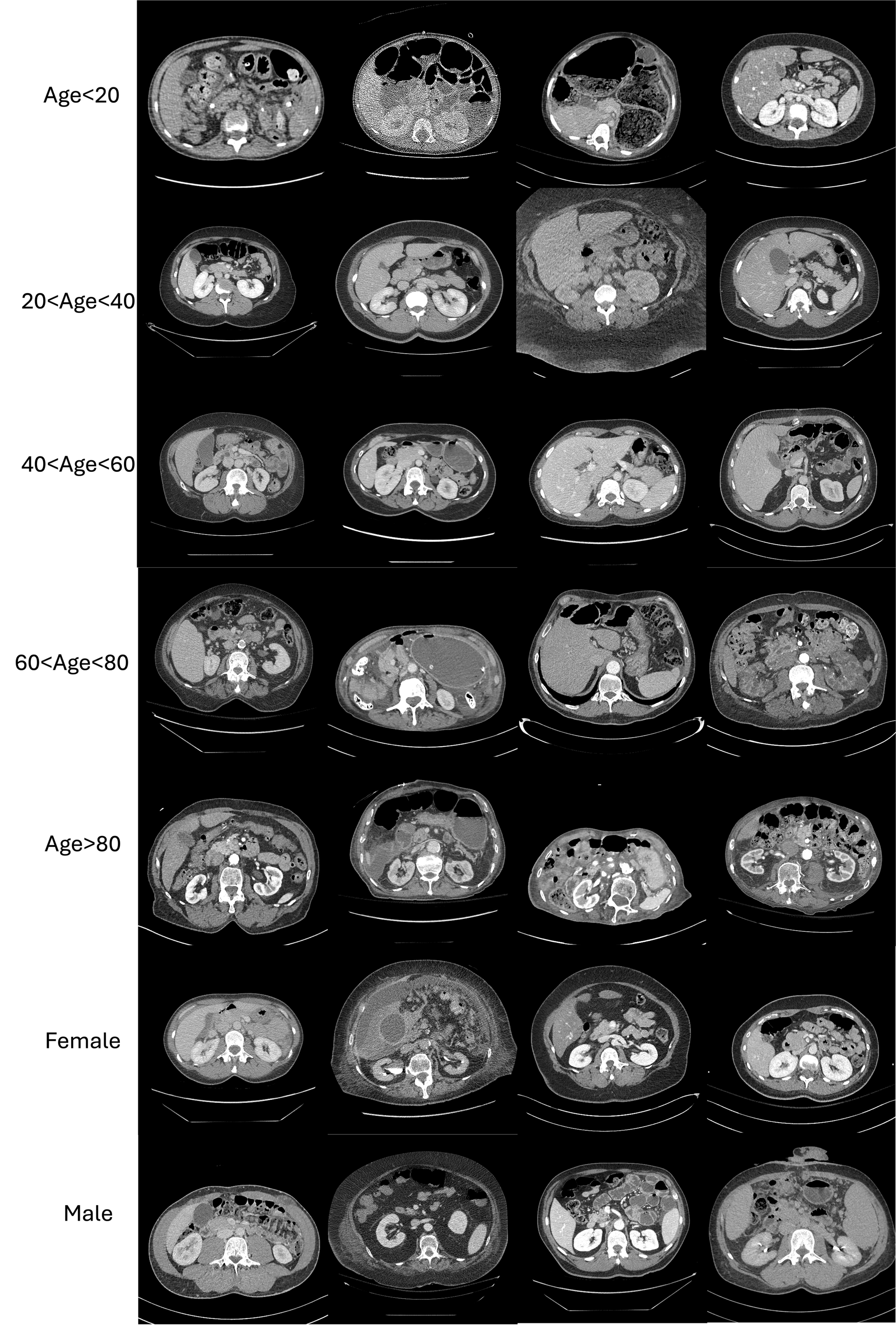}
    \caption{\textbf{Example CT slices across age and sex subgroups in the Merlin dataset.}
We visualize representative cases from five age groups (\emph{Age}<20, 20<\emph{Age}<40, 
40<\emph{Age}<60, 60<\emph{Age}<80, and \emph{Age}>80) as well as \emph{sex} subgroups 
(female, male). All images are preprocessed by clipping intensities to $[-175, 250]$ Hounsfield units
and linearly normalizing them to the range $[0, 255]$.}

\end{figure*}

\clearpage
\newpage
\section{Example Visualization in PanTS Dataset}\label{sec:vis_pants_appendix}

\begin{figure*}[h]
	\centering
	\includegraphics[width=0.9\linewidth]{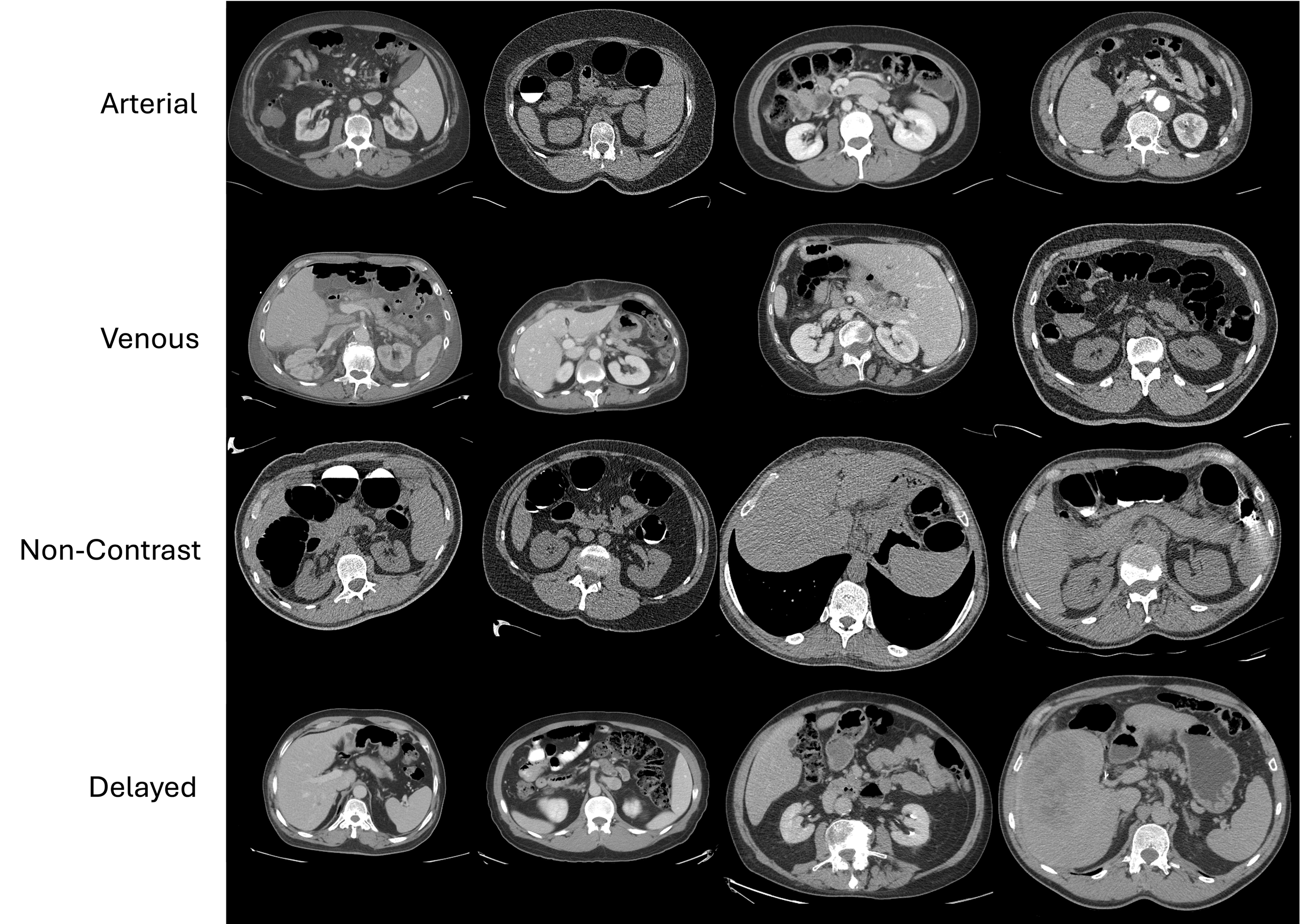}
    \caption{\textbf{Example CT slices across ct phase subgroups in the PanTS dataset.}
We visualize representative cases from four CT contrast phases, including \emph{arterial}, 
\emph{venous}, \emph{non-contrast}, and \emph{delayed} scans. 
All images are preprocessed by clipping intensities to $[-175, 250]$ Hounsfield units
and linearly normalizing them to the range $[0, 255]$.}
\end{figure*}

\begin{figure*}[h]
	\centering
	\includegraphics[width=0.9\linewidth]{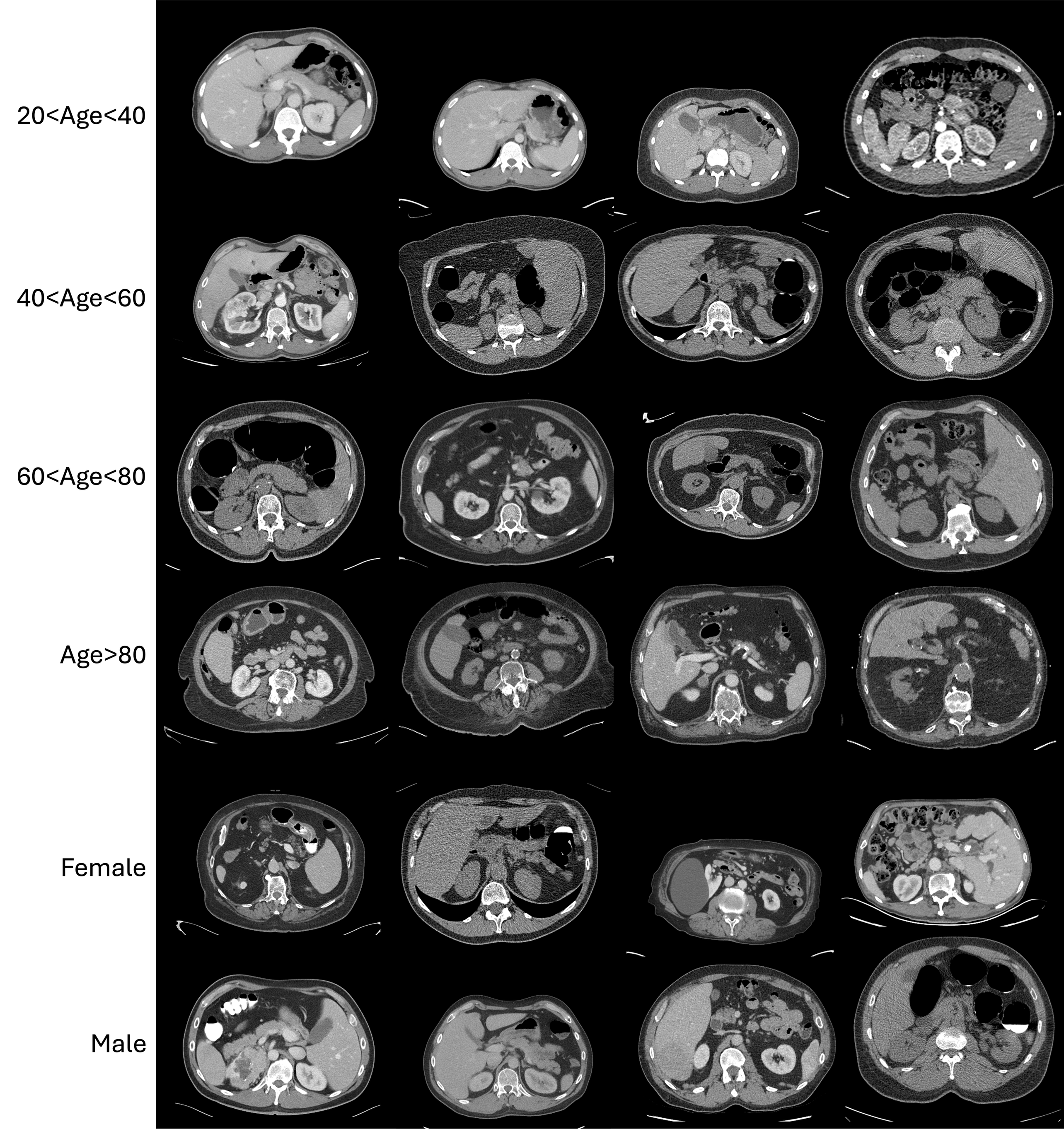}
\caption{\textbf{Example CT slices across age and sex subgroups in the PanTS dataset.}
We visualize representative cases from four age groups (20<\emph{Age}<40, 
40<\emph{Age}<60, 60<\emph{Age}<80, and \emph{Age}>80) as well as \emph{sex} subgroups 
(female, male). All images are preprocessed by clipping intensities to $[-175, 250]$ 
Hounsfield units and linearly normalizing them to the range $[0, 255]$.}

\end{figure*}

\clearpage
\newpage
\section{Example Visualization in N-California Dataset}\label{sec:vis_n-california_appendix}

\begin{figure*}[h]
	\centering
	\includegraphics[width=0.7\linewidth]{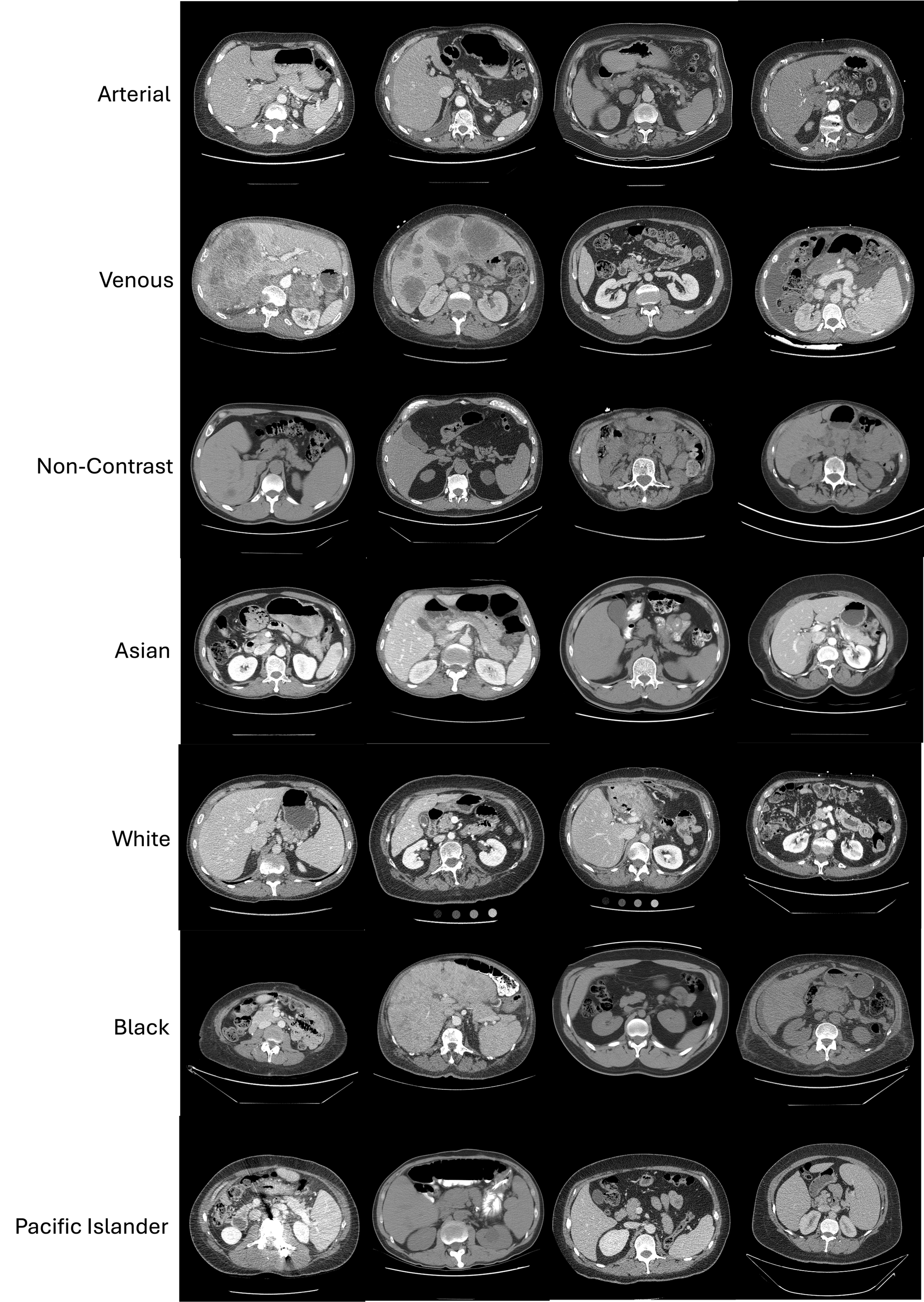}
    \caption{\textbf{Example CT slices across ct phase and race subgroups in the N-California dataset.}
We visualize representative cases from four CT phases (\emph{arterial}, \emph{venous}, and
\emph{non-contrast}) and from four race subgroups (\emph{Asian}, \emph{White}, 
\emph{Black}, and \emph{Pacific Islander}). All images are preprocessed by clipping intensities 
to $[-175, 250]$ Hounsfield units and linearly normalizing them to the range $[0, 255]$.}
\end{figure*}

\begin{figure*}[h]
	\centering
	\includegraphics[width=0.8\linewidth]{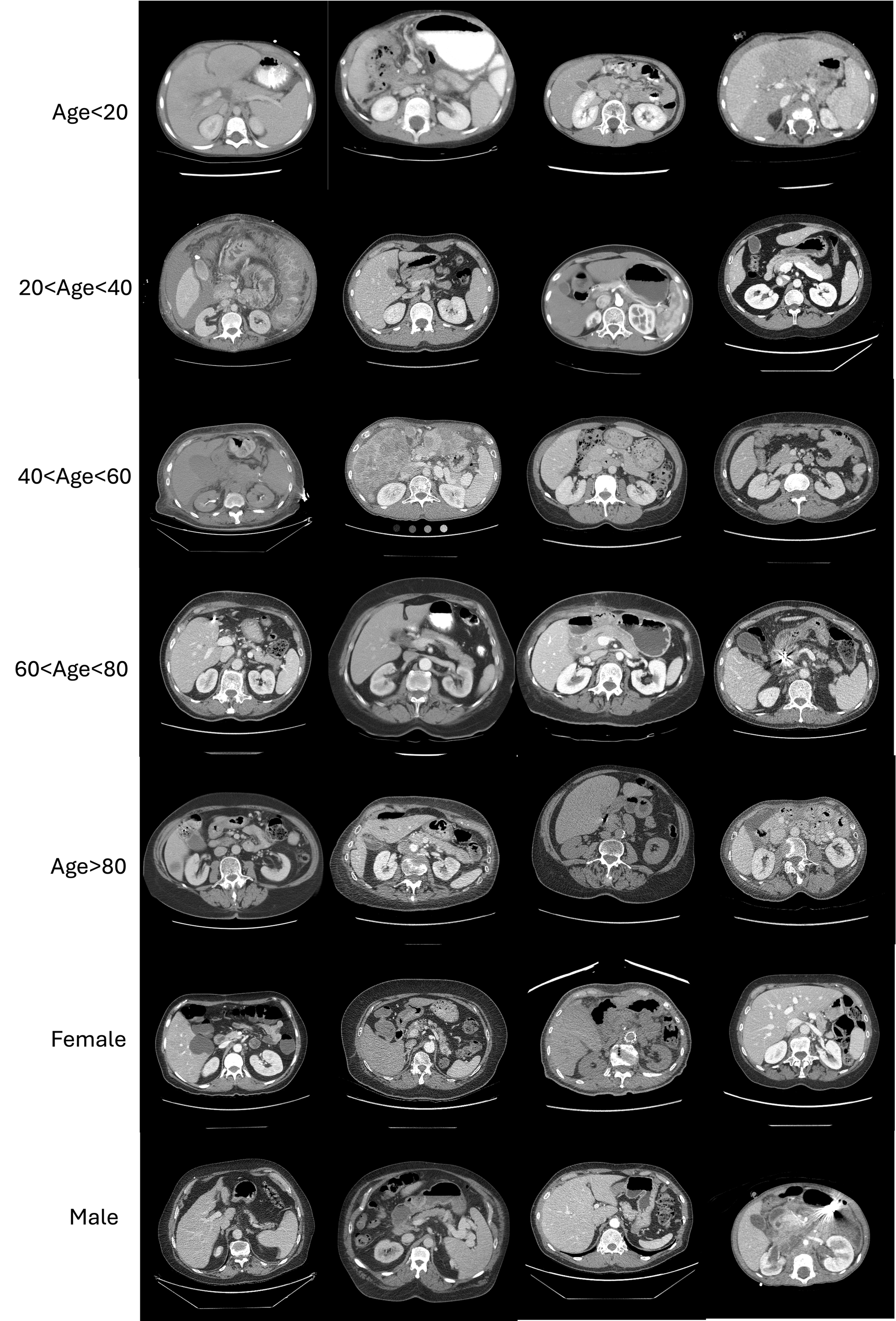}
    \caption{\textbf{Example CT slices across age and sex subgroups in the N-California dataset.}
We visualize representative cases from five age groups (\emph{Age}<20, 20<\emph{Age}<40, 
40<\emph{Age}<60, 60<\emph{Age}<80, and \emph{Age}>80) as well as \emph{sex} subgroups 
(female, male). All images are preprocessed by clipping intensities to $[-175, 250]$ 
Hounsfield units and linearly normalizing them to the range $[0, 255]$.}

\end{figure*}

\clearpage
\newpage
\section{Example Visualization in S-Europe Dataset}\label{sec:vis_s.europe_appendix}

\begin{figure*}[h]
	\centering
	\includegraphics[width=0.7\linewidth]{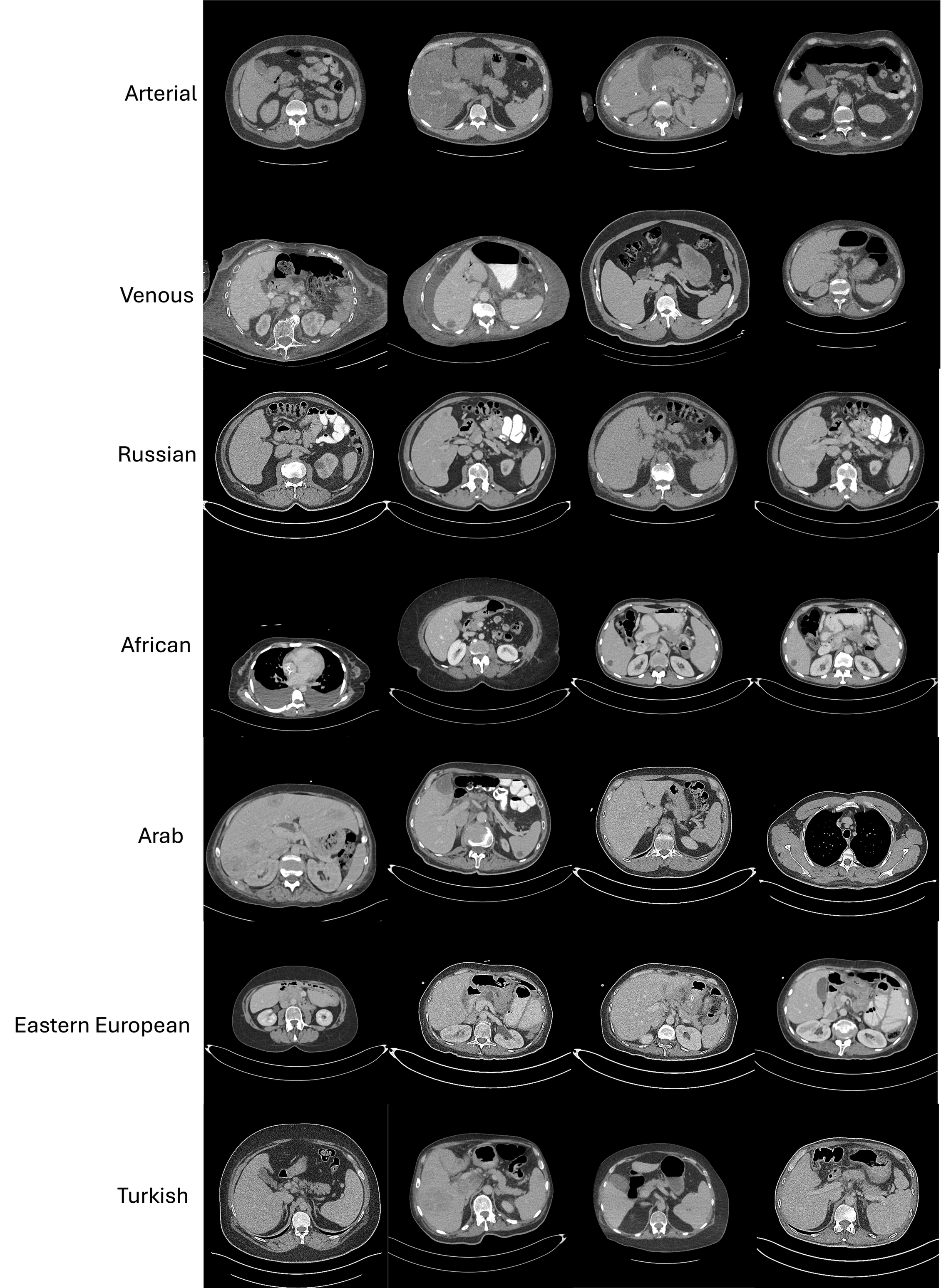}
    \caption{\textbf{Example CT slices across ct phase and race subgroups in the S.\ Europe dataset.}
We visualize representative cases from two CT contrast phases (\emph{arterial}, \emph{venous}) 
and from five race subgroups (\emph{Russian}, \emph{African}, \emph{Arab}, \emph{Eastern~European}, 
and \emph{Turkish}). All images are preprocessed by clipping intensities to $[-175, 250]$ 
Hounsfield units and linearly normalizing them to the range $[0, 255]$.}
\end{figure*}

\begin{figure*}[h]
	\centering
	\includegraphics[width=0.75\linewidth]{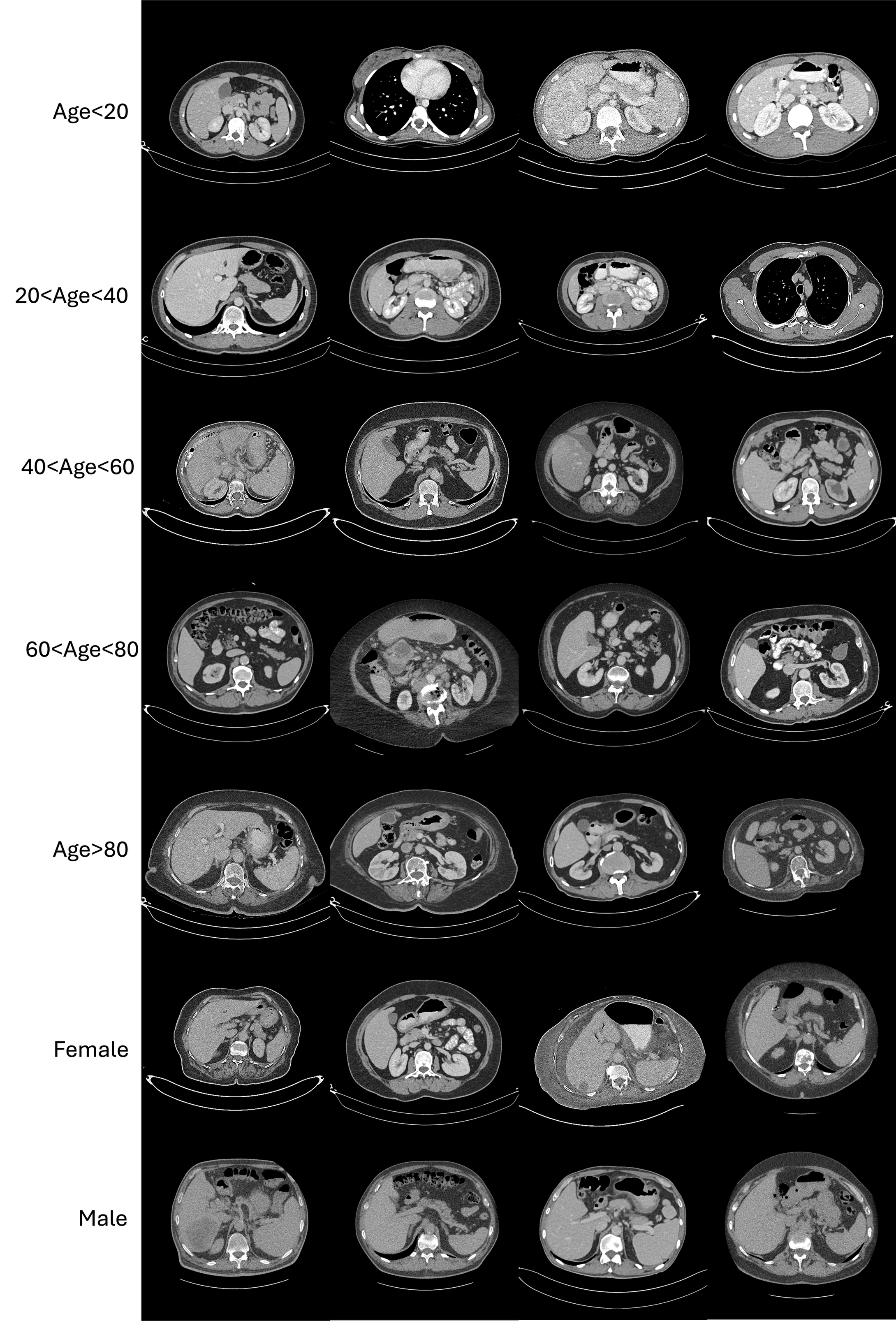}
    \caption{\textbf{Example CT slices across age and sex subgroups in the S.\ Europe dataset.}
We visualize representative cases from five age groups (\emph{Age}<20, 20<\emph{Age}<40,
40<\emph{Age}<60, 60<\emph{Age}<80, and \emph{Age}>80) as well as \emph{sex} subgroups
(female, male). All images are preprocessed by clipping intensities to $[-175, 250]$
Hounsfield units and linearly normalizing them to the range $[0, 255]$.}
\end{figure*}

\clearpage
\newpage
\section{Example Visualization in N-Europe Dataset}\label{sec:vis_n-europe_appendix}

\begin{figure*}[h]
	\centering
	\includegraphics[width=0.75\linewidth]{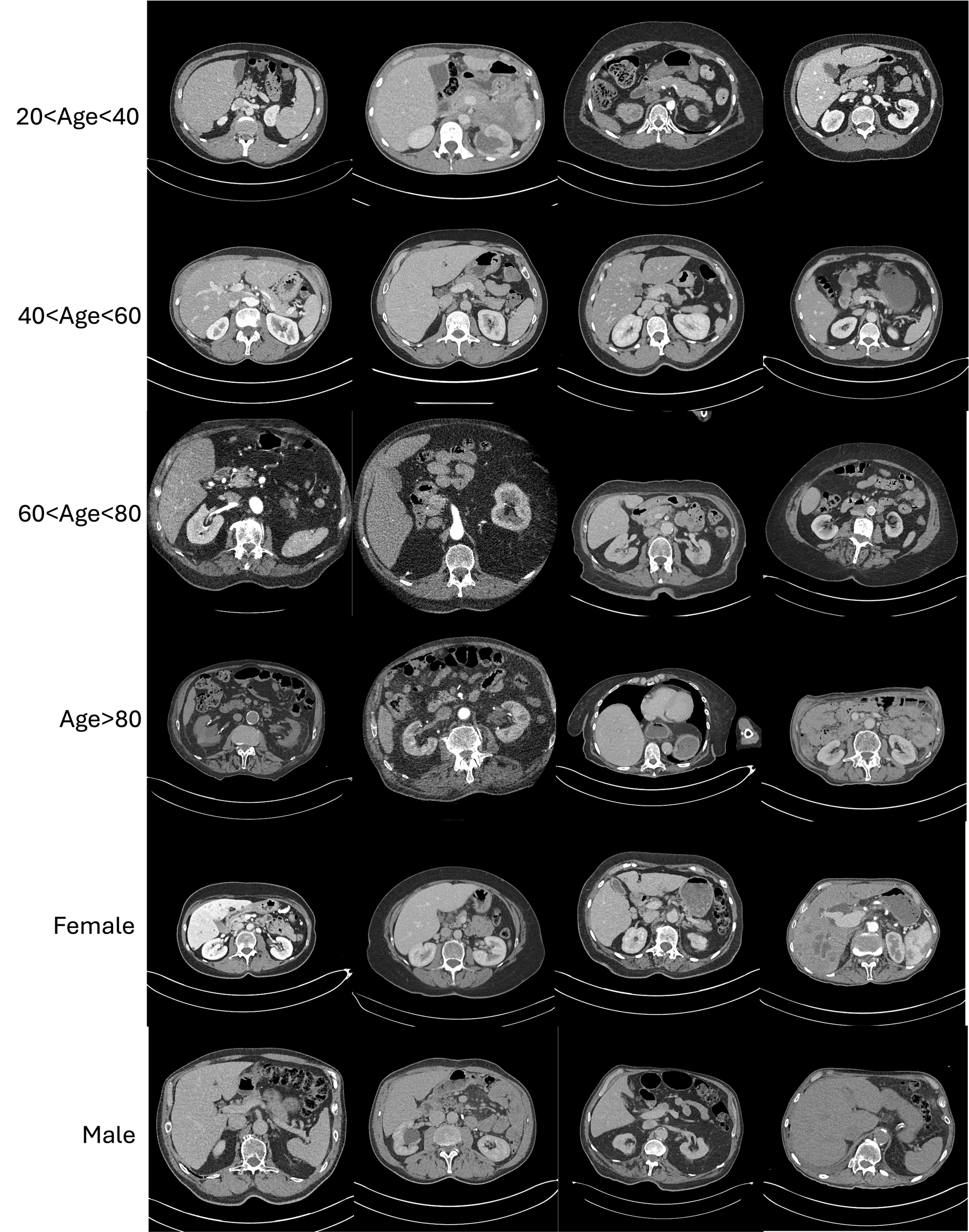}
    \caption{\textbf{Example CT slices across age and sex subgroups in the N-Europe dataset.}
We visualize representative cases from four age groups (20<\emph{Age}<40, 
40<\emph{Age}<60, 60<\emph{Age}<80, and \emph{Age}>80) as well as \emph{sex} subgroups 
(female, male). All images are preprocessed by clipping intensities to $[-175, 250]$ 
Hounsfield units and linearly normalizing them to the range $[0, 255]$.}
\end{figure*}

\begin{figure*}[h]
	\centering
	\includegraphics[width=0.85\linewidth]{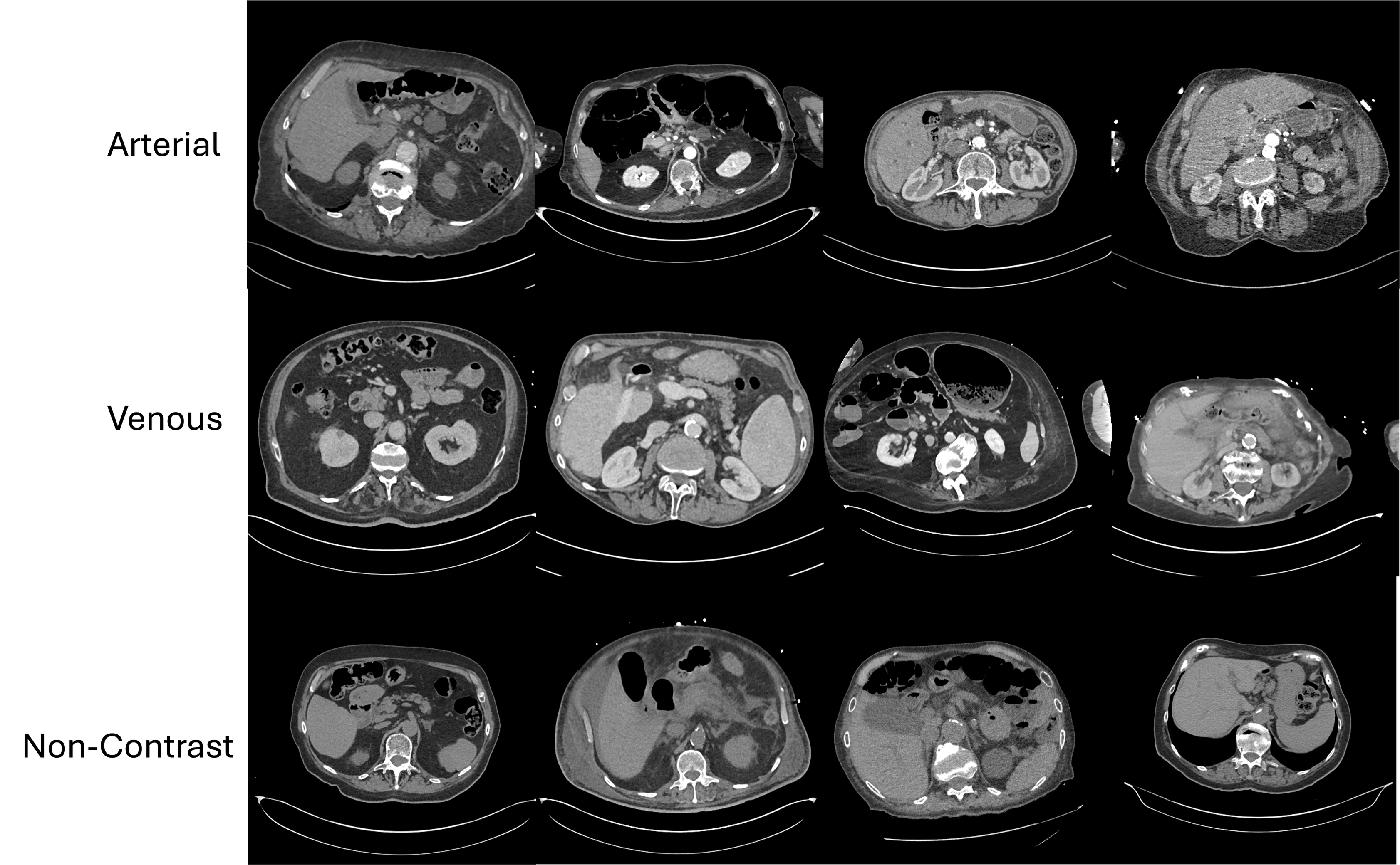}
    \caption{\textbf{Example CT slices across contrast-phase subgroups in the N-Europe dataset.}
We visualize representative cases from three CT contrast phases (\emph{arterial}, \emph{venous}, 
and \emph{non-contrast}). All images are preprocessed by clipping intensities to $[-175, 250]$ 
Hounsfield units and linearly normalizing them to the range $[0, 255]$.}
\end{figure*}

\clearpage
\section{Description of Evaluated Models}\label{sec:AI_Description_appendix}
\ourbench\ evaluates a curated set of architectures that are widely used in medical image segmentation and appear consistently
across our benchmark tables. These models represent three major families—classical CNNs, Transformer-based encoders,
and automated nnU-Net variants—and collectively form a strong and diverse baseline suite for assessing robustness across
demographic and protocol subgroups.

\textbf{U-Net.}
The U-Net \cite{ronneberger2015u} is a fully convolutional encoder–decoder network with long-range skip connections.
The encoder consists of repeated convolution–nonlinearity–pooling blocks that progressively downsample the feature maps and
capture high-level semantic context. The decoder mirrors this structure with upsampling layers that recover spatial resolution,
and each decoder stage concatenates encoder features at the corresponding scale via skip connections to restore fine-grained
local detail. This design provides a strong inductive bias for dense prediction in 2D and 3D medical imaging, balancing context
and localization. Despite its simplicity and relatively small parameter count, U-Net remains the foundational architecture in
biomedical segmentation and serves in BenchX as a canonical, low-complexity CNN baseline.

\textbf{UNETR.}
UNETR \cite{hatamizadeh2022unetr} is a 3D transformer-based segmentation backbone that replaces the convolutional encoder of a U-Net with a Vision Transformer (ViT). Input volumes are partitioned into non-overlapping 3D patches and encoded through a multi-layer ViT, which captures long-range dependencies and global contextual relationships. The encoded features are then fused with a convolutional decoder through skip connections at multiple resolutions, forming a hybrid ViT–CNN architecture. This design enables UNETR to capture global semantics while preserving spatial precision during decoding. In BenchX, UNETR serves as a representative pure-transformer encoder baseline, complementing Swin UNETR and providing insight into how full-attention ViTs behave under distribution shifts and subgroup imbalances in heterogeneous CT cohorts.

\textbf{Swin UNETR-Tiny / Small / Base.}
The Swin UNETR family \cite{tang2022self} integrates Swin Transformer blocks into a 3D U-shaped encoder–decoder.
Input volumes are partitioned into non-overlapping 3D patches that are embedded and processed by a hierarchical Swin
Transformer encoder with shifted-window self-attention, enabling efficient modeling of both local and global dependencies.
Decoder stages fuse multi-resolution Transformer features with convolutional upsampling, yielding a hybrid representation
that combines global context with precise spatial detail. BenchX includes three capacity levels—Tiny, Small, and Base—which
share the same design but differ in depth and width. These variants allow us to systematically study the effect of model scale
on subgroup robustness and to compare Transformer-based volumetric models against CNN-style baselines under the same
training protocol.

\textbf{Universal Model a/b.}
The Universal Models are lightweight 3D segmentation backbones implemented within MONAI and trained using
carefully curated medical priors and regularization. Architecturally, they follow a compact encoder–decoder design with
standard convolutional blocks and limited depth, targeting low-memory and low-latency deployment scenarios. Compared
with heavier CNN or Transformer models, they trade representational capacity for efficiency. In BenchX, Universal Model a
and b provide a controlled pair of “small-model” baselines, enabling us to assess how capacity-constrained networks behave
under distribution shifts across age, sex, race, and CT phases, and to quantify the robustness gap between lightweight and
high-capacity architectures.

\textbf{nnU-Net (Classic).}
nnU-Net \cite{isensee2021nnu} is a self-configuring framework that automatically designs and trains
segmentation pipelines based on dataset-specific statistics. For each new task, it derives image resampling strategies, intensity
normalization, patch and batch sizes, network depth, data augmentation policies, and post-processing rules through a set of
heuristics validated across many benchmarks. The classic nnU-Net architecture is a 3D U-Net–like encoder–decoder with
plain convolutional blocks and instance normalization, trained under a standardized optimization protocol. As a fully automated
pipeline, nnU-Net (Classic) in BenchX serves as a strong, reproducible reference: performance differences across subgroups
can be attributed primarily to data and model behavior rather than manual hyperparameter tuning.

\textbf{ResEnc.}
ResEnc \cite{isensee2024nnu} is the latest default backbone in nnU-Net. It replaces the plain encoder
blocks of the classic model with residual blocks that include identity shortcuts, improving gradient flow and representational
capacity while keeping the decoder lightweight. The architecture maintains the same overall U-shaped topology but increases
feature expressiveness in deeper layers, which is especially beneficial for complex multi-organ and tumor patterns. In BenchX,
ResEnc represents a “modernized” nnU-Net baseline that preserves the auto-configuration benefits while providing a stronger
backbone; comparing ResEnc and Classic nnU-Net highlights how architectural refinements affect robustness across rare and
underrepresented subgroups.

\textbf{STU-Net-B.}
STU-Net \cite{huang2023stu} is a family of scalable U-Net variants designed to study the effect of depth and width
scaling in medical segmentation. It introduces refined residual convolutional blocks and weight-free interpolation operations
to improve transferability across tasks and resolutions. The B variant (STU-Net-B) strikes a balance
between capacity and computational cost: it is substantially larger than a standard U-Net, yet more tractable than the largest
STU-Net configurations. BenchX uses STU-Net-B as a high-capacity CNN baseline, enabling comparison between hand-scaled
architectures (STU-Net-B, MedNeXt) and automatically configured ones (nnU-Net, ResEnc) under the same dataset and
subgroup evaluation protocol.

\textbf{MedNeXt.}
MedNeXt \cite{roy2023mednext} adapts ConvNeXt-style design principles to 3D medical image segmentation. Its basic
block employs large-kernel depthwise convolutions, pointwise convolutions, and inverted bottlenecks, coupled with residual
connections and layer normalization. This design combines the strong inductive bias of convolutions with Transformer-like
scaling behavior, facilitating training of deep and wide networks that remain stable on high-resolution volumes. In BenchX,
MedNeXt serves as a state-of-the-art CNN-style backbone that competes directly with Swin UNETR and nnU-Net variants.
Comparing MedNeXt with U-Net and STU-Net-B helps disentangle the contributions of modern architectural refinements,
network scale, and automated configuration to subgroup-level robustness.

\textbf{VSmTrans.}
VSmTrans \cite{liu2024vsmtrans} is a hybrid volumetric Transformer designed to overcome two key limitations of Vision Transformers in 3D medical segmentation: the high computational cost of capturing global context and the lack of strong local inductive bias needed for delineating fine anatomical boundaries. To address these challenges, VSmTrans introduces a Variable-Shape self-attention mechanism that groups multiple attention windows with flexible spatial shapes into a single module, enabling rapid expansion of the receptive field without incurring significant computational overhead. This design achieves an effective balance between global awareness and local detail preservation. In parallel, VSmTrans integrates a convolution-enhancing branch into each Transformer block. This hybrid design combines the representational capacity of self-attention with the locality and robustness provided by convolutions, while learnable weighting parameters adaptively control the contribution of each pathway during training. In BenchX, VSmTrans serves as a high-capacity Transformer baseline that captures fine-grained tumor details while remaining stable across demographic and protocol subgroups.

\textbf{Summary.}
The architectures included in BenchX were deliberately chosen to cover the principal design paradigms that define modern
3D medical image segmentation. Together, they span (i) classical convolutional encoder–decoders (U-Net, Universal Models),
(ii) scalable and high-capacity convolutional architectures (STU-Net-B, MedNeXt), (iii) Transformer-based volumetric models
that capture global context (UNETR, Swin UNETR-Tiny/Small/Base, VSmTrans), and (iv) fully automated self-configuring
systems (nnU-Net and its modern ResEnc variant). These categories represent the methodological foundations that have
shaped the field over the past decade, ensuring that BenchX provides broad and representative coverage of architectural
philosophies, inductive biases, and model capacities.

Crucially, this curated baseline suite is intentionally balanced: it includes lightweight models, high-capacity CNNs,
Transformer encoders of varying scales, and algorithmically selected nnU-Net pipelines. This diversity allows BenchX to
isolate how different design principles—local inductive bias, global receptive field, scaling behavior, or automated configuration—
impact robustness across demographic subgroups and imaging protocols.

Given the substantial data scale and heterogeneity of BenchX—over 85k CT volumes spanning multiple hospitals,
demographics, and contrast phases—training every emerging architecture is computationally impractical. We therefore focus
on well-established, community-adopted models that provide reproducible, interpretable, and fair reference points for
subgroup-level evaluation. Future iterations of BenchX will expand this set to include more architectures such as
foundation models, diffusion-based segmenters, and hybrid CNN–Transformer designs as computational resources permit.

\newpage
\section{Description of AI Frameworks}
\label{sec:framework_description_appendix}

\noindent\textbf{nnU-Net} \cite{isensee2021nnu} is a framework for automatically configuring AI-based semantic segmentation pipelines. Given a new segmentation dataset, it analyzes dataset-specific characteristics—such as voxel spacing, image geometry, and class imbalance—and automatically derives preprocessing steps, network depth, patch size, data augmentation, and training hyperparameters. Despite its first release dating back to 2019 and its use of a classical U-Net \cite{ronneberger2015u}, nnU-Net has remained competitive and continues to produce state-of-the-art results across diverse medical imaging benchmarks. Its lasting impact stems from demonstrating that robust pipeline configuration and validation often matter more than architectural novelty.

Because nnU-Net standardizes the entire segmentation workflow, it has become the reference framework for fair and reproducible benchmarking. The community widely extends it \cite{ye2023uniseg,huang2023stu,roy2023mednext}, making it one of the most stable baselines available for medical imaging research. A recent update \cite{isensee2024nnu} further provides modernized presets—such as residual encoder variants tuned for different VRAM budgets—ensuring consistent performance across hardware configurations.

\textbf{Why we use it in BenchX.}
For BenchX, nnU-Net serves as an essential baseline because it eliminates human-driven tuning and enforces a fully automated, reproducible pipeline. This ensures that performance differences across age, sex, race, and imaging phases arise from data- and model-dependent factors rather than researcher-dependent hyperparameter choices. nnU-Net therefore provides a stable reference point for measuring robustness, fairness, and subgroup generalization under the heterogeneous data distributions present in \ourbench.

\noindent\textbf{MONAI} (Medical Open Network for AI) \cite{cardoso2022monai} is an open-source, PyTorch-based framework developed to accelerate AI research and deployment in healthcare. It provides modular components covering the entire development lifecycle, including high-throughput data loaders for 3D volumes, standardized preprocessing and augmentation transforms, sliding-window inference utilities, and a suite of widely used architectures for segmentation, classification, and registration. MONAI also supports emerging paradigms such as self-supervised learning, federated learning, and reproducible experiment management.

Designed with extensibility and stability in mind, MONAI has become the foundation for many production-grade medical AI pipelines. Its unified interface and optimized GPU primitives make it particularly well suited for large-scale benchmarking and model comparison, where consistency and fairness across methods are critical.

\textbf{Why we use it in BenchX.}
BenchX includes MONAI-based implementations to ensure that multiple architectures can be trained under a unified, reproducible infrastructure. MONAI minimizes engineering variance by providing standardized transforms, dataloaders, and inference utilities, enabling us to compare models in a controlled and fair setting. The framework’s robustness, strong community support, and compatibility with clinical deployment pipelines also make it an ideal platform for evaluating real-world generalizability across heterogeneous subgroups in \ourbench.

\newpage
\section{Implementation and Configuration Details}\label{sec:implementation_details_appendix}
\subsection{Evaluated Models}
In \ourbench, all baseline models are trained and evaluated on AbdomenAtlas~2.0~\cite{chen2025scaling}, currently the largest publicly available abdominal tumor CT dataset. It contains 10,134 contrast-enhanced CT scans with 13,223 voxel-level tumor annotations across six organs (pancreas, liver, kidney, colon, esophagus, and uterus), along with 6,511 control scans. All annotations are manually curated by 23 expert radiologists, making the dataset several orders of magnitude larger than any existing public tumor dataset.

For fair and consistent benchmark comparisons, we extract from AbdomenAtlas~2.0 all cases containing pancreatic tumors and additionally sample 1,000 healthy controls to construct the training subset used for all baselines in \ourbench. Model-specific training hyper-parameters strictly follow the recommended configurations in Touchstone~1.0~\cite{bassi2024touchstone}, as these settings have been empirically validated to provide strong and stable performance across diverse 3D medical segmentation tasks.

The complete training configurations—covering iteration budgets, GPU requirements, optimizer choices, learning rate policies, patch sizes, and loss functions—are summarized in \tableautorefname~\ref{tab:supp_training_details} and \tableautorefname~\ref{tab:supp_hyper_param}.

\begin{table*}[h]
\caption{\textbf{Training configuration of all baselines used in \ourbench.}}
\centering
\label{tab:supp_training_details}
\scriptsize
\begin{tabular}{p{0.2\linewidth}p{0.1\linewidth}p{0.1\linewidth}p{0.08\linewidth}p{0.12\linewidth}p{0.10\linewidth}p{0.13\linewidth}}
\toprule
architecture & pre-trained & iterations$^\dagger$ & hours & GPU$^\ddagger$ & GPU memory & hyper-parameter \\ 
\midrule
U-Net & No & 250K & 7.5 & 1$\times$A6000 & 7 GB & Pre-defined \\
UNETR & No & 250K & 24 & 8$\times$V100 & 12 GB & Pre-defined \\
SwinUNETR-Tiny & Yes & 250K & 24 & 8$\times$V100 & 32 GB & Pre-defined \\
SwinUNETR-Small & Yes & 250K & 28 & 4$\times$A6000 & 24 GB & Pre-defined \\
SwinUNETR-Base & Yes & 250K & 30 & 4$\times$A6000 & 32 GB & Pre-defined \\
Universal Model$^a$ & No & 200K & 18 & 1$\times$A6000 & 10 GB & Pre-defined \\
Universal Model$^b$ & No & 200K & 18 & 1$\times$A6000 & 10 GB & Pre-defined \\
nnU-Net & No & 250K & 28 & 1$\times$A6000 & 24 GB & Auto-searched \\
ResEnc & No & 250K & 28 & 1$\times$A6000 & 24 GB & Auto-searched  \\
STU-Net-B & No & 500K & 30 & 1$\times$A6000 & 8.8 GB & Self-configuring \\
MedNeXt & No & 250K & 67 & 4$\times$A6000 & 17.6 GB & Self-configuring \\
VSmTrans & No & 300K & 48 & 4$\times$A6000 & 20 GB & Self-configuring \\
\bottomrule
\end{tabular}

\begin{tablenotes}
\item $^\dagger$1 iteration equals one batch update, not a full epoch.
\item $^\ddagger$GPU: number of GPUs $\times$ GPU model.
\end{tablenotes}
\end{table*}

\begin{table*}[h]
\caption{\textbf{Additional hyper-parameters for baselines in \ourbench.}}
\centering
\scriptsize
\setlength{\tabcolsep}{5pt}
\begin{tabular}{lcccccc}
\toprule
architecture & patch size & batch & optimizer & lr & loss & WD \\
\midrule
U-Net & [96,96,96] & 2 & SGD & 1e-2 (Poly) & Dice, CE & 3e-5 \\
UNETR & [96,96,96] & 2 & AdamW & 1e-3 (cosine) & Dice, CE & 1e-5 \\
SwinUNETR-Tiny & [96,96,96] & 2 & AdamW & 1e-3 (cosine) & Dice, CE & 1e-5 \\
SwinUNETR-Small & [96,96,96] & 2 & AdamW & 1e-3 (cosine) & Dice, CE & 1e-5 \\
SwinUNETR-Base & [96,96,96] & 2 & AdamW & 1e-3 (cosine) & Dice, CE & 1e-5 \\
Universal Model$^a$ & [96,96,96] & 2 & AdamW & 1e-4 & Dice, CE & 1e-5 \\
Universal Model$^b$ & [96,96,96] & 2 & AdamW & 1e-4 & Dice, CE & 1e-5 \\
nnU-Net & Auto & 2 & SGD & 1e-2 & Dice, CE & 3e-5 \\
ResEnc & Auto & 2 & SGD & 1e-2 & Dice, CE & 3e-5 \\
STU-Net-B & [96,96,96] & 2 & SGD & 1e-2 (Poly) & Dice, CE & 3e-5 \\
MedNeXt & [128,128,128] & 8 & AdamW & 1e-3 & Dice, CE & 3e-5 \\
VSmTrans & [128,128,128] & 4 & AdamW & 6e-4 & Dice, CE & 5e-2 \\
\bottomrule
\end{tabular}
\label{tab:supp_hyper_param}
\end{table*}

\subsection{SynthX}

\clearpage
\newpage
\section{Evaluation Metrics}\label{sec:metrics_appendix}
Reliable subgroup analysis requires metrics that capture complementary aspects of model behavior under demographic and protocol variability. Each metric highlights a different failure mode—such as missed small tumors, boundary inaccuracies, false alarms, or class-imbalance sensitivity—and together they form a comprehensive basis for assessing fairness and robustness. In \ourbench, we adopt a suite of widely used and clinically meaningful metrics: Sensitivity and Specificity for decomposing error patterns, F1 score for balancing recall and precision under skewed prevalence, and Balanced AUC for evaluating discriminative ability in a class-imbalance–aware manner. This combination provides an equitable and rigorous foundation for comparing model performance across diverse subgroups.

\noindent\textbf{Sensitivity \& Specificity.} Sensitivity (also known as recall or true positive rate) quantifies the proportion of actual positives correctly identified, while Specificity measures the proportion of actual negatives correctly identified. They are defined as:
\begin{equation}\footnotesize
\text{Sensitivity} = \frac{TP}{TP + FN}, \quad \text{Specificity} = \frac{TN}{TN + FP},
\end{equation}
where $TP$, $TN$, $FP$, and $FN$ are the numbers of true positives, true negatives, false positives, and false negatives, respectively. High sensitivity is critical for minimizing missed detections, whereas high specificity is important to reduce false alarms.

\noindent\textbf{F1 score.} F1 score provides a balanced measure of segmentation or detection performance by combining Sensitivity and Precision into a single harmonic-mean statistic. It is particularly informative when evaluating tasks with class imbalance, where a model may achieve high accuracy by favoring the majority class. Formally, F1 score is defined as:

\begin{equation}\footnotesize
\text{F1} = \frac{2 \cdot \text{Precision} \cdot \text{Sensitivity}}
{\text{Precision} + \text{Sensitivity}}
= \frac{2TP}{2TP + FP + FN},
\end{equation}

where $TP$, $FP$, and $FN$ denote true positives, false positives, and false negatives, respectively. A high F1 score requires the model to achieve both strong recall (low false negatives) and strong precision (low false positives), making it well suited for tumor detection tasks where missing small lesions or generating excessive false alarms can be clinically consequential. Compared with single-aspect metrics such as Sensitivity or Specificity, the F1 score offers a more holistic assessment of model reliability under distribution shifts and subgroup disparities, which is essential for evaluating robustness in \ourbench.

\noindent\textbf{AUC.} Balanced Area Under the ROC Curve evaluates a model’s ability to distinguish positive from negative cases while compensating for class imbalance. Standard AUC can be disproportionately influenced by the majority class, especially in medical imaging tasks where tumor prevalence varies widely across cohorts and subgroups. AUC mitigates this by computing ROC curves under a balanced sampling strategy, ensuring equal contribution from positive and negative samples regardless of their natural frequency.

Formally, AUC is defined as the area under a ROC curve constructed by weighting or resampling positives and negatives such that both classes contribute equally to the true positive rate (TPR) and false positive rate (FPR). This yields a metric that better reflects separability independent of class prevalence, unlike conventional AUC which may appear high even when the model performs poorly on minority or rare cases.

Balanced AUC is particularly important in \ourbench, where demographic (age, sex, race) and protocol-specific (phase) subgroups exhibit significant shifts in tumor prevalence. In such settings, AUC provides a stable, fairness-aware assessment of subgroup robustness by isolating the model’s discriminative ability from prevalence-induced distortions. As a result, AUC complements threshold-dependent metrics such as F1 score and highlights subgroup disparities that would otherwise be underestimated by standard AUC.

\newpage
\section{Full Benchmark Results}\label{sec:full_benchmark_results}

\begin{table}[t]
    \centering
    \tiny
    \caption{\textbf{Validation on \textsl{E-Coast}.}
Architectures are grouped by framework. 
Subgroup analysis reveals substantial performance disparities across 
\colorbox{orange!20}{age}, with markedly lower F1 scores for younger patients 
(Age $\leq$20 and 20$<$Age$\leq$40). 
Detection performance across \colorbox{blue!20}{sex} shows minimal variation. 
For \colorbox{green!20}{race}, Black and White subgroups exhibit noticeably 
lower F1 scores compared with Asian patients. 
Differences across \colorbox{purple!20}{imaging phases} are modest, as the cohort 
contains only arterial and venous scans.}

    \begin{tabular}{p{0.26\linewidth}P{0.05\linewidth}P{0.05\linewidth}P{0.05\linewidth}P{0.05\linewidth}P{0.05\linewidth}P{0.05\linewidth}P{0.05\linewidth}P{0.05\linewidth}P{0.05\linewidth}P{0.05\linewidth}P{0.05\linewidth}P{0.05\linewidth}} 
    \toprule
    \multicolumn{1}{c}{\cellcolor{orange!20}\textbf{Age Groups}} & \multicolumn{4}{c}{\cellcolor{orange!20}Age $\leq 20$} & \multicolumn{4}{c}{\cellcolor{orange!20}20 \textless Age $\leq 40$} & \multicolumn{4}{c}{\cellcolor{orange!20}40 \textless Age $\leq 60$} \\
    \cmidrule(lr){2-5}\cmidrule(lr){6-9}\cmidrule(lr){10-13}
    Method & Sen. & Spe. & AUC & F1 & Sen. & Spe. & AUC & F1 & Sen. & Spe. & AUC & F1 \\
    \midrule
    U-Net~\cite{ronneberger2015u} & 100.0 & 45.5 & 72.8 & 25.0 & 62.5 & 65.0 & 63.8 & 41.0 & 70.2 & 62.2 & 66.2 & 63.9 \\
    Swin UNETR-Tiny~\cite{tang2022self} & 100.0 & 45.5 & 72.8 & 25.0 & 66.7 & 61.1 & 63.9 & 41.0 & 70.3 & 54.7 & 62.5 & 61.2 \\
    Swin UNETR-Small~\cite{tang2022self} & 100.0 & 54.5 & 77.3 & 28.6 & 66.7 & 75.5 & 71.1 & 50.0 & 69.3 & 68.2 & 68.8 & 65.7 \\
    Swin UNETR-Base~\cite{tang2022self} & 100.0 & 63.6 & 81.8 & 33.3 & 51.0 & 79.8 & 65.4 & 43.8 & 67.1 & 76.3 & 71.7 & 67.7 \\
    Universal Model$^a$~\cite{liu2023clip} & 0.0 & 63.6 & 31.8 & 0.0 & 55.2 & 62.9 & 59.1 & 36.0 & 65.1 & 59.5 & 62.3 & 59.7 \\
    Universal Model$^b$~\cite{liu2023clip} & 0.0 & 90.9 & 45.5 & 0.0 & 35.4 & 89.3 & 62.4 & 39.5 & 52.5 & 84.9 & 68.7 & 61.0 \\ \hline
    nnU-Net~\cite{isensee2021nnu} & 100.0 & 90.9 & 95.5 & 66.7 & 52.1 & 90.5 & 71.3 & 54.6 & 65.5 & 80.8 & 73.2 & 68.7 \\
    ResEnc~\cite{isensee2024nnu} & 50.0 & 100.0 & 75.0 & 66.7 & 30.2 & 100.0 & 65.1 & 46.4 & 33.5 & 99.5 & 66.5 & 49.9 \\
    UNETR~\cite{hatamizadeh2022unetr} & 100.0 & 50.0 & 75.0 & 26.7 & 66.7 & 49.1 & 57.9 & 35.6 & 72.9 & 42.4 & 57.7 & 58.7 \\
    STU-Net-B~\cite{huang2023stu} & 100.0 & 86.4 & 93.2 & 57.1 & 47.9 & 87.7 & 67.8 & 48.4 & 64.3 & 75.9 & 70.1 & 65.6 \\
    MedNeXt~\cite{roy2023mednext} & 0.0 & 100.0 & 50.0 & 0.0 & 41.7 & 99.7 & 70.7 & 58.4 & 49.9 & 98.9 & 74.4 & 65.9 \\
    VSmTrans~\cite{liu2024vsmtrans} & 100.0 & 59.1 & 79.6 & 30.8 & 61.5 & 85.4 & 73.5 & 55.7 & 75.6 & 79.0 & 77.3 & 74.4 \\
    \midrule

    \multicolumn{1}{c}{\cellcolor{blue!20}\textbf{Sex Groups}} & \multicolumn{4}{c}{\cellcolor{blue!20}Female} & \multicolumn{4}{c}{\cellcolor{blue!20}Male} & \multicolumn{4}{c}{\cellcolor{orange!20}60 \textless Age $\leq 80$} \\
    \cmidrule(lr){2-5}\cmidrule(lr){6-9}\cmidrule(lr){10-13}
    Method & Sen. & Spe. & AUC & F1 & Sen. & Spe. & AUC & F1 & Sen. & Spe. & AUC & F1 \\
    \midrule
    U-Net~\cite{ronneberger2015u} & 72.2 & 64.9 & 68.6 & 72.0 & 76.7 & 60.4 & 68.6 & 77.6 & 76.0 & 65.2 & 70.6 & 83.6 \\
    Swin UNETR-Tiny~\cite{tang2022self} & 72.0 & 60.5 & 66.3 & 70.7 & 82.5 & 51.0 & 66.8 & 79.1 & 80.7 & 58.9 & 69.8 & 86.0 \\
    Swin UNETR-Small~\cite{tang2022self} & 70.3 & 70.2 & 70.3 & 72.3 & 78.7 & 71.1 & 74.9 & 81.1 & 76.3 & 72.8 & 74.6 & 84.4 \\
    Swin UNETR-Base~\cite{tang2022self} & 68.3 & 81.7 & 75.0 & 74.6 & 78.0 & 70.4 & 74.2 & 80.5 & 76.2 & 78.1 & 77.2 & 84.7 \\
    Universal Model$^a$~\cite{liu2023clip} & 52.6 & 64.1 & 58.4 & 57.9 & 70.0 & 53.7 & 61.9 & 72.0 & 59.4 & 57.1 & 58.3 & 71.3 \\
    Universal Model$^b$~\cite{liu2023clip} & 42.3 & 87.8 & 65.1 & 55.6 & 58.5 & 82.9 & 70.7 & 69.8 & 49.1 & 83.9 & 66.5 & 64.7 \\ \hline
    nnU-Net~\cite{isensee2021nnu} & 70.7 & 82.7 & 76.7 & 76.5 & 74.0 & 82.8 & 78.4 & 80.8 & 75.6 & 76.8 & 76.2 & 84.2 \\
    ResEnc~\cite{isensee2024nnu} & 32.3 & 99.5 & 65.9 & 48.6 & 32.3 & 99.5 & 65.9 & 48.7 & 30.6 & 98.7 & 64.7 & 46.8 \\
    UNETR~\cite{hatamizadeh2022unetr} & 69.9 & 49.5 & 59.7 & 66.4 & 77.5 & 34.0 & 55.8 & 72.9 & 74.1 & 38.4 & 56.3 & 80.4 \\
    STU-Net-B~\cite{huang2023stu} & 69.4 & 79.3 & 74.4 & 74.6 & 73.1 & 77.8 & 75.5 & 79.0 & 75.1 & 75.5 & 75.3 & 83.8 \\
    MedNeXt~\cite{roy2023mednext} & 52.6 & 98.8 & 75.7 & 68.5 & 53.6 & 99.2 & 76.4 & 69.6 & 53.7 & 97.8 & 75.8 & 69.7 \\
    VSmTrans~\cite{liu2024vsmtrans} & 78.4 & 77.5 & 78.0 & 79.8 & 80.4 & 85.5 & 83.0 & 85.5 & 81.3 & 79.0 & 80.2 & 88.0 \\
    \midrule

    \multicolumn{1}{c}{\cellcolor{purple!20}\textbf{Imaging Phase Groups}} & \multicolumn{4}{c}{\cellcolor{purple!20}Arterial} & \multicolumn{4}{c}{\cellcolor{purple!20}Venous} & \multicolumn{4}{c}{\cellcolor{orange!20}Age \textgreater 80} \\
    \cmidrule(lr){2-5}\cmidrule(lr){6-9}\cmidrule(lr){10-13}
    Method & Sen. & Spe. & AUC & F1 & Sen. & Spe. & AUC & F1 & Sen. & Spe. & AUC & F1 \\
    \midrule
    U-Net~\cite{ronneberger2015u} & 72.0 & 52.0 & 62.0 & 73.9 & 70.6 & 74.4 & 72.5 & 77.2 & 86.4 & 100.0 & 93.2 & 92.7 \\
    Swin UNETR-Tiny~\cite{tang2022self} & 69.6 & 59.9 & 64.8 & 73.7 & 79.6 & 54.4 & 67.0 & 79.0 & 83.3 & 50.0 & 66.7 & 90.6 \\
    Swin UNETR-Small~\cite{tang2022self} & 67.9 & 67.3 & 67.6 & 74.0 & 75.2 & 73.7 & 74.5 & 80.0 & 84.6 & 50.0 & 67.3 & 91.3 \\
    Swin UNETR-Base~\cite{tang2022self} & 68.8 & 77.7 & 73.3 & 76.7 & 72.7 & 77.5 & 75.1 & 79.2 & 85.2 & 100.0 & 92.6 & 92.0 \\
    Universal Model$^a$~\cite{liu2023clip} & 57.9 & 62.3 & 60.1 & 65.8 & 64.3 & 59.2 & 61.8 & 69.9 & 61.1 & 100.0 & 80.6 & 75.9 \\
    Universal Model$^b$~\cite{liu2023clip} & 46.4 & 85.6 & 66.0 & 60.6 & 52.4 & 86.0 & 69.2 & 65.8 & 57.4 & 100.0 & 78.7 & 72.9 \\ \hline
    nnU-Net~\cite{isensee2021nnu} & 72.3 & 79.1 & 75.7 & 79.3 & 66.9 & 86.5 & 76.7 & 77.2 & 85.8 & 100.0 & 92.9 & 92.4 \\
    ResEnc~\cite{isensee2024nnu} & 28.0 & 99.6 & 63.8 & 43.6 & 36.6 & 99.3 & 68.0 & 53.4 & 41.4 & 100.0 & 70.7 & 58.5 \\
    UNETR~\cite{hatamizadeh2022unetr} & 68.7 & 47.2 & 58.0 & 70.8 & 75.7 & 40.9 & 58.3 & 74.2 & 76.5 & 50.0 & 63.3 & 86.4 \\
    STU-Net-B~\cite{huang2023stu} & 70.8 & 69.4 & 70.1 & 76.3 & 66.3 & 88.2 & 77.3 & 77.1 & 82.7 & 50.0 & 66.4 & 90.2 \\
    MedNeXt~\cite{roy2023mednext} & 49.1 & 99.2 & 74.2 & 65.7 & 55.5 & 98.7 & 77.1 & 71.1 & 69.8 & 100.0 & 84.9 & 82.2 \\
    VSmTrans~\cite{liu2024vsmtrans} & 77.7 & 79.8 & 78.8 & 82.9 & 77.2 & 81.2 & 79.2 & 82.9 & 90.7 & 100.0 & 95.4 & 95.2 \\
    \midrule

    \multicolumn{1}{c}{\cellcolor{green!20}\textbf{Race Groups}} & \multicolumn{4}{c}{\cellcolor{green!20}Asian} & \multicolumn{4}{c}{\cellcolor{green!20}Black} & \multicolumn{4}{c}{\cellcolor{green!20}White} \\
    \cmidrule(lr){2-5}\cmidrule(lr){6-9}\cmidrule(lr){10-13}
    Method & Sen. & Spe. & AUC & F1 & Sen. & Spe. & AUC & F1 & Sen. & Spe. & AUC & F1 \\
    \midrule
    U-Net~\cite{ronneberger2015u} & 100.0 & 57.1 & 78.6 & 36.4 & 75.0 & 64.7 & 69.9 & 30.0 & 78.6 & 64.0 & 71.3 & 30.4 \\
    Swin UNETR-Tiny~\cite{tang2022self} & 100.0 & 63.3 & 81.7 & 40.0 & 75.0 & 53.8 & 64.4 & 25.0 & 82.5 & 57.0 & 69.8 & 28.1 \\
    Swin UNETR-Small~\cite{tang2022self} & 83.3 & 75.5 & 79.4 & 43.5 & 70.0 & 65.8 & 67.9 & 28.9 & 84.9 & 71.5 & 78.2 & 37.5 \\
    Swin UNETR-Base~\cite{tang2022self} & 83.3 & 77.5 & 80.4 & 45.5 & 85.0 & 76.6 & 80.8 & 42.5 & 82.5 & 77.9 & 80.2 & 42.3 \\
    Universal Model$^a$~\cite{liu2023clip} & 66.7 & 81.6 & 74.2 & 42.1 & 60.0 & 63.0 & 61.5 & 24.0 & 66.7 & 59.5 & 63.1 & 24.4 \\
    Universal Model$^b$~\cite{liu2023clip} & 50.0 & 91.8 & 70.9 & 46.1 & 45.0 & 90.8 & 67.9 & 39.1 & 58.7 & 85.6 & 72.2 & 39.9 \\ \hline
    nnU-Net~\cite{isensee2021nnu} & 100.0 & 87.8 & 93.9 & 66.7 & 70.0 & 81.0 & 75.5 & 40.6 & 84.9 & 82.8 & 83.9 & 49.1 \\
    ResEnc~\cite{isensee2024nnu} & 50.0 & 100.0 & 75.0 & 66.7 & 30.0 & 100.0 & 65.0 & 46.1 & 24.6 & 99.4 & 62.0 & 37.8 \\
    UNETR~\cite{hatamizadeh2022unetr} & 83.3 & 55.1 & 69.2 & 30.3 & 75.0 & 52.7 & 63.9 & 24.6 & 78.6 & 42.6 & 60.6 & 21.9 \\
    STU-Net-B~\cite{huang2023stu} & 100.0 & 85.7 & 92.9 & 63.2 & 65.0 & 76.6 & 70.8 & 34.2 & 78.6 & 79.5 & 79.1 & 42.3 \\
    MedNeXt~\cite{roy2023mednext} & 66.7 & 100.0 & 83.4 & 80.0 & 55.0 & 100.0 & 77.5 & 71.0 & 42.9 & 99.0 & 71.0 & 56.2 \\
    VSmTrans~\cite{liu2024vsmtrans} & 100.0 & 87.8 & 93.9 & 66.7 & 80.0 & 80.4 & 80.2 & 44.4 & 83.3 & 80.5 & 81.9 & 45.5 \\
    \bottomrule
    \end{tabular} 
    \label{tab:benchmark_jhh}
\end{table}

\begin{table*}[t]
    \centering
    \tiny
    \caption{\textbf{Validation on PanTS}. 
Architectures are grouped by framework. 
Subgroup analysis shows substantial variation across \colorbox{orange!20}{age}, 
with younger (20$<$Age$\leq$40) and older (60$<$Age$\leq$80) patients exhibiting 
much lower F1 scores compared with mid-age groups. 
Performance across \colorbox{blue!20}{sex} subgroups is moderately different, 
with males showing noticeably reduced F1. 
Across \colorbox{purple!20}{imaging phases} (arterial, venous, non-contrast, delayed), 
arterial and non-contrast scans remain the most challenging, displaying significantly 
lower sensitivity and F1.}

    \begin{tabular}{p{0.26\linewidth}P{0.05\linewidth}P{0.05\linewidth}P{0.05\linewidth}P{0.05\linewidth}P{0.05\linewidth}P{0.05\linewidth}P{0.05\linewidth}P{0.05\linewidth}P{0.05\linewidth}P{0.05\linewidth}P{0.05\linewidth}P{0.05\linewidth}} 
    \toprule
    \multicolumn{1}{c}{\cellcolor{orange!20}\textbf{Age Groups}} & \multicolumn{4}{c}{\cellcolor{orange!20}20 \textless Age $\leq 40$} & \multicolumn{4}{c}{\cellcolor{orange!20}40 \textless Age $\leq 60$} & \multicolumn{4}{c}{\cellcolor{orange!20}60 \textless Age $\leq 80$} \\
    \cmidrule(lr){2-5}\cmidrule(lr){6-9}\cmidrule(lr){10-13}
    Method & Sen. & Spe. & AUC & F1 & Sen. & Spe. & AUC & F1 & Sen. & Spe. & AUC & F1 \\
    \midrule
    U-Net~\cite{ronneberger2015u} & 0.0 & 62.9 & 31.5 & 0.0 & 77.8 & 71.3 & 74.6 & 34.1 & 40.0 & 72.3 & 56.2 & 19.1 \\
    Swin UNETR-Tiny~\cite{tang2022self} & 50.0 & 54.3 & 52.2 & 10.5 & 72.2 & 64.4 & 68.3 & 28.0 & 70.0 & 53.5 & 61.8 & 21.9 \\
    Swin UNETR-Small~\cite{tang2022self} & 100.0 & 68.6 & 84.3 & 26.7 & 66.7 & 74.1 & 70.4 & 32.0 & 50.0 & 71.3 & 60.7 & 22.7 \\
    Swin UNETR-Base~\cite{tang2022self} & 0.0 & 77.1 & 38.6 & 0.0 & 66.7 & 80.5 & 73.6 & 37.5 & 50.0 & 71.3 & 60.7 & 22.7 \\
    Universal Model$^a$~\cite{liu2023clip} & 100.0 & 51.4 & 75.7 & 19.1 & 55.6 & 44.2 & 49.9 & 16.0 & 60.0 & 44.1 & 52.1 & 16.6 \\
    Universal Model$^b$~\cite{liu2023clip} & 0.0 & 62.9 & 31.5 & 0.0 & 50.0 & 66.7 & 58.4 & 21.2 & 55.0 & 67.3 & 61.2 & 22.7 \\ \hline
    nnU-Net~\cite{isensee2021nnu} & 50.0 & 88.6 & 69.3 & 28.6 & 72.2 & 83.3 & 77.8 & 43.3 & 30.0 & 87.1 & 58.6 & 23.1 \\
    ResEnc~\cite{isensee2024nnu} & 50.0 & 100.0 & 75.0 & 66.7 & 77.8 & 98.8 & 88.3 & 82.3 & 25.0 & 99.0 & 62.0 & 37.0 \\
    UNETR~\cite{hatamizadeh2022unetr} & 50.0 & 65.7 & 57.9 & 13.3 & 66.7 & 73.0 & 69.9 & 31.2 & 55.0 & 56.4 & 55.7 & 18.5 \\
    STU-Net-B~\cite{huang2023stu} & 50.0 & 88.6 & 69.3 & 28.6 & 72.2 & 89.1 & 80.7 & 52.0 & 30.0 & 85.6 & 57.8 & 21.8 \\
    MedNeXt~\cite{roy2023mednext} & 50.0 & 100.0 & 75.0 & 66.7 & 77.8 & 100.0 & 88.9 & 87.5 & 40.0 & 100.0 & 70.0 & 57.1 \\
    VSmTrans~\cite{liu2024vsmtrans} & 0.0 & 100.0 & 50.0 & 0.0 & 38.9 & 100.0 & 69.5 & 56.0 & 5.0 & 100.0 & 52.5 & 9.5 \\
    \midrule

    \multicolumn{1}{c}{\cellcolor{blue!20}\textbf{Sex Groups}} & \multicolumn{4}{c}{\cellcolor{blue!20}Female} & \multicolumn{4}{c}{\cellcolor{blue!20}Male} & \multicolumn{4}{c}{\cellcolor{purple!20}Arterial} \\
    \cmidrule(lr){2-5}\cmidrule(lr){6-9}\cmidrule(lr){10-13}
    Method & Sen. & Spe. & AUC & F1 & Sen. & Spe. & AUC & F1 & Sen. & Spe. & AUC & F1 \\
    \midrule
    U-Net~\cite{ronneberger2015u} & 62.1 & 75.1 & 68.6 & 37.9 & 46.1 & 68.4 & 57.3 & 12.0 & 74.1 & 65.9 & 70.0 & 43.0 \\
    Swin UNETR-Tiny~\cite{tang2022self} & 72.4 & 64.8 & 68.6 & 35.6 & 69.2 & 53.5 & 61.4 & 12.8 & 77.8 & 60.0 & 68.9 & 41.2 \\
    Swin UNETR-Small~\cite{tang2022self} & 62.1 & 74.6 & 68.4 & 37.5 & 61.5 & 71.9 & 66.7 & 17.2 & 81.5 & 78.5 & 80.0 & 56.4 \\
    Swin UNETR-Base~\cite{tang2022self} & 62.1 & 81.9 & 72.0 & 43.9 & 46.1 & 69.1 & 57.6 & 12.2 & 81.5 & 80.0 & 80.8 & 57.9 \\
    Universal Model$^a$~\cite{liu2023clip} & 55.2 & 50.3 & 52.8 & 22.7 & 76.9 & 41.0 & 59.0 & 11.5 & 44.4 & 53.3 & 48.9 & 23.5 \\
    Universal Model$^b$~\cite{liu2023clip} & 48.3 & 74.6 & 61.5 & 30.4 & 53.9 & 61.3 & 57.6 & 11.8 & 44.4 & 74.8 & 59.6 & 32.9 \\ \hline
    nnU-Net~\cite{isensee2021nnu} & 58.6 & 86.0 & 72.3 & 46.6 & 30.8 & 84.8 & 57.8 & 14.3 & 70.4 & 89.6 & 80.0 & 63.3 \\
    ResEnc~\cite{isensee2024nnu} & 55.2 & 98.5 & 76.9 & 66.7 & 38.5 & 99.2 & 68.9 & 50.0 & 66.7 & 99.3 & 83.0 & 78.3 \\
    UNETR~\cite{hatamizadeh2022unetr} & 55.2 & 71.0 & 63.1 & 31.7 & 69.2 & 60.9 & 65.1 & 14.8 & 55.6 & 63.0 & 59.3 & 32.6 \\
    STU-Net-B~\cite{huang2023stu} & 58.6 & 89.6 & 74.1 & 51.5 & 30.8 & 85.9 & 58.4 & 15.1 & 66.7 & 87.4 & 77.1 & 58.1 \\
    MedNeXt~\cite{roy2023mednext} & 58.6 & 100.0 & 79.3 & 73.9 & 53.9 & 100.0 & 77.0 & 70.0 & 77.8 & 100.0 & 88.9 & 87.5 \\
    VSmTrans~\cite{liu2024vsmtrans} & 27.6 & 100.0 & 63.8 & 43.2 & 0.0 & 100.0 & 50.0 & 0.0 & 14.8 & 100.0 & 57.4 & 25.8 \\
    \midrule

    \multicolumn{1}{c}{} & \multicolumn{4}{c}{\cellcolor{purple!20}Venous} & \multicolumn{4}{c}{\cellcolor{purple!20}Non-contrast} & \multicolumn{4}{c}{\cellcolor{purple!20}Delay} \\
    \cmidrule(lr){2-5}\cmidrule(lr){6-9}\cmidrule(lr){10-13}
    Method & Sen. & Spe. & AUC & F1 & Sen. & Spe. & AUC & F1 & Sen. & Spe. & AUC & F1 \\
    \midrule
    U-Net~\cite{ronneberger2015u} & 68.5 & 68.0 & 68.3 & 43.9 & 100.0 & 72.7 & 86.4 & 44.1 & 66.7 & 88.9 & 77.8 & 66.7 \\
    Swin UNETR-Tiny~\cite{tang2022self} & 77.8 & 46.0 & 61.9 & 37.0 & 76.9 & 74.4 & 75.7 & 37.0 & 100.0 & 44.4 & 72.2 & 54.5 \\
    Swin UNETR-Small~\cite{tang2022self} & 70.4 & 66.2 & 68.3 & 43.7 & 69.2 & 77.7 & 73.5 & 36.7 & 100.0 & 77.8 & 88.9 & 75.0 \\
    Swin UNETR-Base~\cite{tang2022self} & 70.4 & 68.9 & 69.7 & 45.4 & 76.9 & 72.7 & 74.8 & 35.7 & 66.7 & 88.9 & 77.8 & 66.7 \\
    Universal Model$^a$~\cite{liu2023clip} & 64.8 & 46.6 & 55.7 & 32.0 & 76.9 & 60.3 & 68.6 & 28.2 & 33.3 & 33.3 & 33.3 & 20.0 \\
    Universal Model$^b$~\cite{liu2023clip} & 56.5 & 62.9 & 59.7 & 35.0 & 61.5 & 80.2 & 70.9 & 35.6 & 0.0 & 55.6 & 27.8 & 0.0 \\ \hline
    nnU-Net~\cite{isensee2021nnu} & 64.8 & 85.0 & 74.9 & 55.8 & 100.0 & 79.3 & 89.7 & 51.0 & 66.7 & 77.8 & 72.3 & 57.1 \\
    ResEnc~\cite{isensee2024nnu} & 56.5 & 97.9 & 77.2 & 68.2 & 92.3 & 99.2 & 95.8 & 92.3 & 33.3 & 100.0 & 66.7 & 50.0 \\
    UNETR~\cite{hatamizadeh2022unetr} & 65.7 & 59.2 & 62.5 & 37.7 & 61.5 & 82.6 & 72.1 & 38.1 & 66.7 & 88.9 & 77.8 & 66.7 \\
    STU-Net-B~\cite{huang2023stu} & 63.0 & 87.0 & 75.0 & 56.9 & 100.0 & 84.3 & 92.2 & 57.8 & 66.7 & 77.8 & 72.3 & 57.1 \\
    MedNeXt~\cite{roy2023mednext} & 69.4 & 99.4 & 84.4 & 80.7 & 92.3 & 99.2 & 95.8 & 92.3 & 33.3 & 100.0 & 66.7 & 50.0 \\
    VSmTrans~\cite{liu2024vsmtrans} & 23.1 & 99.8 & 61.5 & 37.3 & 23.1 & 100.0 & 61.6 & 37.5 & 0.0 & 100.0 & 50.0 & 0.0 \\
    \bottomrule
    \end{tabular} 
    \label{tab:benchmark_pants}
\end{table*}

\begin{table*}[t]
    \centering
    \tiny
    \caption{\textbf{Validation on Merlin}. 
Architectures are grouped by framework. 
Subgroup analysis reveals substantial variation across \colorbox{orange!20}{age}, 
with younger (20$<$Age$\leq$40) and older (60$<$Age$\leq$80) patients exhibiting 
markedly lower F1 scores than mid-age groups. 
Performance across \colorbox{blue!20}{sex} shows moderate differences, with males 
generally achieving lower sensitivity and F1. 
Across \colorbox{green!20}{race}, Asian patients consistently achieve higher F1, 
while White and Black subgroups show pronounced drops. 
For \colorbox{purple!20}{imaging phases}, arterial cases remain the most challenging, 
displaying considerably lower sensitivity and F1 compared with venous and non-contrast scans.}

    \begin{tabular}{p{0.26\linewidth}P{0.05\linewidth}P{0.05\linewidth}P{0.05\linewidth}P{0.05\linewidth}P{0.05\linewidth}P{0.05\linewidth}P{0.05\linewidth}P{0.05\linewidth}P{0.05\linewidth}P{0.05\linewidth}P{0.05\linewidth}P{0.05\linewidth}} 
    \toprule
    \multicolumn{1}{c}{\cellcolor{orange!20}\textbf{Age Groups}} & \multicolumn{4}{c}{\cellcolor{orange!20}20 \textless Age $\leq 40$} & \multicolumn{4}{c}{\cellcolor{orange!20}40 \textless Age $\leq 60$} & \multicolumn{4}{c}{\cellcolor{orange!20}60 \textless Age $\leq 80$} \\
    \cmidrule(lr){2-5}\cmidrule(lr){6-9}\cmidrule(lr){10-13}
    Method & Sen. & Spe. & AUC & F1 & Sen. & Spe. & AUC & F1 & Sen. & Spe. & AUC & F1 \\
    \midrule
    U-Net~\cite{ronneberger2015u} & 25.0 & 74.2 & 49.6 & 3.6 & 38.9 & 72.3 & 55.6 & 13.0 & 55.6 & 67.2 & 61.4 & 21.6 \\
    Swin UNETR-Tiny~\cite{tang2022self} & 75.0 & 60.8 & 67.9 & 7.2 & 55.6 & 51.7 & 53.7 & 11.6 & 63.0 & 49.3 & 56.2 & 17.5 \\
    Swin UNETR-Small~\cite{tang2022self} & 50.0 & 62.9 & 56.5 & 5.1 & 61.1 & 68.0 & 64.6 & 17.6 & 55.6 & 60.5 & 58.1 & 18.9 \\
    Swin UNETR-Base~\cite{tang2022self} & 50.0 & 71.7 & 60.9 & 6.6 & 55.6 & 68.0 & 61.8 & 16.1 & 48.1 & 62.8 & 55.5 & 17.3 \\
    Universal Model$^a$~\cite{liu2023clip} & 0.0 & 77.8 & 38.9 & 0.0 & 16.7 & 74.0 & 45.4 & 6.1 & 29.6 & 81.1 & 55.4 & 17.6 \\
    Universal Model$^b$~\cite{liu2023clip} & 0.0 & 86.1 & 43.1 & 0.0 & 5.6 & 86.0 & 45.8 & 3.3 & 14.8 & 88.5 & 51.7 & 12.3 \\ \hline
    nnU-Net~\cite{isensee2021nnu} & 50.0 & 86.1 & 68.1 & 12.1 & 33.3 & 83.7 & 58.5 & 16.4 & 51.9 & 82.1 & 67.0 & 29.8 \\
    ResEnc~\cite{isensee2024nnu} & 25.0 & 97.9 & 61.5 & 22.2 & 11.1 & 95.7 & 53.4 & 12.1 & 25.9 & 96.0 & 61.0 & 30.4 \\
    UNETR~\cite{hatamizadeh2022unetr} & 50.0 & 59.8 & 54.9 & 4.8 & 55.6 & 50.7 & 53.2 & 11.4 & 51.9 & 56.8 & 54.4 & 16.6 \\
    STU-Net-B~\cite{huang2023stu} & 25.0 & 88.1 & 56.6 & 7.1 & 22.2 & 88.3 & 55.3 & 14.0 & 55.6 & 84.1 & 69.9 & 33.7 \\
    MedNeXt~\cite{roy2023mednext} & 25.0 & 95.7 & 60.4 & 15.4 & 29.4 & 95.3 & 62.4 & 28.6 & 52.2 & 92.9 & 72.6 & 44.4 \\
    VSmTrans~\cite{liu2024vsmtrans} & 0.0 & 99.5 & 49.8 & 0.0 & 0.0 & 99.3 & 49.7 & 0.0 & 7.4 & 99.0 & 53.2 & 12.5 \\
    \midrule

    \multicolumn{1}{c}{\cellcolor{blue!20}\textbf{Sex Groups}} & \multicolumn{4}{c}{\cellcolor{blue!20}Male} & \multicolumn{4}{c}{\cellcolor{blue!20}Female} & \multicolumn{4}{c}{\cellcolor{orange!20}Age \textgreater 80} \\
    \cmidrule(lr){2-5}\cmidrule(lr){6-9}\cmidrule(lr){10-13}
    Method & Sen. & Spe. & AUC & F1 & Sen. & Spe. & AUC & F1 & Sen. & Spe. & AUC & F1 \\
    \midrule
    U-Net~\cite{ronneberger2015u} & 52.4 & 70.6 & 61.5 & 23.9 & 41.2 & 72.8 & 57.0 & 15.4 & 46.1 & 78.0 & 62.1 & 41.4 \\
    Swin UNETR-Tiny & 66.7 & 49.5 & 58.1 & 20.3 & 50.0 & 57.9 & 54.0 & 13.2 & 53.9 & 60.4 & 57.2 & 36.8 \\
    Swin UNETR-Small~\cite{tang2022self} & 66.7 & 63.7 & 65.2 & 25.7 & 41.2 & 65.8 & 53.5 & 13.0 & 50.0 & 72.5 & 61.3 & 40.6 \\
    Swin UNETR-Base~\cite{tang2022self} & 59.5 & 64.5 & 62.0 & 23.6 & 38.2 & 70.3 & 54.3 & 13.5 & 46.1 & 73.6 & 59.9 & 38.7 \\
    Universal Model$^a$~\cite{liu2023clip} & 19.1 & 83.1 & 51.1 & 13.4 & 20.6 & 76.0 & 48.3 & 8.8 & 15.4 & 92.3 & 53.9 & 21.6 \\
    Universal Model$^b$~\cite{liu2023clip} & 11.9 & 88.2 & 50.1 & 10.5 & 14.7 & 86.4 & 50.6 & 9.4 & 19.2 & 86.8 & 53.0 & 23.3 \\ \hline
    nnU-Net~\cite{isensee2021nnu} & 52.4 & 84.6 & 68.5 & 34.6 & 47.1 & 83.9 & 65.5 & 24.8 & 57.7 & 85.7 & 71.7 & 55.6 \\
    ResEnc~\cite{isensee2024nnu} & 21.4 & 96.6 & 59.0 & 27.7 & 26.5 & 96.1 & 61.3 & 29.0 & 26.9 & 95.6 & 61.3 & 37.8 \\
    UNETR~\cite{hatamizadeh2022unetr} & 64.3 & 54.4 & 59.4 & 21.2 & 29.4 & 58.3 & 43.9 & 8.0 & 42.3 & 69.2 & 55.8 & 33.9 \\
    STU-Net-B~\cite{huang2023stu} & 47.6 & 88.0 & 67.8 & 36.0 & 38.2 & 86.2 & 62.2 & 22.6 & 46.1 & 90.1 & 68.1 & 51.1 \\
    MedNeXt~\cite{roy2023mednext} & 43.6 & 96.0 & 69.8 & 47.9 & 35.7 & 92.9 & 64.3 & 28.6 & 36.4 & 90.7 & 63.6 & 43.2 \\
    VSmTrans~\cite{liu2024vsmtrans} & 0.0 & 99.0 & 49.5 & 0.0 & 5.9 & 99.2 & 52.6 & 10.0 & 0.0 & 97.8 & 48.9 & 0.0 \\
    \midrule

    \multicolumn{1}{c}{\cellcolor{green!20}\textbf{Race Groups}} & \multicolumn{4}{c}{\cellcolor{green!20}Asian} & \multicolumn{4}{c}{\cellcolor{green!20}White} & \multicolumn{4}{c}{\cellcolor{green!20}Black} \\
    \cmidrule(lr){2-5}\cmidrule(lr){6-9}\cmidrule(lr){10-13}
    Method & Sen. & Spe. & AUC & F1 & Sen. & Spe. & AUC & F1 & Sen. & Spe. & AUC & F1 \\
    \midrule
    U-Net~\cite{ronneberger2015u} & 54.5 & 80.7 & 67.6 & 31.6 & 47.9 & 68.7 & 58.3 & 21.5 & 0.0 & 72.5 & 36.3 & 0.0 \\
    Swin UNETR-Tiny~\cite{tang2022self} & 54.5 & 56.9 & 55.7 & 18.8 & 56.2 & 52.1 & 54.2 & 18.4 & 50.0 & 52.2 & 51.1 & 5.6 \\
    Swin UNETR-Small~\cite{tang2022self} & 63.6 & 72.5 & 68.1 & 29.2 & 50.0 & 62.6 & 56.3 & 19.8 & 50.0 & 65.2 & 57.6 & 7.4 \\
    Swin UNETR-Base~\cite{tang2022self} & 45.5 & 76.2 & 60.9 & 23.8 & 45.8 & 64.8 & 55.3 & 19.1 & 50.0 & 73.9 & 62.0 & 9.5 \\
    Universal Model$^a$~\cite{liu2023clip} & 0.0 & 90.8 & 45.4 & 0.0 & 20.8 & 79.4 & 50.1 & 13.2 & 0.0 & 79.7 & 39.9 & 0.0 \\
    Universal Model$^b$~\cite{liu2023clip} & 9.1 & 93.6 & 51.4 & 10.5 & 8.3 & 88.6 & 48.5 & 7.7 & 50.0 & 81.2 & 65.6 & 12.5 \\ \hline
    nnU-Net~\cite{isensee2021nnu} & 54.5 & 88.1 & 71.3 & 40.0 & 45.8 & 82.3 & 64.1 & 29.1 & 0.0 & 82.6 & 41.3 & 0.0 \\
    ResEnc~\cite{isensee2024nnu} & 36.4 & 97.2 & 66.8 & 44.4 & 20.8 & 96.9 & 58.9 & 27.8 & 0.0 & 94.2 & 47.1 & 0.0 \\
    UNETR~\cite{hatamizadeh2022unetr} & 54.5 & 62.4 & 58.5 & 20.7 & 41.7 & 54.7 & 48.2 & 14.6 & 0.0 & 62.3 & 31.2 & 0.0 \\
    STU-Net-B~\cite{huang2023stu} & 54.5 & 90.8 & 72.7 & 44.4 & 39.6 & 84.9 & 62.3 & 27.9 & 0.0 & 87.0 & 43.5 & 0.0 \\
    MedNeXt~\cite{roy2023mednext} & 60.0 & 95.2 & 77.6 & 57.1 & 40.0 & 94.5 & 67.3 & 40.5 & 0.0 & 91.2 & 45.6 & 0.0 \\
    VSmTrans~\cite{liu2024vsmtrans} & 0.0 & 98.2 & 49.1 & 0.0 & 2.1 & 99.1 & 50.6 & 3.8 & 0.0 & 100.0 & 50.0 & 0.0 \\
    \midrule

    \multicolumn{1}{c}{\cellcolor{purple!20}\textbf{Phase Groups}} & \multicolumn{4}{c}{\cellcolor{purple!20}Arterial} & \multicolumn{4}{c}{\cellcolor{purple!20}Venous} & \multicolumn{4}{c}{\cellcolor{green!20}Pacific Isl.} \\
    \cmidrule(lr){2-5}\cmidrule(lr){6-9}\cmidrule(lr){10-13}
    Method & Sen. & Spe. & AUC & F1 & Sen. & Spe. & AUC & F1 & Sen. & Spe. & AUC & F1 \\
    \midrule
    U-Net~\cite{ronneberger2015u} & 43.8 & 71.3 & 57.6 & 25.0 & 51.2 & 71.4 & 61.3 & 17.8 & 0.0 & 56.2 & 28.1 & 0.0 \\
    Swin UNETR-Tiny~\cite{tang2022self} & 53.1 & 59.1 & 56.1 & 23.8 & 65.1 & 50.8 & 58.0 & 14.6 & 0.0 & 43.8 & 21.9 & 0.0 \\
    Swin UNETR-Small~\cite{tang2022self} & 50.0 & 70.0 & 60.0 & 27.4 & 60.5 & 62.4 & 61.5 & 16.9 & 0.0 & 50.0 & 25.0 & 0.0 \\
    Swin UNETR-Base~\cite{tang2022self} & 43.8 & 73.9 & 58.9 & 26.4 & 55.8 & 65.2 & 60.5 & 16.7 & 100.0 & 43.8 & 71.9 & 18.2 \\
    Universal Model$^a$~\cite{liu2023clip} & 18.8 & 82.2 & 50.5 & 15.2 & 20.9 & 77.7 & 49.3 & 9.3 & 0.0 & 50.0 & 25.0 & 0.0 \\
    Universal Model$^b$~\cite{liu2023clip} & 12.5 & 88.3 & 50.4 & 12.7 & 13.9 & 86.3 & 50.1 & 8.8 & 0.0 & 68.8 & 34.4 & 0.0 \\ \hline
    nnU-Net~\cite{isensee2021nnu} & 50.0 & 86.1 & 68.1 & 40.0 & 51.2 & 83.3 & 67.3 & 25.7 & 100.0 & 81.2 & 90.6 & 40.0 \\
    ResEnc~\cite{isensee2024nnu} & 15.6 & 97.8 & 56.7 & 23.8 & 30.2 & 95.9 & 63.1 & 31.7 & 0.0 & 100.0 & 50.0 & 0.0 \\
    UNETR~\cite{hatamizadeh2022unetr} & 37.5 & 71.7 & 54.6 & 22.0 & 58.1 & 49.4 & 53.8 & 12.8 & 100.0 & 56.2 & 78.1 & 22.2 \\
    STU-Net-B~\cite{huang2023stu} & 43.8 & 87.4 & 65.6 & 37.3 & 44.2 & 87.0 & 65.6 & 26.2 & 0.0 & 75.0 & 37.5 & 0.0 \\
    MedNeXt~\cite{roy2023mednext} & 37.5 & 94.0 & 65.8 & 41.9 & 41.9 & 94.3 & 68.1 & 37.1 & 0.0 & 92.9 & 46.5 & 0.0 \\
    VSmTrans~\cite{liu2024vsmtrans} & 0.0 & 100.0 & 50.0 & 0.0 & 4.7 & 98.7 & 51.7 & 7.5 & 0.0 & 100.0 & 50.0 & 0.0 \\
    \bottomrule
    \end{tabular} 
    \label{tab:benchmark_merlin}
\end{table*}

\begin{table*}[t]
    \centering
    \tiny
    \caption{\textbf{Validation on N--California}. 
Architectures are grouped by framework. 
Subgroup analysis reveals substantial variation across \colorbox{orange!20}{age}, 
with both younger (20$<$Age$\leq$40) and older (60$<$Age$\leq$80) patients showing 
notably reduced F1 scores compared with mid-age groups. 
Differences across \colorbox{blue!20}{sex} are moderate, with males consistently 
exhibiting lower sensitivity and F1. 
For \colorbox{green!20}{race}, Asian patients generally achieve higher F1, while 
White and Black subgroups show considerable drops. 
Across \colorbox{purple!20}{phase} groups (venous, non-contrast, arterial), arterial scans 
remain the most challenging, demonstrating the lowest sensitivity and F1.}
    \begin{tabular}{p{0.26\linewidth}P{0.05\linewidth}P{0.05\linewidth}P{0.05\linewidth}P{0.05\linewidth}P{0.05\linewidth}P{0.05\linewidth}P{0.05\linewidth}P{0.05\linewidth}P{0.05\linewidth}P{0.05\linewidth}P{0.05\linewidth}P{0.05\linewidth}} 
    \toprule
    \multicolumn{1}{c}{\cellcolor{orange!20}\textbf{Age Groups}} & \multicolumn{4}{c}{\cellcolor{orange!20}20 \textless Age $\leq 40$} & \multicolumn{4}{c}{\cellcolor{orange!20}40 \textless Age $\leq 60$} & \multicolumn{4}{c}{\cellcolor{orange!20}60 \textless Age $\leq 80$} \\
    \cmidrule(lr){2-5}\cmidrule(lr){6-9}\cmidrule(lr){10-13}
    Method & Sen. & Spe. & AUC & F1 & Sen. & Spe. & AUC & F1 & Sen. & Spe. & AUC & F1 \\
    \midrule
    U-Net~\cite{ronneberger2015u} & 33.3 & 83.9 & 58.6 & 15.4 & 46.7 & 77.0 & 61.9 & 16.1 & 53.1 & 76.7 & 64.9 & 28.3 \\
    Swin UNETR-Tiny~\cite{tang2022self} & 66.7 & 82.1 & 74.4 & 26.7 & 60.0 & 68.1 & 64.1 & 15.8 & 49.0 & 66.4 & 57.7 & 20.9 \\
    Swin UNETR-Small~\cite{tang2022self} & 33.3 & 78.6 & 56.0 & 12.5 & 53.3 & 79.4 & 66.4 & 19.8 & 38.8 & 83.3 & 61.1 & 26.0 \\
    Swin UNETR-Base~\cite{tang2022self} & 66.7 & 74.0 & 70.4 & 22.2 & 78.6 & 75.8 & 77.2 & 24.7 & 42.5 & 74.1 & 58.3 & 22.4 \\
    Universal Model$^a$~\cite{liu2023clip} & 66.7 & 76.8 & 71.8 & 22.2 & 13.3 & 77.3 & 45.3 & 4.9 & 14.3 & 82.4 & 48.4 & 10.1 \\
    Universal Model$^b$~\cite{liu2023clip} & 33.3 & 91.1 & 62.2 & 22.2 & 6.7 & 85.5 & 46.1 & 3.5 & 10.2 & 92.1 & 51.2 & 11.0 \\ \hline
    nnU-Net~\cite{isensee2021nnu} & 100.0 & 83.9 & 92.0 & 40.0 & 60.0 & 84.4 & 72.2 & 26.5 & 38.8 & 87.2 & 63.0 & 29.7 \\
    ResEnc~\cite{isensee2024nnu} & 66.7 & 92.9 & 79.8 & 44.4 & 33.3 & 96.5 & 64.9 & 33.3 & 18.4 & 98.3 & 58.4 & 27.3 \\
    UNETR~\cite{hatamizadeh2022unetr} & 66.7 & 87.5 & 77.1 & 33.3 & 33.3 & 85.5 & 59.4 & 16.4 & 30.6 & 81.4 & 56.0 & 19.9 \\
    STU-Net-B~\cite{huang2023stu} & 66.7 & 87.5 & 77.1 & 33.3 & 53.3 & 82.6 & 68.0 & 22.2 & 28.6 & 84.8 & 56.7 & 20.9 \\
    MedNeXt~\cite{roy2023mednext} & 66.7 & 92.9 & 79.8 & 44.4 & 53.3 & 97.9 & 75.6 & 55.2 & 39.6 & 96.1 & 67.9 & 44.7 \\
    VSmTrans~\cite{liu2024vsmtrans} & 0.0 & 98.2 & 49.1 & 0.0 & 6.7 & 99.3 & 53.0 & 11.1 & 2.0 & 99.8 & 50.9 & 3.9 \\
    \midrule

    \multicolumn{1}{c}{\cellcolor{blue!20}\textbf{Sex Groups}} & \multicolumn{4}{c}{\cellcolor{blue!20}Female} & \multicolumn{4}{c}{\cellcolor{blue!20}Male} & \multicolumn{4}{c}{\cellcolor{orange!20}Age \textgreater 80} \\
    \cmidrule(lr){2-5}\cmidrule(lr){6-9}\cmidrule(lr){10-13}
    Method & Sen. & Spe. & AUC & F1 & Sen. & Spe. & AUC & F1 & Sen. & Spe. & AUC & F1 \\
    \midrule
    U-Net~\cite{ronneberger2015u} & 40.0 & 84.8 & 62.4 & 28.8 & 52.3 & 70.9 & 61.6 & 21.8 & 31.8 & 76.5 & 54.2 & 26.9 \\
    Swin UNETR-Tiny~\cite{tang2022self} & 51.1 & 75.8 & 63.5 & 27.5 & 50.0 & 61.6 & 55.8 & 17.2 & 45.5 & 67.3 & 56.4 & 31.2 \\
    Swin UNETR-Small~\cite{tang2022self} & 46.7 & 83.9 & 65.3 & 31.8 & 38.6 & 78.4 & 58.5 & 20.2 & 45.5 & 74.5 & 60.0 & 35.1 \\
    Swin UNETR-Base~\cite{tang2022self} & 47.7 & 77.4 & 62.6 & 28.0 & 53.9 & 70.1 & 62.0 & 21.2 & 47.4 & 62.2 & 54.8 & 29.0 \\
    Universal Model$^a$~\cite{liu2023clip} & 22.2 & 82.9 & 52.6 & 16.0 & 13.6 & 78.8 & 46.2 & 7.7 & 22.7 & 83.7 & 53.2 & 23.3 \\
    Universal Model$^b$~\cite{liu2023clip} & 11.1 & 93.4 & 52.3 & 13.0 & 13.6 & 87.7 & 50.7 & 10.8 & 18.2 & 94.9 & 56.6 & 25.8 \\ \hline
    nnU-Net~\cite{isensee2021nnu} & 46.7 & 89.5 & 68.1 & 38.5 & 34.1 & 82.4 & 58.3 & 20.6 & 22.7 & 82.7 & 52.7 & 22.7 \\
    ResEnc~\cite{isensee2024nnu} & 24.4 & 97.1 & 60.8 & 32.4 & 18.2 & 97.2 & 57.7 & 24.2 & 13.6 & 95.9 & 54.8 & 20.7 \\
    UNETR~\cite{hatamizadeh2022unetr} & 26.7 & 83.6 & 55.2 & 19.4 & 36.4 & 81.8 & 59.1 & 21.3 & 27.3 & 77.5 & 52.4 & 24.0 \\
    STU-Net-B~\cite{huang2023stu} & 35.6 & 87.5 & 61.6 & 28.6 & 34.1 & 81.2 & 57.7 & 19.7 & 31.8 & 82.7 & 57.3 & 30.4 \\
    MedNeXt~\cite{roy2023mednext} & 44.4 & 95.3 & 69.9 & 47.6 & 30.2 & 97.0 & 63.6 & 36.6 & 18.2 & 93.9 & 56.1 & 25.0 \\
    VSmTrans~\cite{liu2024vsmtrans} & 2.2 & 99.5 & 50.9 & 4.2 & 2.3 & 99.2 & 50.8 & 4.1 & 0.0 & 98.0 & 49.0 & 0.0 \\
    \midrule

    \multicolumn{1}{c}{\cellcolor{green!20}\textbf{Race Groups}} & \multicolumn{4}{c}{\cellcolor{green!20}Asian} & \multicolumn{4}{c}{\cellcolor{green!20}White} & \multicolumn{4}{c}{\cellcolor{green!20}Black} \\
    \cmidrule(lr){2-5}\cmidrule(lr){6-9}\cmidrule(lr){10-13}
    Method & Sen. & Spe. & AUC & F1 & Sen. & Spe. & AUC & F1 & Sen. & Spe. & AUC & F1 \\
    \midrule
    U-Net~\cite{ronneberger2015u} & 47.6 & 79.2 & 63.4 & 31.8 & 41.2 & 76.2 & 58.7 & 21.3 & 16.7 & 79.3 & 48.0 & 8.3 \\
    Swin UNETR-Tiny~\cite{tang2022self} & 52.4 & 75.3 & 63.9 & 31.4 & 52.9 & 66.9 & 59.9 & 21.4 & 33.3 & 69.5 & 51.4 & 12.1 \\
    Swin UNETR-Small~\cite{tang2022self} & 38.1 & 84.4 & 61.3 & 30.2 & 47.1 & 80.2 & 63.7 & 26.8 & 33.3 & 81.7 & 57.5 & 17.4 \\
    Swin UNETR-Base~\cite{tang2022self} & 47.4 & 81.9 & 64.7 & 33.3 & 58.3 & 71.4 & 64.9 & 26.2 & 0.0 & 77.9 & 39.0 & 0.0 \\
    Universal Model$^a$~\cite{liu2023clip} & 4.8 & 89.0 & 46.9 & 5.1 & 21.6 & 80.6 & 51.1 & 13.4 & 16.7 & 82.9 & 49.8 & 9.5 \\
    Universal Model$^b$~\cite{liu2023clip} & 0.0 & 97.4 & 48.7 & 0.0 & 13.7 & 89.5 & 51.6 & 12.4 & 16.7 & 92.7 & 54.7 & 15.4 \\ \hline
    nnU-Net~\cite{isensee2021nnu} & 19.1 & 88.3 & 53.7 & 18.6 & 43.1 & 84.8 & 64.0 & 28.8 & 33.3 & 84.2 & 58.8 & 19.1 \\
    ResEnc~\cite{isensee2024nnu} & 14.3 & 95.5 & 54.9 & 19.4 & 23.5 & 97.7 & 60.6 & 32.0 & 16.7 & 98.8 & 57.8 & 25.0 \\
    UNETR~\cite{hatamizadeh2022unetr} & 28.6 & 83.1 & 55.9 & 22.6 & 35.3 & 83.8 & 59.6 & 23.4 & 0.0 & 84.2 & 42.1 & 0.0 \\
    STU-Net-B~\cite{huang2023stu} & 19.1 & 87.7 & 53.4 & 18.2 & 43.1 & 82.9 & 63.0 & 27.0 & 0.0 & 87.8 & 43.9 & 0.0 \\
    MedNeXt~\cite{roy2023mednext} & 23.8 & 94.8 & 59.3 & 29.4 & 44.0 & 96.6 & 70.3 & 48.9 & 0.0 & 96.3 & 48.2 & 0.0 \\
    VSmTrans~\cite{liu2024vsmtrans} & 0.0 & 100.0 & 50.0 & 0.0 & 2.0 & 99.4 & 50.7 & 3.6 & 0.0 & 100.0 & 50.0 & 0.0 \\
    \midrule

    \multicolumn{1}{c}{\cellcolor{purple!20}\textbf{Phase Groups}} & \multicolumn{4}{c}{\cellcolor{purple!20}Venous} & \multicolumn{4}{c}{\cellcolor{purple!20}Non-Contrast} & \multicolumn{4}{c}{\cellcolor{purple!20}Arterial} \\
    \cmidrule(lr){2-5}\cmidrule(lr){6-9}\cmidrule(lr){10-13}
    Method & Sen. & Spe. & AUC & F1 & Sen. & Spe. & AUC & F1 & Sen. & Spe. & AUC & F1 \\
    \midrule
    U-Net~\cite{ronneberger2015u} & 67.9 & 71.5 & 69.7 & 31.9 & 30.6 & 79.7 & 55.2 & 15.5 & 44.0 & 78.8 & 61.4 & 29.3 \\
    Swin UNETR-Tiny~\cite{tang2022self} & 85.7 & 51.8 & 68.8 & 27.6 & 19.4 & 76.9 & 48.2 & 9.3 & 56.0 & 67.9 & 62.0 & 28.6 \\
    Swin UNETR-Small~\cite{tang2022self} & 85.7 & 67.6 & 76.7 & 35.8 & 5.6 & 89.3 & 47.5 & 4.5 & 48.0 & 77.7 & 62.9 & 30.8 \\
    Swin UNETR-Base~\cite{tang2022self} & 70.4 & 76.9 & 73.7 & 37.6 & 26.5 & 71.4 & 49.0 & 10.9 & 63.6 & 73.4 & 68.5 & 34.1 \\
    Universal Model$^a$~\cite{liu2023clip} & 21.4 & 79.5 & 50.5 & 13.9 & 22.2 & 78.8 & 50.5 & 11.2 & 8.0 & 87.0 & 47.5 & 7.8 \\
    Universal Model$^b$~\cite{liu2023clip} & 10.7 & 86.2 & 48.5 & 9.1 & 13.9 & 91.2 & 52.6 & 12.2 & 12.0 & 93.5 & 52.8 & 15.0 \\ \hline
    nnU-Net~\cite{isensee2021nnu} & 53.6 & 86.6 & 70.1 & 39.0 & 30.6 & 84.8 & 57.7 & 18.6 & 40.0 & 86.4 & 63.2 & 33.3 \\
    ResEnc~\cite{isensee2024nnu} & 46.4 & 92.5 & 69.5 & 43.3 & 5.6 & 99.6 & 52.6 & 10.0 & 16.0 & 97.3 & 56.7 & 23.5 \\
    UNETR~\cite{hatamizadeh2022unetr} & 64.3 & 70.8 & 67.6 & 30.0 & 8.3 & 89.9 & 49.1 & 7.0 & 28.0 & 80.4 & 54.2 & 20.6 \\
    STU-Net-B~\cite{huang2023stu} & 60.7 & 86.6 & 73.7 & 43.0 & 16.7 & 80.3 & 48.5 & 9.0 & 32.0 & 90.2 & 61.1 & 31.4 \\
    MedNeXt~\cite{roy2023mednext} & 71.4 & 91.3 & 81.4 & 57.1 & 13.9 & 98.5 & 56.2 & 20.8 & 33.3 & 97.3 & 65.3 & 43.2 \\
    VSmTrans~\cite{liu2024vsmtrans} & 7.1 & 97.6 & 52.4 & 11.1 & 0.0 & 100.0 & 50.0 & 0.0 & 0.0 & 100.0 & 50.0 & 0.0 \\
    \bottomrule
    \end{tabular} 
    \label{tab:benchmark_N-California}
\end{table*}

\begin{table*}[t]
    \centering
    \tiny
    \caption{\textbf{Validation on S.\ Europe}. 
Architectures are grouped by framework. 
Subgroup analysis reveals substantial variation across \colorbox{orange!20}{age}, 
with both middle-aged (40$<$Age$\leq$60) and older (Age$>$80) groups showing 
notably reduced F1 scores. 
Across \colorbox{blue!20}{sex}, Turkish males exhibit the lowest sensitivity and F1. 
For \colorbox{green!20}{race}, Turkish and Eastern European subgroups show consistently 
lower performance than Asian patients. 
Across \colorbox{purple!20}{imaging phases}, arterial scans remain the most challenging, 
while venous scans yield higher AUC and F1.}
    \begin{tabular}{p{0.26\linewidth}P{0.05\linewidth}P{0.05\linewidth}P{0.05\linewidth}P{0.05\linewidth}P{0.05\linewidth}P{0.05\linewidth}P{0.05\linewidth}P{0.05\linewidth}P{0.05\linewidth}P{0.05\linewidth}P{0.05\linewidth}P{0.05\linewidth}} 
    \toprule
    \multicolumn{1}{c}{\cellcolor{orange!20}\textbf{Age Groups}} & \multicolumn{4}{c}{\cellcolor{orange!20}40 \textless Age $\leq 60$} & \multicolumn{4}{c}{\cellcolor{orange!20}60 \textless Age $\leq 80$} & \multicolumn{4}{c}{\cellcolor{orange!20}Age \textgreater 80} \\
    \cmidrule(lr){2-5}\cmidrule(lr){6-9}\cmidrule(lr){10-13}
    Method & Sen. & Spe. & AUC & F1 & Sen. & Spe. & AUC & F1 & Sen. & Spe. & AUC & F1 \\
    \midrule
    U-Net~\cite{ronneberger2015u} & 45.0 & 56.9 & 51.0 & 9.2 & 71.4 & 58.1 & 64.8 & 13.6 & 50.0 & 79.0 & 64.5 & 13.3 \\
    Swin UNETR-Tiny~\cite{tang2022self} & 60.0 & 47.9 & 54.0 & 10.3 & 76.2 & 42.0 & 59.1 & 10.9 & 50.0 & 57.1 & 53.6 & 7.4 \\
    Swin UNETR-Small~\cite{tang2022self} & 20.0 & 67.4 & 43.7 & 5.3 & 61.9 & 60.5 & 61.2 & 12.5 & 50.0 & 75.0 & 62.5 & 11.8 \\
    Swin UNETR-Base~\cite{tang2022self} & 35.0 & 69.2 & 52.1 & 9.6 & 61.9 & 58.3 & 60.1 & 11.9 & 50.0 & 71.4 & 60.7 & 10.5 \\
    Universal Model$^a$~\cite{liu2023clip} & 75.0 & 47.8 & 61.4 & 12.7 & 47.6 & 61.9 & 54.8 & 10.1 & 0.0 & 52.6 & 26.3 & 0.0 \\
    Universal Model$^b$~\cite{liu2023clip} & 30.0 & 73.9 & 52.0 & 9.4 & 23.8 & 70.8 & 47.3 & 6.5 & 0.0 & 79.0 & 39.5 & 0.0 \\ \hline
    nnU-Net~\cite{isensee2021nnu} & 65.0 & 78.8 & 71.9 & 22.6 & 66.7 & 77.1 & 71.9 & 20.6 & 50.0 & 92.9 & 71.5 & 28.6 \\
    ResEnc~\cite{isensee2024nnu} & 10.0 & 97.7 & 53.9 & 12.9 & 19.1 & 95.5 & 57.3 & 17.8 & 0.0 & 94.6 & 47.3 & 0.0 \\
    UNETR~\cite{hatamizadeh2022unetr} & 35.0 & 58.5 & 46.8 & 7.5 & 52.4 & 51.5 & 52.0 & 8.9 & 100.0 & 69.6 & 84.8 & 19.1 \\
    STU-Net-B~\cite{huang2023stu} & 70.0 & 75.4 & 72.7 & 21.7 & 61.9 & 73.7 & 67.8 & 17.3 & 50.0 & 82.1 & 66.1 & 15.4 \\
    MedNeXt~\cite{roy2023mednext} & 45.0 & 92.8 & 68.9 & 31.6 & 47.6 & 94.1 & 70.9 & 35.1 & 50.0 & 94.6 & 72.3 & 33.3 \\
    VSmTrans~\cite{liu2024vsmtrans} & 10.0 & 99.7 & 54.9 & 17.4 & 9.5 & 99.5 & 54.5 & 16.0 & 0.0 & 100.0 & 50.0 & 0.0 \\
    \midrule

    \multicolumn{1}{c}{\cellcolor{blue!20}\textbf{Sex Groups}} & \multicolumn{4}{c}{\cellcolor{blue!20}Female} & \multicolumn{4}{c}{\cellcolor{blue!20}Male} & \multicolumn{4}{c}{\cellcolor{green!20}Turkish} \\
    \cmidrule(lr){2-5}\cmidrule(lr){6-9}\cmidrule(lr){10-13}
    Method & Sen. & Spe. & AUC & F1 & Sen. & Spe. & AUC & F1 & Sen. & Spe. & AUC & F1 \\
    \midrule
    U-Net~\cite{ronneberger2015u} & 35.3 & 63.6 & 49.5 & 5.8 & 73.1 & 53.5 & 63.3 & 13.8 & 54.3 & 57.5 & 55.9 & 8.8 \\
    Swin UNETR-Tiny~\cite{tang2022self} & 70.6 & 48.8 & 59.7 & 8.4 & 65.4 & 42.2 & 53.8 & 10.3 & 65.7 & 44.2 & 55.0 & 8.3 \\
    Swin UNETR-Small~\cite{tang2022self} & 29.4 & 67.9 & 48.7 & 5.5 & 50.0 & 57.6 & 53.8 & 10.4 & 40.0 & 63.4 & 51.7 & 7.5 \\
    Swin UNETR-Base~\cite{tang2022self} & 35.3 & 65.1 & 50.2 & 6.1 & 57.7 & 61.2 & 59.5 & 12.9 & 40.0 & 62.3 & 51.2 & 7.3 \\
    Universal Model$^a$~\cite{liu2023clip} & 82.3 & 58.0 & 70.2 & 11.6 & 42.3 & 54.2 & 48.3 & 8.3 & 60.0 & 55.5 & 57.8 & 9.3 \\
    Universal Model$^b$~\cite{liu2023clip} & 29.4 & 77.3 & 53.4 & 7.3 & 23.1 & 69.0 & 46.1 & 6.5 & 25.7 & 73.0 & 49.4 & 6.3 \\ \hline
    nnU-Net~\cite{isensee2021nnu} & 47.1 & 80.7 & 63.9 & 13.1 & 76.9 & 76.2 & 76.6 & 24.4 & 62.9 & 78.5 & 70.7 & 17.7 \\
    ResEnc~\cite{isensee2024nnu} & 11.8 & 96.4 & 54.1 & 10.8 & 15.4 & 95.8 & 55.6 & 15.7 & 17.1 & 95.6 & 56.4 & 15.0 \\
    UNETR~\cite{hatamizadeh2022unetr} & 23.5 & 54.4 & 39.0 & 3.2 & 61.5 & 57.2 & 59.4 & 12.6 & 37.1 & 55.1 & 46.1 & 5.8 \\
    STU-Net-B~\cite{huang2023stu} & 64.7 & 76.1 & 70.4 & 14.9 & 65.4 & 74.5 & 70.0 & 20.1 & 60.0 & 75.1 & 67.6 & 15.1 \\
    MedNeXt~\cite{roy2023mednext} & 29.4 & 93.0 & 61.2 & 17.5 & 57.7 & 91.9 & 74.8 & 37.0 & 40.0 & 91.7 & 65.9 & 22.8 \\
    VSmTrans~\cite{liu2024vsmtrans} & 23.5 & 99.6 & 61.6 & 34.8 & 0.0 & 99.8 & 49.9 & 0.0 & 11.4 & 99.7 & 55.6 & 19.1 \\
    \midrule

    \multicolumn{1}{c}{\cellcolor{purple!20}\textbf{Phase Groups}} & \multicolumn{4}{c}{\cellcolor{purple!20}Venous} & \multicolumn{4}{c}{\cellcolor{purple!20}Arterial} & \multicolumn{4}{c}{\cellcolor{purple!20}Non-contrast} \\
    \cmidrule(lr){2-5}\cmidrule(lr){6-9}\cmidrule(lr){10-13}
    Method & Sen. & Spe. & AUC & F1 & Sen. & Spe. & AUC & F1 & Sen. & Spe. & AUC & F1 \\
    \midrule
    U-Net~\cite{ronneberger2015u} & 60.0 & 56.2 & 58.1 & 10.7 & 50.0 & 67.5 & 58.8 & 8.9 & 100.0 & 80.8 & 90.4 & 28.6 \\
    Swin UNETR-Tiny~\cite{tang2022self} & 74.3 & 42.1 & 58.2 & 10.3 & 50.0 & 60.0 & 55.0 & 7.4 & 0.0 & 54.2 & 27.1 & 0.0 \\
    Swin UNETR-Small~\cite{tang2022self} & 48.6 & 59.1 & 53.9 & 9.3 & 0.0 & 79.2 & 39.6 & 0.0 & 0.0 & 87.5 & 43.8 & 0.0 \\
    Swin UNETR-Base~\cite{tang2022self} & 54.3 & 60.9 & 57.6 & 10.7 & 25.0 & 68.3 & 46.7 & 4.7 & 0.0 & 91.7 & 45.9 & 0.0 \\
    Universal Model$^a$~\cite{liu2023clip} & 57.1 & 52.9 & 55.0 & 9.6 & 75.0 & 66.7 & 70.9 & 12.8 & 100.0 & 84.6 & 92.3 & 33.3 \\
    Universal Model$^b$~\cite{liu2023clip} & 28.6 & 71.0 & 49.8 & 7.5 & 25.0 & 86.7 & 55.9 & 9.5 & 0.0 & 96.2 & 48.1 & 0.0 \\ \hline
    nnU-Net~\cite{isensee2021nnu} & 71.4 & 79.1 & 75.3 & 22.6 & 25.0 & 75.8 & 50.4 & 5.9 & 100.0 & 87.5 & 93.8 & 40.0 \\
    ResEnc~\cite{isensee2024nnu} & 17.1 & 95.3 & 56.2 & 15.6 & 0.0 & 100.0 & 50.0 & 0.0 & 0.0 & 100.0 & 50.0 & 0.0 \\
    UNETR~\cite{hatamizadeh2022unetr} & 45.7 & 52.0 & 48.9 & 7.6 & 50.0 & 76.7 & 63.4 & 11.8 & 0.0 & 87.5 & 43.8 & 0.0 \\
    STU-Net-B~\cite{huang2023stu} & 71.4 & 76.4 & 73.9 & 20.7 & 50.0 & 67.5 & 58.8 & 8.9 & 0.0 & 79.2 & 39.6 & 0.0 \\
    MedNeXt~\cite{roy2023mednext} & 51.4 & 90.9 & 71.2 & 29.3 & 25.0 & 99.2 & 62.1 & 33.3 & 0.0 & 100.0 & 50.0 & 0.0 \\
    VSmTrans~\cite{liu2024vsmtrans} & 11.4 & 99.6 & 55.5 & 19.1 & 0.0 & 100.0 & 50.0 & 0.0 & 0.0 & 100.0 & 50.0 & 0.0 \\
    \bottomrule
    \end{tabular} 
    \label{tab:benchmark_ct_rate}
\end{table*}

\begin{table*}[t]
    \centering
    \tiny
    \caption{\textbf{Validation on \textsl{N--Europe}}. 
Architectures are grouped by framework. 
Subgroup analysis shows notable variation across \colorbox{orange!20}{age}, 
with both young (20$<$Age$\leq$40) and older (60$<$Age$\leq$80) groups exhibiting 
reduced F1 scores. 
Performance differences across \colorbox{blue!20}{sex} are moderate, with males 
typically achieving lower sensitivity and F1 than females. 
Across \colorbox{purple!20}{imaging phases} (non-contrast, arterial, venous), venous scans 
generally yield higher balanced AUC and F1, while non-contrast scans remain more challenging.}

    \begin{tabular}{p{0.26\linewidth}P{0.05\linewidth}P{0.05\linewidth}P{0.05\linewidth}P{0.05\linewidth}P{0.05\linewidth}P{0.05\linewidth}P{0.05\linewidth}P{0.05\linewidth}P{0.05\linewidth}P{0.05\linewidth}P{0.05\linewidth}P{0.05\linewidth}} 
    \toprule
    \multicolumn{1}{c}{\cellcolor{orange!20}\textbf{Age Groups}} & \multicolumn{4}{c}{\cellcolor{orange!20}20 \textless Age $\leq 40$} & \multicolumn{4}{c}{\cellcolor{orange!20}40 \textless Age $\leq 60$} & \multicolumn{4}{c}{\cellcolor{orange!20}60 \textless Age $\leq 80$} \\
    \cmidrule(lr){2-5}\cmidrule(lr){6-9}\cmidrule(lr){10-13}
    Method & Sen. & Spe. & AUC & F1 & Sen. & Spe. & AUC & F1 & Sen. & Spe. & AUC & F1 \\
    \midrule
    U-Net~\cite{ronneberger2015u} & 60.0 & 100.0 & 80.0 & 75.0 & 53.5 & 54.5 & 54.0 & 66.0 & 52.2 & 57.6 & 54.9 & 64.5 \\
    Swin UNETR-Tiny~\cite{tang2022self} & 60.0 & 100.0 & 80.0 & 75.0 & 68.1 & 45.5 & 56.8 & 76.3 & 61.5 & 44.4 & 53.0 & 70.6 \\
    Swin UNETR-Small~\cite{tang2022self} & 60.0 & 0.0 & 30.0 & 66.7 & 54.3 & 54.5 & 54.4 & 66.7 & 55.5 & 48.5 & 52.0 & 66.3 \\
    Swin UNETR-Base~\cite{tang2022self} & 60.0 & 0.0 & 30.0 & 66.7 & 58.6 & 50.0 & 54.3 & 69.7 & 51.8 & 54.5 & 53.2 & 63.8 \\
    Universal Model$^a$~\cite{liu2023clip} & 0.0 & 100.0 & 50.0 & 0.0 & 16.4 & 81.8 & 49.1 & 27.3 & 16.3 & 90.9 & 53.6 & 27.6 \\
    Universal Model$^b$~\cite{liu2023clip} & 0.0 & 100.0 & 50.0 & 0.0 & 10.3 & 86.4 & 48.4 & 18.3 & 13.5 & 90.9 & 52.2 & 23.4 \\ \hline
    nnU-Net~\cite{isensee2021nnu} & 40.0 & 0.0 & 20.0 & 50.0 & 49.1 & 68.2 & 58.7 & 63.3 & 48.2 & 58.6 & 53.4 & 61.1 \\
    ResEnc~\cite{isensee2024nnu} & 20.0 & 100.0 & 60.0 & 33.3 & 26.7 & 86.4 & 56.6 & 41.3 & 23.8 & 81.8 & 52.8 & 37.2 \\
    UNETR~\cite{hatamizadeh2022unetr} & 40.0 & 0.0 & 20.0 & 50.0 & 62.9 & 59.1 & 61.0 & 73.7 & 59.2 & 55.6 & 57.4 & 69.9 \\
    STU-Net-B~\cite{huang2023stu} & 20.0 & 0.0 & 10.0 & 28.6 & 44.0 & 81.8 & 62.9 & 59.6 & 45.5 & 62.6 & 54.1 & 59.0 \\
    MedNeXt~\cite{roy2023mednext} & 40.0 & 0.0 & 20.0 & 50.0 & 42.2 & 77.3 & 59.8 & 57.6 & 46.6 & 59.6 & 53.1 & 59.8 \\
    VSmTrans~\cite{liu2024vsmtrans} & 20.0 & 100.0 & 60.0 & 33.3 & 13.8 & 95.5 & 54.7 & 24.1 & 7.7 & 95.0 & 51.4 & 14.1 \\
    \midrule

    \multicolumn{1}{c}{\cellcolor{blue!20}\textbf{Sex Groups}} & \multicolumn{4}{c}{\cellcolor{blue!20}Female} & \multicolumn{4}{c}{\cellcolor{blue!20}Male} & \multicolumn{4}{c}{\cellcolor{orange!20}Age \textgreater 80} \\
    \cmidrule(lr){2-5}\cmidrule(lr){6-9}\cmidrule(lr){10-13}
    Method & Sen. & Spe. & AUC & F1 & Sen. & Spe. & AUC & F1 & Sen. & Spe. & AUC & F1 \\
    \midrule
    U-Net~\cite{ronneberger2015u} & 49.1 & 48.1 & 48.6 & 62.1 & 60.9 & 60.4 & 60.7 & 71.8 & 57.9 & 49.1 & 53.5 & 69.9 \\
    Swin UNETR-Tiny~\cite{tang2022self} & 60.8 & 43.2 & 52.0 & 71.3 & 69.0 & 41.7 & 55.4 & 75.8 & 67.4 & 36.4 & 51.9 & 76.1 \\
    Swin UNETR-Small~\cite{tang2022self} & 52.4 & 46.9 & 49.7 & 64.9 & 63.0 & 50.0 & 56.5 & 72.3 & 60.7 & 47.3 & 54.0 & 72.0 \\
    Swin UNETR-Base~\cite{tang2022self} & 49.1 & 51.9 & 50.5 & 62.4 & 66.4 & 53.1 & 59.8 & 75.1 & 63.5 & 50.9 & 57.2 & 74.3 \\
    Universal Model$^a$~\cite{liu2023clip} & 12.8 & 85.2 & 49.0 & 22.2 & 18.1 & 87.5 & 52.8 & 29.9 & 13.9 & 80.0 & 47.0 & 23.8 \\
    Universal Model$^b$~\cite{liu2023clip} & 8.8 & 86.4 & 47.6 & 15.8 & 16.2 & 86.5 & 51.4 & 27.2 & 11.7 & 78.2 & 45.0 & 20.3 \\ \hline
    nnU-Net~\cite{isensee2021nnu} & 48.0 & 56.8 & 52.4 & 61.8 & 55.8 & 53.1 & 54.5 & 67.1 & 56.8 & 43.6 & 50.2 & 68.7 \\
    ResEnc~\cite{isensee2024nnu} & 21.0 & 76.5 & 48.8 & 33.6 & 27.3 & 83.3 & 55.3 & 41.7 & 23.4 & 74.5 & 49.0 & 36.8 \\
    UNETR~\cite{hatamizadeh2022unetr} & 52.4 & 51.9 & 52.2 & 65.3 & 64.1 & 51.0 & 57.6 & 73.3 & 55.1 & 41.8 & 48.5 & 67.2 \\
    STU-Net-B~\cite{huang2023stu} & 39.6 & 60.5 & 50.1 & 54.1 & 54.6 & 61.5 & 58.1 & 67.0 & 49.6 & 50.9 & 50.3 & 63.1 \\
    MedNeXt~\cite{roy2023mednext} & 42.6 & 56.8 & 49.7 & 56.8 & 56.2 & 58.3 & 57.3 & 68.0 & 54.3 & 47.3 & 50.8 & 66.9 \\
    VSmTrans~\cite{liu2024vsmtrans} & 7.1 & 96.3 & 51.7 & 13.2 & 9.7 & 95.8 & 52.8 & 17.6 & 7.2 & 98.2 & 52.7 & 13.5 \\
    \midrule

    \multicolumn{1}{c}{\cellcolor{purple!20}\textbf{Phase Groups}} & \multicolumn{4}{c}{\cellcolor{purple!20}Non-contrast} & \multicolumn{4}{c}{\cellcolor{purple!20}Arterial} & \multicolumn{4}{c}{\cellcolor{purple!20}Venous} \\
    \cmidrule(lr){2-5}\cmidrule(lr){6-9}\cmidrule(lr){10-13}
    Method & Sen. & Spe. & AUC & F1 & Sen. & Spe. & AUC & F1 & Sen. & Spe. & AUC & F1 \\
    \midrule
    U-Net~\cite{ronneberger2015u} & 54.5 & 58.3 & 56.4 & 67.7 & 48.6 & 58.3 & 53.5 & 61.9 & 56.4 & 53.5 & 55.0 & 68.1 \\
    Swin UNETR-Tiny~\cite{tang2022self} & 33.8 & 75.0 & 54.4 & 49.1 & 47.5 & 50.0 & 48.8 & 60.3 & 73.0 & 37.2 & 55.1 & 78.8 \\
    Swin UNETR-Small~\cite{tang2022self} & 22.1 & 75.0 & 48.6 & 35.0 & 49.2 & 52.8 & 51.0 & 62.0 & 63.9 & 45.0 & 54.5 & 73.1 \\
    Swin UNETR-Base~\cite{tang2022self} & 46.8 & 58.3 & 52.6 & 61.0 & 44.7 & 58.3 & 51.5 & 58.4 & 62.0 & 50.4 & 56.2 & 72.2 \\
    Universal Model$^a$~\cite{liu2023clip} & 18.2 & 83.3 & 50.8 & 30.1 & 7.8 & 88.9 & 48.4 & 14.2 & 17.0 & 86.0 & 51.5 & 28.4 \\
    Universal Model$^b$~\cite{liu2023clip} & 10.4 & 100.0 & 55.2 & 18.8 & 5.6 & 88.9 & 47.3 & 10.4 & 14.4 & 84.5 & 49.5 & 24.5 \\ \hline
    nnU-Net~\cite{isensee2021nnu} & 45.5 & 58.3 & 51.9 & 59.8 & 43.6 & 50.0 & 46.8 & 56.7 & 54.7 & 55.8 & 55.3 & 66.9 \\
    ResEnc~\cite{isensee2024nnu} & 13.0 & 100.0 & 56.5 & 23.0 & 14.0 & 83.3 & 48.7 & 23.8 & 28.0 & 77.5 & 52.8 & 42.3 \\
    UNETR~\cite{hatamizadeh2022unetr} & 15.6 & 83.3 & 49.5 & 26.4 & 40.8 & 66.7 & 53.8 & 55.3 & 67.7 & 44.2 & 56.0 & 75.8 \\
    STU-Net-B~\cite{huang2023stu} & 20.8 & 75.0 & 47.9 & 33.3 & 34.1 & 63.9 & 49.0 & 48.2 & 53.3 & 58.9 & 56.1 & 66.0 \\
    MedNeXt~\cite{roy2023mednext} & 41.6 & 66.7 & 54.2 & 56.6 & 32.4 & 52.8 & 42.6 & 45.7 & 54.5 & 58.1 & 56.3 & 67.0 \\
    VSmTrans~\cite{liu2024vsmtrans} & 2.6 & 100.0 & 51.3 & 5.1 & 4.5 & 100.0 & 52.3 & 8.6 & 10.1 & 94.6 & 52.4 & 18.2 \\
    \bottomrule
    \end{tabular} 
    \label{tab:benchmark_totalsegmentator}
\end{table*}

\end{document}